\documentclass[fleqn,10pt]{wlscirep}
\usepackage[utf8]{inputenc}
\usepackage[T1]{fontenc}
\DeclareUnicodeCharacter{1D467}{\ensuremath{\mathit{z}}}
\DeclareUnicodeCharacter{1D44D}{\ensuremath{\mathit{Z}}}
\usepackage{bm}
\usepackage{graphicx}
\usepackage{svg}
\usepackage{graphicx,verbatim}
\usepackage{subcaption}
\usepackage{caption}
\usepackage[export]{adjustbox}
\usepackage{booktabs}
\usepackage{etoolbox}
\usepackage{float}
\usepackage{array}
\usepackage{xcolor}
\usepackage[commandnameprefix=always,final]{changes}
\usepackage{pdfpages}
\newcolumntype{L}[1]{>{\raggedright\arraybackslash}m{#1}}
\newcolumntype{C}[1]{>{\centering\arraybackslash}m{#1}}

\title{MedicalPatchNet: A Patch-Based Self-Explainable AI Architecture for Chest X-ray Classification}

\author[1,*]{Patrick Wienholt}
\author[1]{Christiane Kuhl}
\author[2,3,4,5,6]{Jakob Nikolas Kather}
\author[1]{Sven Nebelung}
\author[1]{Daniel Truhn}
\affil[1]{Department of Diagnostic and Interventional Radiology of the University Hospital RWTH Aachen, Aachen, Germany}
\affil[2]{Else Kroener Fresenius Center for Digital Health, Technical University Dresden, Dresden, Germany}
\affil[3]{Department of Medicine III, University Hospital RWTH Aachen, Aachen, Germany}
\affil[4]{Pathology \& Data Analytics, Leeds Institute of Medical Research at St James's, University of Leeds, Leeds, United Kingdom}
\affil[5]{Department of Medicine I, University Hospital Dresden, Dresden, Germany}
\affil[6]{Medical Oncology, National Center for Tumor Diseases (NCT), University Hospital Heidelberg, Heidelberg, Germany}

\affil[*]{e-mail: pwienholt@ukaachen.de}

\begin{abstract}
Deep neural networks excel in radiological image classification but frequently suffer from poor interpretability, limiting clinical acceptance.
We present MedicalPatchNet, an inherently self-explainable architecture for chest X-ray classification that transparently attributes decisions to distinct image regions.
MedicalPatchNet splits images into non-overlapping patches, independently classifies each patch, and aggregates predictions, enabling intuitive visualization of each patch’s diagnostic contribution without post-hoc techniques.
Trained on the CheXpert dataset (223,414 images), MedicalPatchNet matches the classification performance (AUROC 0.907 vs. 0.908) of EfficientNet\chreplaced{V2-S}{-B0}, while \chdeleted{substantially} improving interpretability:
MedicalPatchNet demonstrates \chdeleted{substantially} improved interpretability with higher pathology localization accuracy (mean hit-rate 0.485 vs. 0.376 with Grad-CAM) on the CheXlocalize dataset.
By providing explicit, reliable explanations accessible even to non-AI experts, MedicalPatchNet mitigates risks associated with shortcut learning, thus improving clinical trust.
Our model is publicly available with reproducible training and inference scripts and contributes to safer, explainable AI-assisted diagnostics across medical imaging domains.
We make the code publicly available: \href{https://github.com/TruhnLab/MedicalPatchNet}{github.com/TruhnLab/MedicalPatchNet}

\end{abstract}

\begin{document}

\flushbottom
\maketitle

\thispagestyle{empty}

\section*{Introduction}
Neural networks are increasingly integrated into clinical workflows \cite{Shahid2019applications}.
In particular for radiological tasks they have shown remarkable results.
Applications range from breast cancer screening \cite{Marinovich2023BreastCancer}, where neural networks can identify suspicious cases earlier than human experts, over mortality prediction in intensive care units \cite{Xia2012mortality}, and diagnosing common diseases from chest X-rays, where deep learning models have consistently proven beneficial.\chadded{ Similar methodological trends can also be observed in a range of recent studies on machine learning and medical data analysis.\cite{REV1YIN202487,REV2JiangYYCHYWFSLLZCZHL24,REV3FENG2025143668,REV4shi2023b2,REV5LI2024108080,REV6Gan25,REV7Gan25a,REV8,REV9xu2023clinical}}%\ct{REV_1,REV_2,REV_3,REV_4,REV_5,REV_6,REV_7,REV_8,REV_9}}

Despite their strong and sometimes superior performance compared to expert radiologists \cite{McKinney2020international}, these models often lack explainability regarding their decision-making process \cite{Elton2020HardExplainableAI}.
Providing explanations is crucial for correctly interpreting the results, especially in cases where the network's behavior differs from the radiologist's expectations.
Because neural networks often rely on different image features than humans \cite{Makino2022Differences}, knowing the underlying reasons for a decision is particularly important.
Without explainability, a neural network with superior performance is of reduced value, as a medical professional may be unable to determine the reasoning behind the decision, making it difficult to assess its trustworthiness.
Furthermore, explanations provided by neural networks should be easily comprehensible, especially for the people using them, not just for the people designing them.
To be practically useful in clinical settings, individuals without extensive deep learning knowledge should be able to easily and intuitively interpret the decisions.

However, most modern deep learning architectures have been designed primarily to maximize performance rather than explainability.
This results in architectures that operate as so-called black boxes, making it difficult to gain insight into them \cite{Elton2020HardExplainableAI}.
Nevertheless, since explainability is essential, methods have been developed to render these black box approaches post-hoc explainable.
Examples include Grad-CAM \cite{Selvaraju2017GradCAM}, GradCam++ \cite{Chattopadhyay2018GradCam++}, Eigen-CAM \cite{Muhammad2020EigenCAM}, and other methods \cite{Sundararajan2017Integratedgradients,Shrikumar2017DeepLIFT,Bach2015LRP,Zeiler2014UnderstandCNN}.
These techniques are widely utilized in medical contexts \cite{Muhammad2024UnveilingTheBlackBox}, as they generate saliency maps that attempt to shed light on the underlying decision process.
However, post-hoc explainability methods have inherent limitations \cite{Rudin2019StopExplainingBlackBox}.
Previous research has shown that common saliency methods like Grad-CAM do not necessarily highlight the most critical regions influencing the decision \cite{Saporta2022CheXlocalize,Mittelstadt2019ExplainingExplanationsInAI}.

Correctly interpreting results of methods like Grad-CAM requires deep understanding of gradient-based mechanisms in deep learning, a level of knowledge beyond what is typically expected from clinical personnel who should routinely use these networks.
Depending on these methods, which produce visually appealing but potentially misinterpreted saliency maps, may create a false sense of comprehension regarding the network's decision-making process.
Consequently, this misplaced trust could lead to relying on predictions that are not truly trustworthy.
Rudin et al. \cite{Rudin2019StopExplainingBlackBox} therefore argues against using these post-hoc methods, as they produce misleading and unreliable explanations, and instead advocate for inherently explainable approaches.
Self-explainable AI refers to designing AI that is explainable by design \cite{Hou2024SXAI,Rudin2019StopExplainingBlackBox}.
\chreplaced{One approach is to learn, for each class, a set of representative image patches that capture characteristic visual patterns, and then classify new images by comparing their local regions to these class-specific prototypes and aggregating the resulting evidence}{One approach involves comparing image parts to prototypical parts of other images, making predictions based on evidence provided by these prototypical parts} \cite{Chen2019ThisLooksLikeThat}.

A non-gradient-based approach is to occlude parts of the image and then evaluate the change in the output compared to the non-occluded image \cite{Zeiler2014UnderstandCNN}.
To make brain MRI scan decisions explainable, Zhang et al. \cite{Zhang2023sMRI-PatchNet} developed a method that identifies and extracts the most discriminative 3D patches.
Their network then utilizes only these patches, allowing the decision to be known to rely exclusively on them, rather than on other patches.
Additionally, to interpret model predictions, patches are perturbed by zeroing them out, and the network’s behavior is observed.
This allows for quantifying the respective contributions.
Previous studies have used different patch-based approaches to address classification tasks.
For instance, Oh et al. \cite{Oh2020PatchCov19} and Szczepanski et al. \cite{Szczepanski2022POTHER} classified COVID-19 by extracting randomly overlapping image patches, training a model on these patches individually rather than on the entire image.
This patch-wise approach was shown to effectively reduce overfitting, allowing more efficient training on limited datasets.
However, their interpretability depended on applying Grad-CAM \cite{Selvaraju2017GradCAM} to each patch individually and aggregating the resulting saliency maps for visualization, thereby inheriting the limitations of post-hoc methods.

In recent work, several Multiple Instance Learning \chadded{(MIL)} frameworks \cite{Ilse18MIL,Shao21TransMIL,lu2021data} similarly divide images into patches and aggregate patch-level evidence.
Ilse et al. \cite{Ilse18MIL} propose attention‑based MIL, learning permutation‑invariant attention weights to identify diagnostically relevant instances while preserving interpretability.
Lu et al. \cite{Shao21TransMIL} introduce CLAM, which combines attention pooling with instance‑level clustering to refine feature separation and generate high‑contrast heatmaps under weak supervision.
Shao et al. \cite{lu2021data} present TransMIL, a transformer-based correlated MIL approach that models spatial and morphological relationships between patches to improve performance and visual interpretability.

A related field that partially overlaps with explainability is weakly supervised segmentation \cite{Zhu2023WeaklySupervisedSegmentationReview,Ciga2021WeaklySupervisedPathologySegmentation}.
In this approach, segmentation masks are derived solely from classification labels.
Techniques such as Grad-CAM \cite{Selvaraju2017GradCAM} and attention maps \cite{Abnar2020AttentionRollout} are frequently employed in this context.
Unlike explainability techniques, the primary goal of this field is to produce accurate segmentation masks, without necessarily focusing on transparency regarding biases or underlying decision mechanisms.
This stands in contrast to explainability methods, whose focus is to reflect the true underlying decision process of the neural network.
An explainability method that correctly segments a pleural effusion, for example, but does so despite the true underlying reason for classification being elsewhere, is a poor explainability method—even if the segmentation of the pathology is correct.
It has been shown that for chest X-rays, networks tend to rely on dataset biases when classifying pathologies \cite{Zech2018VariableGeneralizationPerformance,DeGrave2021Shortcut}.
Zech et al., for example, show that their network sometimes relied on the presence of specific laterality markers\chadded{---small radiopaque ``L'' or ``R'' labels that indicate the patient's left or right side on the radiograph---}to detect pneumonia \cite{Zech2018VariableGeneralizationPerformance}.
Similarly, DeGrave et al. \cite{DeGrave2021Shortcut} demonstrated that some networks trained for COVID-19 classification relied on shortcuts—such as laterality markers or radiopacity at image borders—rather than on genuine pathological features.

All these factors highlight the need for a neural network for clinical applications that combines high performance, self-explainability, and ease of interpretation, even for users without extensive deep learning expertise.
This is precisely where our method, MedicalPatchNet, stands out.
We introduce MedicalPatchNet, which demonstrates strong performance in chest X-ray classification.
Instead of complex interpretation techniques that require advanced insight into a model’s inner workings, our approach is straightforward yet effective: MedicalPatchNet splits the input image into patches, classifies each patch independently, and determines the final diagnosis by averaging these patch-level predictions.
Visualizing the classification results at the patch level clearly illustrates each patch’s contribution to the overall decision.
This explicit visualization ensures that every influential visual feature is transparently represented.
Our architecture can be viewed as a special case of the Multiple Instance Learning frameworks prevalent in computational pathology.
In contrast to such approaches, which often rely on complex, learnable aggregation functions like attention mechanisms, MedicalPatchNet employs a fixed, non-learnable arithmetic mean of the patch logits.
This simplification makes our model a more constrained but inherently transparent variant, where each patch's contribution to the final decision is explicitly defined and not learned.

\begin{figure}
\centering
\includegraphics[width=\linewidth]{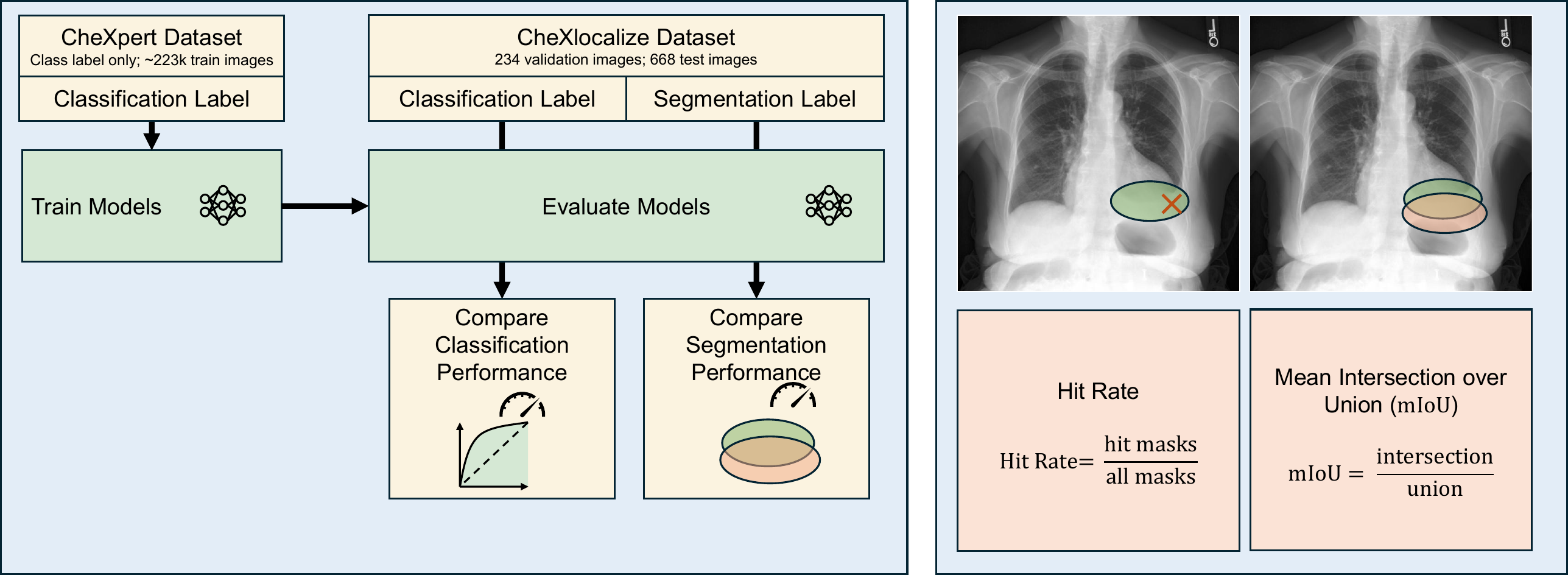}
\caption{\chreplaced{ Evaluation framework for classification and localization.
Models are trained on the CheXpert dataset using image-level classification labels only.
Evaluation is performed on CheXlocalize, which provides image-level labels and radiologist-annotated pixel-wise segmentation masks (for 10 findings).
We report both classification performance (e.g., AUROC) and localization performance.
Localization is quantified by (i) hit rate, i.e., the fraction of cases where the pixel with maximum attribution/saliency lies inside the corresponding ground-truth segmentation mask, and (ii) mean Intersection over Union (mIoU) between a binarized localization map and the ground-truth mask.}{Evaluation framework for classification and localization. Models are trained on the CheXpert dataset using only classification labels. For evaluation, the CheXlocalize dataset is used, which provides ground-truth segmentation masks. We compare both classification performance (e.g., using AUROC) and localization performance. For localization, saliency maps from our proposed MedicalPatchNet are compared against a standard EfficientNet-B0 explained by post-hoc methods (Grad-CAM, Grad-CAM++, and Eigen-CAM). The primary localization metrics, illustrated on the right, are the Hit Rate and the Mean Intersection over Union (mIoU).}}
\label{fig:metricsOverview}
\end{figure}

Our proposed architecture can accommodate various backbone architectures.
For our study, we chose EfficientNet\chreplaced{V2-S}{-B0} \cite{EfficientNetV2Tan2021}, as it provides good performance with relatively low computational requirements.
First, we demonstrate that MedicalPatchNet with an EfficientNet\chreplaced{V2-S}{-B0} backbone achieves performance comparable to the standard EfficientNet\chreplaced{V2-S}{-B0} architecture for classification of the 14 classes on the CheXpert \cite{Irvin2019Chexpert} dataset.
Maintaining competitive performance is crucial, as explainability does not necessarily correlate with diagnostic accuracy \cite{Rudin2019StopExplainingBlackBox}.
Second, we evaluate the saliency maps generated by MedicalPatchNet on the CheXlocalize dataset \cite{Saporta2022CheXlocalize} (see Figure~\ref{fig:metricsOverview}), comparing its localization capabilities against Grad-CAM, Grad-CAM++, and Eigen-CAM using the EfficientNet\chreplaced{V2-S}{-B0}.
Here, MedicalPatchNet outperforms these three commonly used explainability methods.

\section{\chadded{Material and} Methods}

\subsection{\chreplaced{Material: Datasets and annotations}{Dataset}}
\chdeleted{We trained MedicalPatchNet as well as EfficientNet\chreplaced{V2-S}{-B0} using the CheXpert dataset \cite{Irvin2019Chexpert}, labeled by the VisualCheXpert labeler \cite{Jain2021VisualCheXbert}.}
\chadded{Our study is based on the CheXpert dataset \cite{Irvin2019Chexpert}, labeled by the VisualCheXpert tool \cite{Jain2021VisualCheXbert}, which provides the large-scale chest X-ray material and image-level annotations used throughout our experiments.}
The training dataset comprised 223,414 chest X-rays from 64,540 patients, each labeled with binary classifications across $C = 14$ distinct classes.\chadded{ The CheXpert label set includes No Finding, Lung Opacity, Lung Lesion, Edema, Consolidation, Pneumonia, Atelectasis, Pneumothorax, Pleural Effusion, Pleural Other, Cardiomegaly, Enlarged Cardiomediastinum, Fracture, and Support Devices}.
\chadded{All radiographs were preprocessed once before training by cropping them to a square field of view and isotropically resizing them to $1024 \times 1024$ pixels. For all experiments, we used the image-level labels provided by the VisualCheXpert \cite{Jain2021VisualCheXbert} tool.}
Of these training images, 191,027 have a frontal projection, while 32,387 are lateral views.
The frontal and lateral views are not combined or used together but are treated as two separate images for both training and evaluation.
To evaluate how effectively the methods can localize pathologies, we use the CheXlocalize dataset \cite{Saporta2022CheXlocalize}, which consists of a validation set comprising 234 chest X-rays from 200 patients and a test set containing 668 chest X-rays from 500 patients.
These images are distinct from the CheXpert training data; they correspond to CheXpert validation and test sets enhanced with segmentation masks annotated by radiologists.\chadded{ For 10 of the 14 CheXpert labels (Lung Opacity, Atelectasis, Cardiomegaly, Consolidation, Edema, Enlarged Cardiomediastinum, Lung Lesion, Pleural Effusion, Pneumothorax, and Support Devices), CheXlocalize provides pixel-wise segmentation masks, which we later use to quantify localization performance.}

\subsection{MedicalPatchNet}

\begin{figure}
\centering
\includegraphics[width=\linewidth]{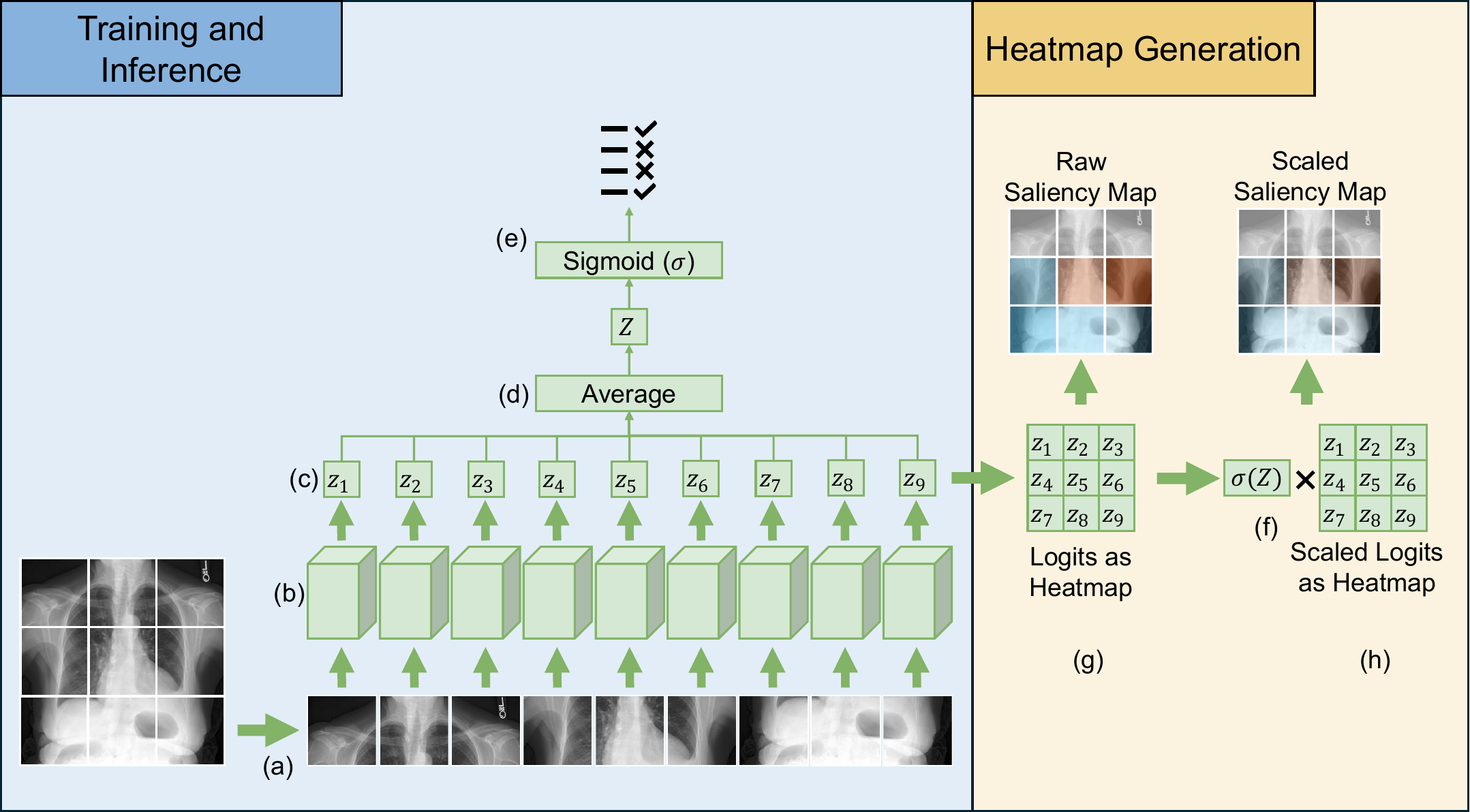}
\caption{Initially, the image is divided into patches (a), \chreplaced{each patch is independently processed}{each independently processed} by \chreplaced{the same}{an identical} EfficientNet\chreplaced{V2-S}{-B0} (b). The resulting patch logits\chadded{, i.e., raw pre-sigmoid class scores} (c), are averaged (d). After applying the sigmoid activation function, the output (e) provides the final classification results of MedicalPatchNet. Multiplying the raw patch logits by the classification probabilities generates scaled patch logits (f). A saliency map can be derived either from raw patch logits (g) or scaled patch logits (h),  illustrating each patch’s contribution to the final decision.}
\label{fig:MedicalPatchNetOverview}
\end{figure}

The core idea of MedicalPatchNet\chadded{, as shown in Figure \ref{fig:MedicalPatchNetOverview},} is to divide the image into patches, predict each patch independently of the others, and then combine the predictions into a global prediction.
Since the global prediction is just a scaled sum of the local predictions, visualizing these local predictions allows direct quantification of each patch’s contribution to the final classification.\chadded{ By construction, all patches are processed independently with a shared backbone and their logits are combined by a fixed arithmetic mean, without any learnable interaction between patches. This simple aggregation strategy allows MedicalPatchNet to act as a drop-in replacement for a conventional image-level classifier that uses the same backbone and training labels, while additionally providing an explicit patch-wise decomposition of the decision.}

As a backbone, we chose EfficientNet\chreplaced{V2-S}{-B0} \cite{EfficientNetV2Tan2021}; however, other architectures are also compatible.
Since the chest X-rays are grayscale, we adapt the backbone network to accept a single input channel instead of three by modifying the first convolution layer.
As a first step, MedicalPatchNet partitions an input image of size $S \times S$ pixels into $P \times P$ non-overlapping patches, each sized $p \times p$ pixels, where $p = \frac{S}{P}$.
These patches $x_i$, for $i \in [1,P^2]$, are stacked along the batch dimension for processing.
Thus, a batch containing $B$ images results in a batch size of $BP^2$ patches, all processed simultaneously by the backbone neural network.
We modify the final layer to output $C$ logits\chadded{, i.e.\ raw, pre-sigmoid scores for each class}, producing an output vector $z_i \in \mathbb{R}^{C}$ for each patch.
It contains one logit for each class.
The global logits $Z \in \mathbb{R}^{C}$ are computed by averaging these local logits: $Z = \frac{1}{P^2}\sum_{i}^{P^2}z_i$.
The final classification output is obtained by applying the sigmoid function: $\hat{y} = \sigma(Z)$.
The idea of applying the sigmoid after averaging the logits is to avoid constraining the patch logits' influence on the global logits.

Using the raw logits vector $z_i$ and the classification vector $\hat{y}$, we calculate scaled patch logits as $\hat{z}_i = \hat{y} \odot z_i$ through element-wise multiplication.
By scaling the raw logits with classification probabilities, we incorporate a limited degree of global context into individual patches in an interpretable manner.

\begin{figure}[htbp]
    \centering
    \begin{subfigure}[t]{0.24\textwidth}
        \centering
        \includegraphics[width=\textwidth]{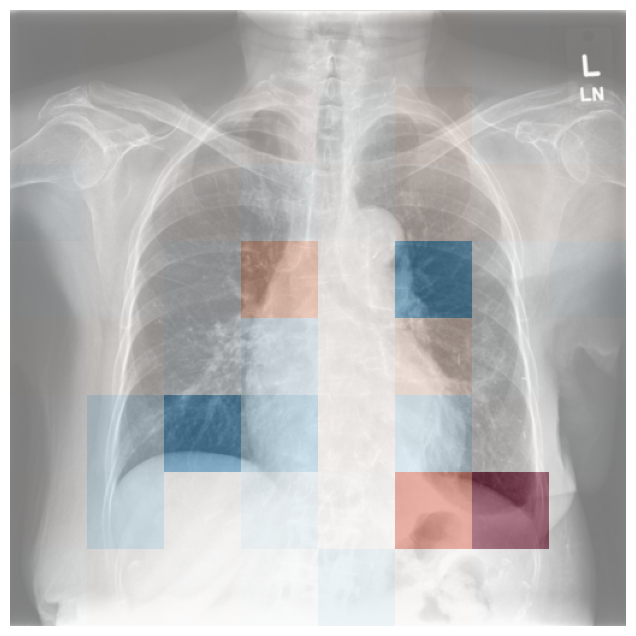}
        \caption{Shift 1 Patch, 64 Pixel}
        \label{fig:shifted_a}
    \end{subfigure}
    \hfill
    \begin{subfigure}[t]{0.24\textwidth}
        \centering
        \includegraphics[width=\textwidth]{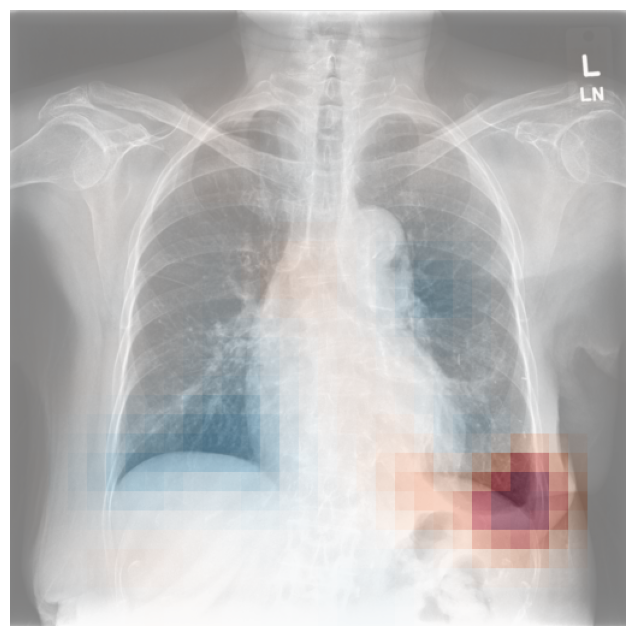}
        \caption{Shift 1/4 Patch, 16 Pixel}
        \label{fig:shifted_b}
    \end{subfigure}
    \hfill
    \begin{subfigure}[t]{0.24\textwidth}
        \centering
        \includegraphics[width=\textwidth]{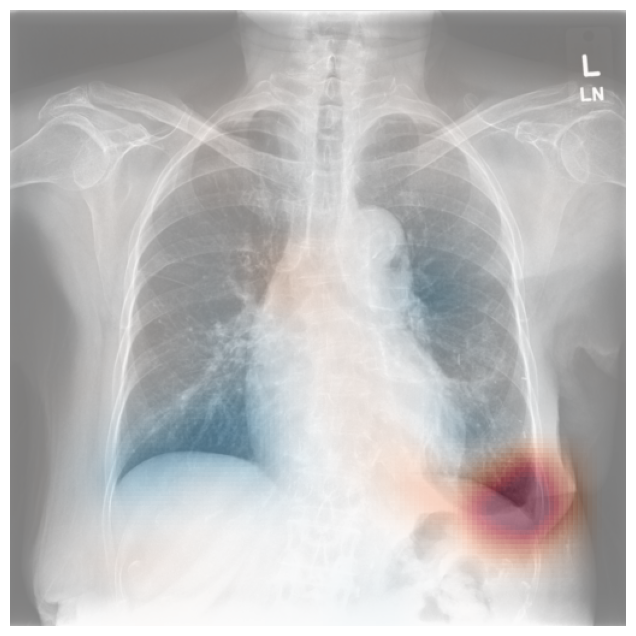}
        \caption{Shift 1/16 Patch, 4 Pixel}
        \label{fig:shifted_c}
    \end{subfigure}
    \hfill
    \begin{subfigure}[t]{0.24\textwidth}
        \centering
        \includegraphics[width=\textwidth]{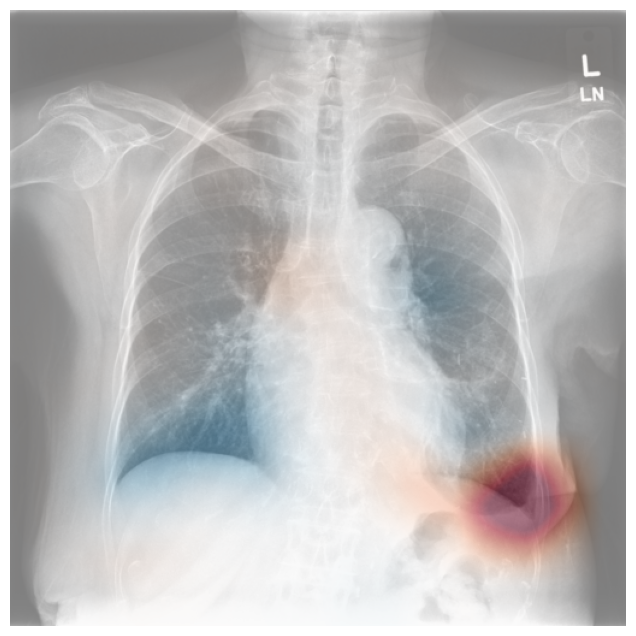}
        \caption{Shift 1/64 Patch, 1 Pixel}
        \label{fig:shifted_d}
    \end{subfigure}
\caption{The saliency map shown in (a) illustrates the influence of each patch on the final classification of pleural effusion. \chreplaced{Patches with large positive logits are shown in red and represent strong evidence supporting the class, patches with large negative logits are shown in blue and represent evidence against the class, and patches visualized in light grey or white have logits close to zero and therefore contribute only minimally to the final decision (effectively ``abstaining'' from the vote).}{Red patches ``vote'' for the classification, while blue patches ``vote'' against it. The closer the patch is to white, the less influence it has on the classification decision.} Image (a) shows a direct visualization of the patch logits from one forward pass. When shifting the image and averaging the generated saliency maps, smoother maps can be produced, as seen in (b), (c), and (d), although this requires more forward passes.}
    \label{fig:shifted}
\end{figure}

Saliency maps derived from raw or scaled logits are generated in the same manner.
To create a coarse saliency map as illustrated in Figure \ref{fig:shifted_a}, we visualize the patch logits as an overlay saliency map.
The saliency map values can be interpreted as follows:
a high positive value indicates that the network strongly ``decides'' the entire image belongs to the specific class based exclusively on the content of that patch.
Conversely, a high negative value signifies that, based solely on the patch, the network strongly ``decides'' against classification into that class.
Patches that are irrelevant to the classification decision have values close to zero.
Computing global logits by averaging local patch logits can thus be viewed as patches ``voting'' for or against the presence of a class, with different strength.

To improve the coarse appearance of patch-based saliency maps, we generate multiple spatially shifted saliency maps and overlay them.
First, we define a shift offset $o$, \chreplaced{which determines the stride of the image shifts used for additional saliency map generation}{determining the extent of image shifts for additional saliency map generation}.
Given a patch size $p$, an offset $o$ results in $\left(\frac{p}{o}\right)^2$ total shifts, since shifts of size $o$ are performed along both spatial dimensions.
Image borders are padded with zero values.
The smoothed saliency map is obtained by averaging all shifted patch saliency maps, which reduces block artifacts.
\chreplaced{The shift offset $o$ is adjustable as a hyperparameter and directly trades off explanation resolution and computational cost.}{The shift offset $o$ is adjustable as a hyperparameter.}
An offset of $o = p$ generates only one saliency map, equivalent to direct patch prediction, \chreplaced{so that each of the original non-overlapping image patches is evaluated exactly once.}{requiring just a single forward pass.}
Conversely, setting the offset to $o = 1$ results in pixel-wise shifts, necessitating $p^2$ forward passes to produce the saliency map shown in \ref{fig:shifted_d}.
\chadded{More generally, for a square image of size $S \times S$, the number of evaluated patches, and thus the explanation runtime, scales proportionally to $\left(\frac{S}{o}\right)^2$.}
It should be noted that MedicalPatchNet with a patch size equal to the image size generates only one output vector, representing the entire image.
\chreplaced{Thus, using this configuration is equivalent to a standard EfficientNetV2-S image-level classifier; only the first convolutional and final linear layers are adapted to our data (single-channel input and the chosen number of output classes).}{Thus, using this configuration is equivalent to using EfficientNet-B0 with adapted first convolutional and final linear layers.}

\subsection{Training}

We trained both MedicalPatchNet and EfficientNet\chreplaced{V2-S}{-B0} on chest X-rays rescaled to $512 \times 512$ pixels.
\chadded{ Concretely, all images were first preprocessed to a square format of $1024 \times 1024$ pixels as described above and, at train and test time, were then resized to $512 \times 512$ pixels.}
MedicalPatchNet divides each image into $8 \times 8$ patches, each measuring $64 \times 64$ pixels.
We initialized EfficientNet\chreplaced{V2-S}{-B0}, both standalone and as the backbone for MedicalPatchNet, with pretrained weights from ImageNet-1k \cite{Deng2009ImageNet}.
All models were trained exclusively on classification labels from the CheXpert training dataset \cite{Irvin2019Chexpert}, generated using the VisualCheXpert labeling tool \cite{Jain2021VisualCheXbert}.
During training, no segmentation masks or explicit localization information were provided.
The loss was computed only for the global classification output and backpropagated through the entire network.

We trained the models using PyTorch version 2.5.1 on Nvidia H100 GPUs.
We set the batch size to 16 images.
For MedicalPatchNet, this corresponded to an effective batch size of $16 \times 8 \times 8 = 1024$ patches, as all patches from one image were processed simultaneously through the backbone network.
As an optimizer, we used AdamW \cite{Loshchilov2019AdamW} and trained for 20 epochs with a OneCycle learning rate schedule \cite{Smith2018OneCycleLR}, peaking at a learning rate of $1 \cdot 10^{-4}$ after the first 5\% of training.
The complete training of a single model required approximately four hours.
\chreplaced{Image augmentation involved randomly cropping a square area covering between 50\% and 100\% of the original image, followed by random rotations within $\pm 5^{\circ}$ and brightness adjustments with a factor between 0.7 and 1.3.}{During training, we applied a composed augmentation pipeline consisting of a RandomResizedCrop that samples a square region covering between 50\% and 100\% of the preprocessed image and resizes it to $512 \times 512$ pixels, followed by a random in-plane rotation within $\pm 5^{\circ}$ and a brightness jitter with a multiplicative factor in the range [0.7, 1.3]. For validation and test images, only a deterministic resize to $512 \times 512$ pixels without any stochastic augmentation was used.}
\chadded{All images were loaded as single-channel grayscale, converted to tensors, and intensity values were scaled to the range $[0,1]$; no additional per-image standardization or histogram equalization was applied.}

\subsection{Evaluation}

To confirm that dividing images into patches did not negatively impact classification performance, we compared the performance of MedicalPatchNet against EfficientNet\chreplaced{V2-S}{-B0}.
\chadded{Because both models share the same backbone, training data, and optimization pipeline, this comparison effectively isolates the impact of enforcing patch-wise independence with fixed averaging aggregation versus standard image-level classification on overall diagnostic performance.}
As performance metrics, we used sensitivity, specificity, accuracy, and the Area Under the Receiver Operating Characteristic curve (AUROC).
Sensitivity (true positive rate) measures the proportion of correctly identified positives, while specificity (true negative rate) measures the proportion of correctly identified negative cases.
Accuracy is the overall proportion of correctly classified instances.
AUROC quantifies the model's ability to distinguish between classes by summarizing its performance across all possible classification thresholds, with a higher AUROC indicating better class separability.
An AUROC of 0.5 represents a random classifier.
%Both networks were trained exclusively on the CheXpert training subset.
For sensitivity, specificity, and accuracy, optimal classification thresholds were selected for each class by maximizing the sum of sensitivity and specificity on the validation dataset.
These thresholds were then used during evaluation on the held-out test dataset.
We estimated 95\% confidence intervals for performance metrics using bootstrapping, resampling the test dataset with replacement 100,000 times.

We evaluated the localization performance of the saliency maps produced by MedicalPatchNet and those generated by Grad-CAM, Grad-CAM++, and Eigen-CAM applied to EfficientNet\chreplaced{V2-S}{-B0}, using the CheXlocalize dataset \cite{Saporta2022CheXlocalize}.\chdeleted{ This dataset is dedicated solely to evaluation, as the models were trained on the CheXpert training dataset.}
\chdeleted{The provided segmentation masks are for 10 out of the 14 classes from the CheXpert dataset.}
\chadded{ We used CheXlocalize purely as an external evaluation set; all model parameters were learned exclusively from the CheXpert training subset.}

\chadded{In addition to these post-hoc baselines, we also trained two prototype-based, inherently interpretable methods (ProtoPNet\cite{Chen2019ThisLooksLikeThat} and PIPNet\cite{PIPNet}) using the authors' public reference implementations and their default training hyperparameters on CheXpert. In our setting, however, both approaches failed to yield a reliable 14-label multi-label chest X-ray classifier (ProtoPNet remained close to chance level, and PIPNet collapsed to meaningful predictions for only a small subset of labels). Because meaningful explanations require a non-trivial classifier, we do not treat these models as competitive localization baselines in the main text; instead, we provide the complete quantitative results and a detailed discussion of possible reasons for this failure in the Supplementary Material.}

We evaluated the saliency maps based on their hit rate, defined as follows \cite{Saporta2022CheXlocalize}: \chreplaced{For each ground-truth segmentation mask, we identify the pixel with the maximum saliency value. 
If this pixel lies within the corresponding ground-truth segmentation mask, we count a hit; 
otherwise, we count a miss. Let $h_i \in \{0,1\}$ denote the hit indicator for case $i$ and 
let $N$ be the number of evaluated cases. The hit rate is then defined as
\[
    \text{hit rate} = \frac{1}{N} \sum_{i=1}^{N} h_i.
\]
}{For each available ground truth segmentation mask, the most salient point of the saliency method is calculated.
If this point lies within the ground truth segmentation mask, it is a hit; otherwise, it is not.
The proportion of hits is the hit rate.}\chadded{ Intuitively, the hit rate measures how often the single most highlighted point of a saliency map falls inside the expert-annotated pathology region, independent of the exact size or shape of that region.}

Additionally, we computed the mean Intersection over Union (mIoU), defined as the average of the intersection areas between predicted saliency maps and ground truth segmentations, divided by their union, calculated across all true positive, false positive, and false negative cases.\chadded{ In this context, a true positive (TP) case denotes an image–pathology pair with a positive ground-truth label and a non-empty predicted segmentation mask after thresholding; a false positive (FP) has a non-empty predicted mask but a negative ground-truth label; a false negative (FN) has a positive ground-truth label but no predicted mask; and a true negative (TN) has a negative ground-truth label and no predicted mask.}
This metric quantifies the spatial overlap quality of the saliency maps with respect to expert-annotated pathology regions.
\chadded{ In contrast to the hit rate, which only checks whether the most salient point lies inside the lesion, mIoU is sensitive to the full extent and shape of the highlighted area and penalizes overly diffuse or misplaced saliency.}
\chreplaced{For the saliency-based localization metrics (hit rate and mIoU), we used the official CheXlocalize evaluation code \cite{Saporta2022CheXlocalize} with its default bootstrapping configuration to compute 95\% confidence intervals.}{For bootstrapping, we used the authors' parameters.}

Unlike Saporta et al. \cite{Saporta2022CheXlocalize}, our mIoU evaluation included not only true positive cases—where both saliency and ground truth segmentation exist—but also evaluated false positive and false negative cases.
\chreplaced{To convert continuous saliency maps into binary segmentation masks, we tuned a single threshold per pathology on the CheXlocalize validation set by maximizing the mean Intersection over Union with respect to the radiologist-annotated ground-truth pixel-level masks provided as part of the CheXlocalize dataset.}{To determine the threshold for segmentation masks, we optimized it based on the validation set segmentation masks.}

\section{Results}

\begin{table}[ht]
\centering
\begin{tabular}{ l c c c c c c c c }
\toprule
 & \multicolumn{2}{c}{AUROC} & \multicolumn{2}{c}{Accuracy} & \multicolumn{2}{c}{Sensitivity} & \multicolumn{2}{c}{Specificity}\\
\cmidrule(lr){2-3} \cmidrule(lr){4-5} \cmidrule(lr){6-7} \cmidrule(lr){8-9}
 Model &  * & All &  * & All &  * & All &  * & All\\ 
 \midrule
 MedicalPatchNet & 0.907 & 0.902 & \textbf{0.836} & \textbf{0.848} & 0.825 & 0.763 & \textbf{0.841} & \textbf{0.851} \\  
 EfficientNet\chadded{V2-S} & \textbf{0.908} & \textbf{0.911} & 0.823 & 0.843 & \textbf{0.834} & \textbf{0.798} & 0.826 & 0.844 \\
 \midrule
 Difference & -0.001 & -0.009 & 0.013 & 0.005 & -0.009 & -0.035 & 0.015 & 0.007 \\
 \bottomrule
\end{tabular}
\caption{\label{tab:performance_comparison}Performance comparison between MedicalPatchNet and EfficientNet\chreplaced{V2-S}{-B0}. The metrics represent average values computed across the 10 classes from the CheXlocalize dataset (*) and across all 14 classes from the CheXpert dataset (All).}
\end{table}

Table \ref{tab:performance_comparison} presents the performance comparison between MedicalPatchNet and EfficientNet\chreplaced{V2-S}{-B0}.
It provides averaged metrics across all classes, as well as specifically across the 10 classes included in the CheXlocalize dataset.
Figure \ref{fig:auroc_comparison} displays the AUROC scores for each individual class.
\chadded{Detailed accuracy, sensitivity, and specificity values for all classes, as well as an additional backbone generalizability experiment comparing MedicalPatchNet--ConvNeXtBase to a standard ConvNeXt\cite{Liu2022ConvNext} classifier, are provided in the Supplementary Material.}\chdeleted{Detailed accuracy, sensitivity, and specificity values for all classes are provided in the Supplementary Material.}
To assess how effectively each saliency method identifies pathologies, we evaluated the hit rate, which is summarized in Table \ref{tab:hitrate_comparison}.\chreplaced{ For each pathology, this table reports the proportion of cases in which the single most salient point of a method’s map falls inside the corresponding expert-annotated segmentation mask (values between 0 and 1, higher is better).}{ For each pathology, this table reports the proportion of cases in which the single most salient point of a saliency map falls inside the corresponding expert-annotated segmentation mask (values between 0 and 1, higher is better), and, in the last column, the analogous human benchmark hit rates from Saporta et al.\cite{Saporta2022CheXlocalize}.}\chadded{ As expected, the radiologist benchmark substantially exceeds all automatic methods (mean hit rate 0.814 versus 0.471 and 0.485 for scaled and raw MedicalPatchNet encodings).}
MedicalPatchNet demonstrates superior performance in nine out of the ten analyzed classes.
Scaled patch \chreplaced{saliency maps}{encodings} outperform other methods in six classes, while raw patch \chreplaced{saliency maps}{encodings} perform best in three classes.
The average mean intersection over union (mIoU), computed across true positives \chdeleted{(TP)}, false positives \chdeleted{(FP)}, and false negatives \chdeleted{(FN)}, is presented in Table \ref{tab:iou_summary}; MedicalPatchNet shows the highest overall performance.
However, when evaluating mIoU exclusively for true positive cases, the best performance is achieved by Grad-CAM++.\chadded{ From a clinical perspective, the aggregate mIoU over true positives, false positives, and false negatives is particularly relevant, because it also penalizes saliency methods that highlight incorrect regions in false positive cases or fail to localize pathology in false negative cases, i.e. precisely in situations where reliable visual feedback is most critical.}

\begin{table}
\centering
\begin{tabular}{ l c c c c c c c c }
\toprule
 & \multicolumn{2}{c}{MedicalPatchNet} & \multicolumn{3}{c}{EfficientNet\chadded{V2-S}} \\
\cmidrule(lr){2-3} \cmidrule(lr){4-6}
Pathology & Scaled \chadded{Patch Saliency} & Raw \chadded{Patch Saliency} & Grad-CAM & Grad-CAM++ & Eigen-CAM \\
 \midrule
TP, FP, FN & \textbf{0.069} & \underline{0.056} & 0.052 & 0.054 & 0.049 \\
TP only & 0.168 & 0.164 & \underline{0.227} & \textbf{0.238} & 0.215 \\
 \bottomrule
\end{tabular}
\caption{\label{tab:iou_summary}Mean intersection over union (mIoU) averaged across true positives (TP), false positives (FP), and false negatives (FN), and reported separately for true positives (TP only).}
\end{table}
\begin{figure}
\centering
\includegraphics[width=0.9\linewidth]{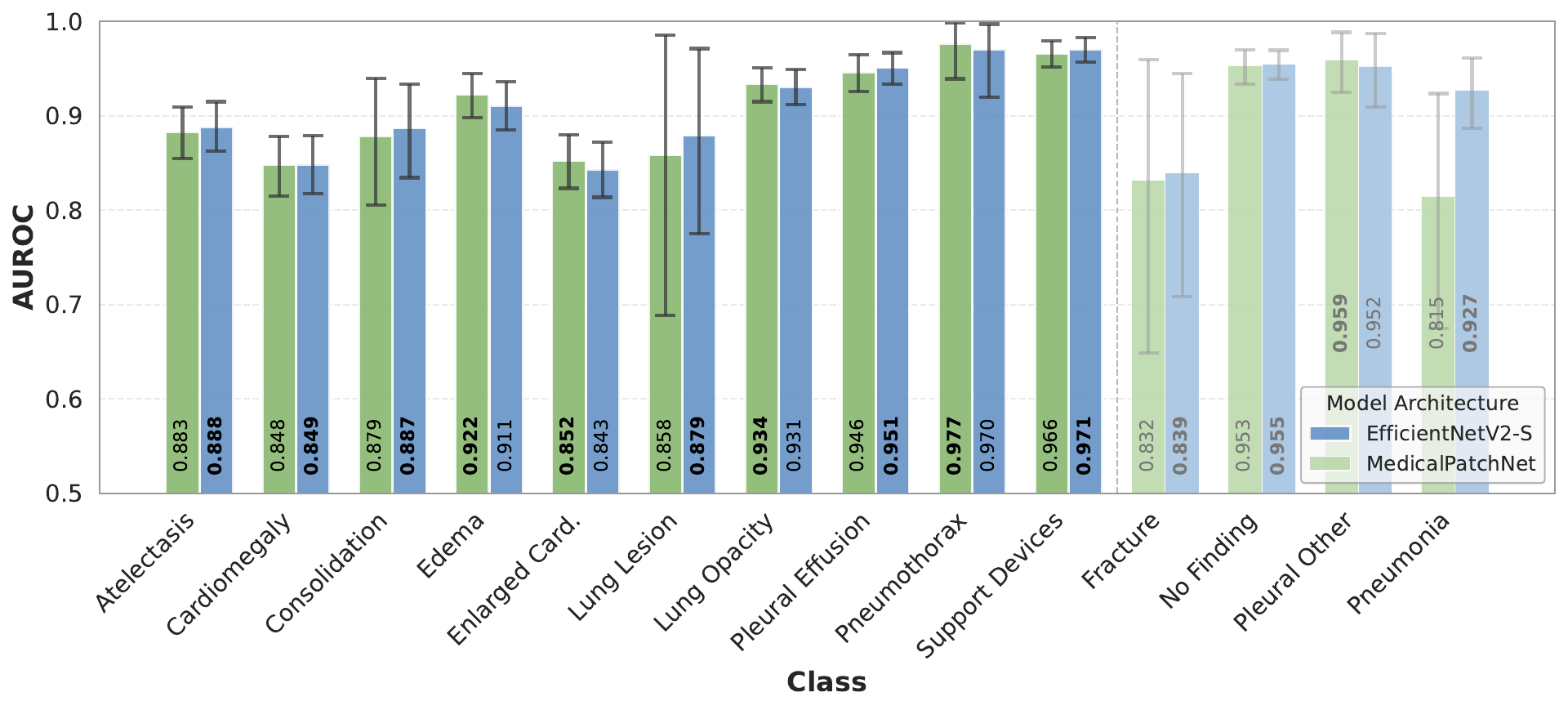}\caption{Comparison of the Area Under the Receiver Operating Characteristic (AUROC) curves for MedicalPatchNet and EfficientNet\chreplaced{V2-S}{-B0}, indicating similar classification performance.On the CheXlocalize dataset \cite{Saporta2022CheXlocalize}, both models yield mean AUROCs of 0.907 and 0.908, respectively; for the full 14-class CheXpert dataset \cite{Irvin2019Chexpert}, AUROCs are 0.902 and 0.911, respectively.}
\label{fig:auroc_comparison}
\end{figure}

\begin{table}[htbp!]
\centering
%\small
\resizebox{\textwidth}{!}{%
\begin{tabular}{llllll|l}
\toprule
 & \multicolumn{2}{c}{MedicalPatchNet} & \multicolumn{3}{c}{EfficientNet\chadded{V2-S}} \\
\cmidrule(lr){2-3} \cmidrule(lr){4-6}
Pathology & Scaled \chadded{Patch Saliency} & Raw \chadded{Patch Saliency} & Grad-CAM & Grad-CAM++ & Eigen-CAM & \chadded{Human} \\
\midrule
Lung Opacity & \textbf{0.493} \scriptsize{[0.438--0.548]} & 0.395 \scriptsize{[0.339--0.451]} & \underline{0.424} \scriptsize{[0.369--0.481]} & 0.414 \scriptsize{[0.359--0.470]} & 0.376 \scriptsize{[0.323--0.430]} & \chadded{0.559}\\
Atelectasis & \textbf{0.490} \scriptsize{[0.413--0.567]} & 0.389 \scriptsize{[0.315--0.463]} & \underline{0.406} \scriptsize{[0.335--0.483]} & 0.367 \scriptsize{[0.296--0.440]} & 0.373 \scriptsize{[0.301--0.451]} & \chadded{0.870}\\
Cardiomegaly & \textbf{0.464} \scriptsize{[0.390--0.535]} & 0.269 \scriptsize{[0.206--0.339]} & \underline{0.376} \scriptsize{[0.305--0.446]} & 0.246 \scriptsize{[0.184--0.311]} & 0.217 \scriptsize{[0.157--0.278]} & \chadded{0.972}\\
Consolidation & \textbf{0.540} \scriptsize{[0.381--0.706]} & 0.482 \scriptsize{[0.323--0.643]} & \underline{0.516} \scriptsize{[0.361--0.688]} & 0.401 \scriptsize{[0.250--0.568]} & 0.433 \scriptsize{[0.265--0.600]} & \chadded{0.510}\\
Edema & \underline{0.648} \scriptsize{[0.543--0.750]} & 0.601 \scriptsize{[0.488--0.707]} & \textbf{0.650} \scriptsize{[0.545--0.747]} & 0.471 \scriptsize{[0.367--0.573]} & 0.518 \scriptsize{[0.407--0.634]} & \chadded{0.769}\\
Enlarged Card. & \textbf{0.419} \scriptsize{[0.361--0.472]} & 0.320 \scriptsize{[0.266--0.374]} & \underline{0.392} \scriptsize{[0.340--0.448]} & 0.349 \scriptsize{[0.294--0.403]} & 0.322 \scriptsize{[0.267--0.374]} & \chadded{0.957}\\
Lung Lesion & 0.281 \scriptsize{[0.062--0.545]} & \textbf{0.564} \scriptsize{[0.267--0.824]} & 0.216 \scriptsize{[0.000--0.444]} & 0.284 \scriptsize{[0.071--0.538]} & \underline{0.354} \scriptsize{[0.100--0.615]} & \chadded{0.850}\\
Pleural Effusion & \textbf{0.466} \scriptsize{[0.378--0.552]} & \underline{0.372} \scriptsize{[0.280--0.458]} & 0.308 \scriptsize{[0.226--0.386]} & 0.293 \scriptsize{[0.213--0.375]} & 0.207 \scriptsize{[0.134--0.283]} & \chadded{0.718}\\
Pneumothorax & \underline{0.498} \scriptsize{[0.182--0.833]} & \textbf{0.794} \scriptsize{[0.500--1.000]} & 0.195 \scriptsize{[0.000--0.500]} & 0.195 \scriptsize{[0.000--0.500]} & 0.097 \scriptsize{[0.000--0.333]} & \chadded{1.000}\\
Support Devices & \underline{0.409} \scriptsize{[0.353--0.462]} & \textbf{0.668} \scriptsize{[0.614--0.717]} & 0.272 \scriptsize{[0.223--0.316]} & 0.225 \scriptsize{[0.179--0.268]} & 0.130 \scriptsize{[0.093--0.169]} & \chadded{0.933}\\
\midrule
Mean & \underline{0.471} & \textbf{0.485} & 0.376 & 0.325 & 0.303 & \chadded{0.814}\\
\bottomrule
\end{tabular}
}

\caption{\label{tab:hitrate_comparison}Hit rate of different saliency methods evaluated on the CheXlocalize dataset. Bold indicates the highest score, and underlining indicates the second-best score for each pathology. The 95\% confidence intervals are presented in brackets. The human benchmark values are obtained from a radiologist and are reported by Saporta et al. \cite{Saporta2022CheXlocalize}.}
\end{table}

\begin{figure}[!p]
    \centering
    \scriptsize
    \setlength{\tabcolsep}{2pt}
    \renewcommand{\arraystretch}{1}
    \begin{tabular}{@{}
        L{3.0cm}        % pathology / label column
        C{0.14\textwidth}
        C{0.14\textwidth}
        C{0.14\textwidth}
        C{0.14\textwidth}
        C{0.14\textwidth}@{}}
        \toprule
        & MedicalPatchNet (raw) (our) & Grad\,-CAM & Grad\,-CAM++ & Eigen\,-CAM & Ground Truth \\
        \midrule
        Pleural Effusion (True) &
        \includegraphics[width=\linewidth]{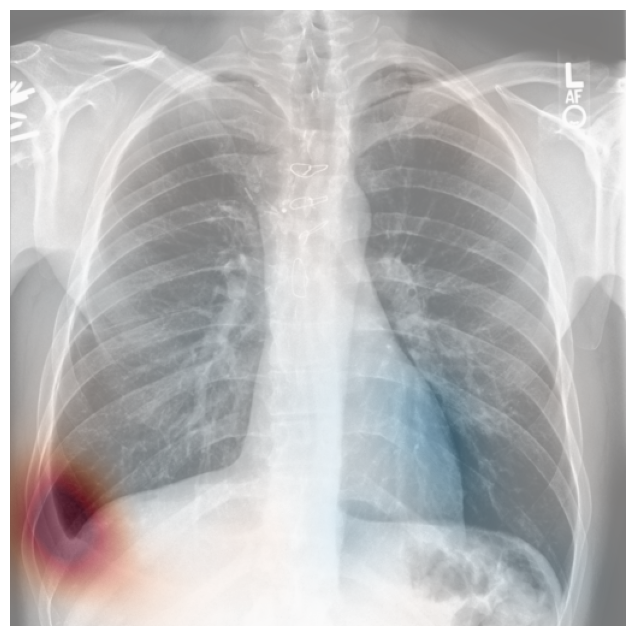} &
        \includegraphics[width=\linewidth]{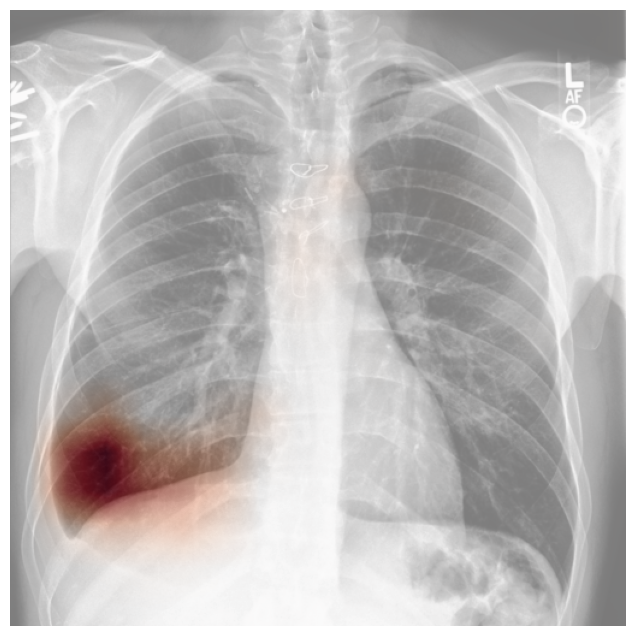} &
        \includegraphics[width=\linewidth]{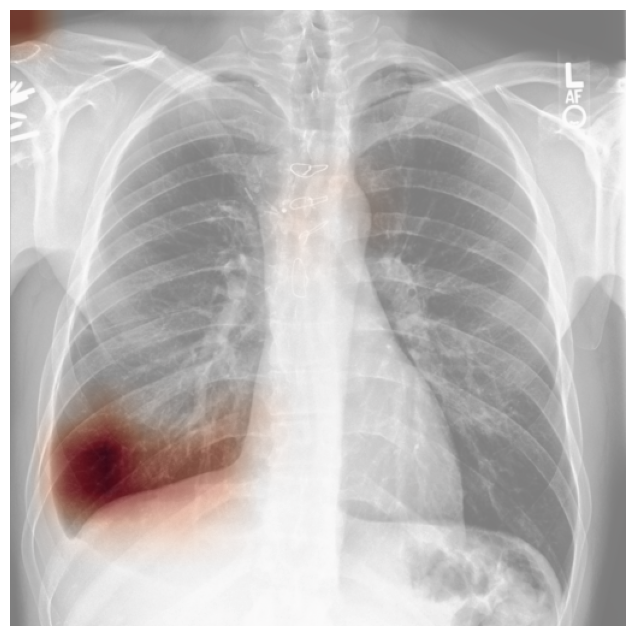} &
        \includegraphics[width=\linewidth]{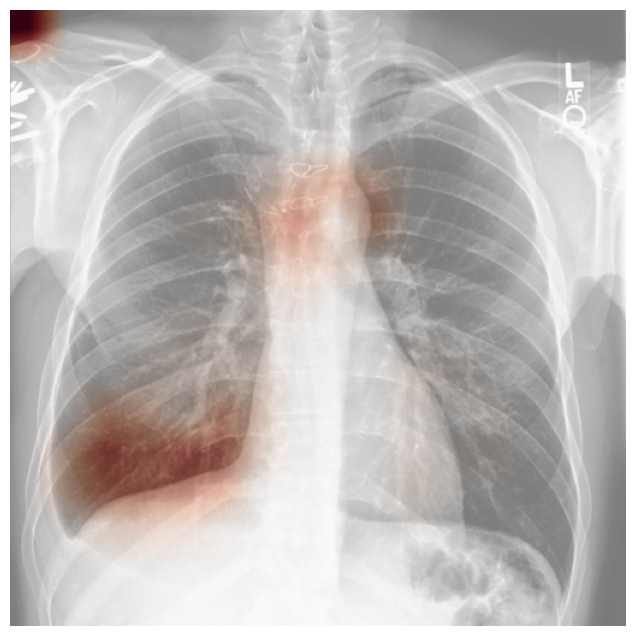} &
        \includegraphics[width=\linewidth]{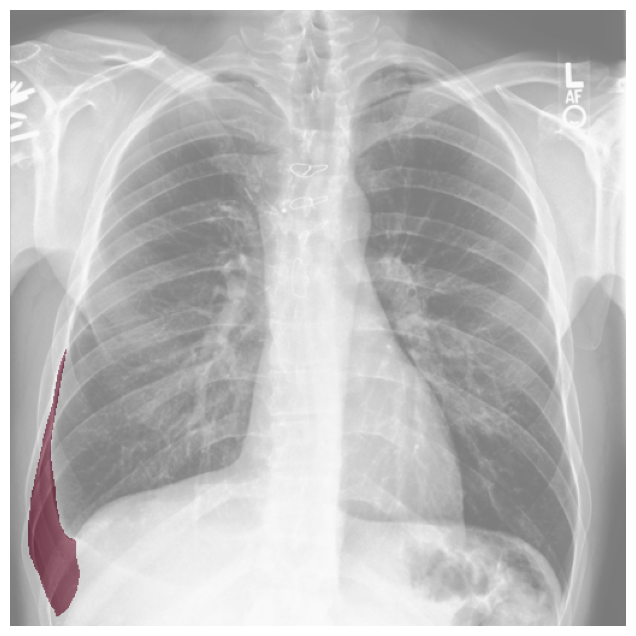} \\

        Pleural Effusion (False) &
        \includegraphics[width=\linewidth]{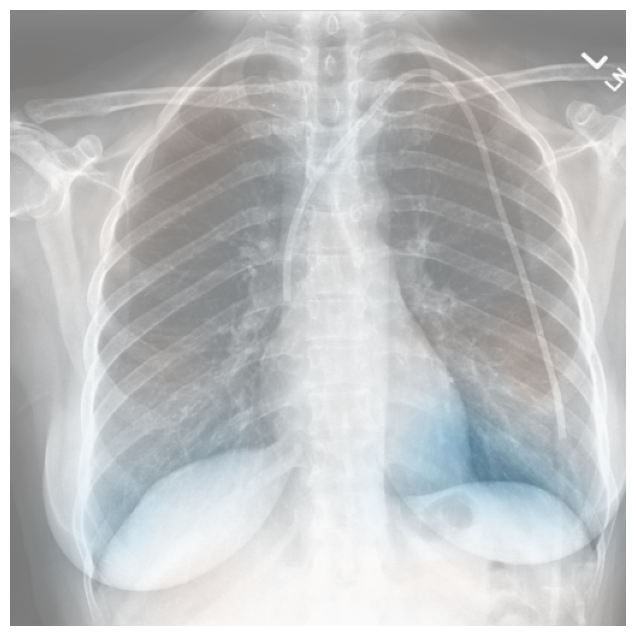} &
        \includegraphics[width=\linewidth]{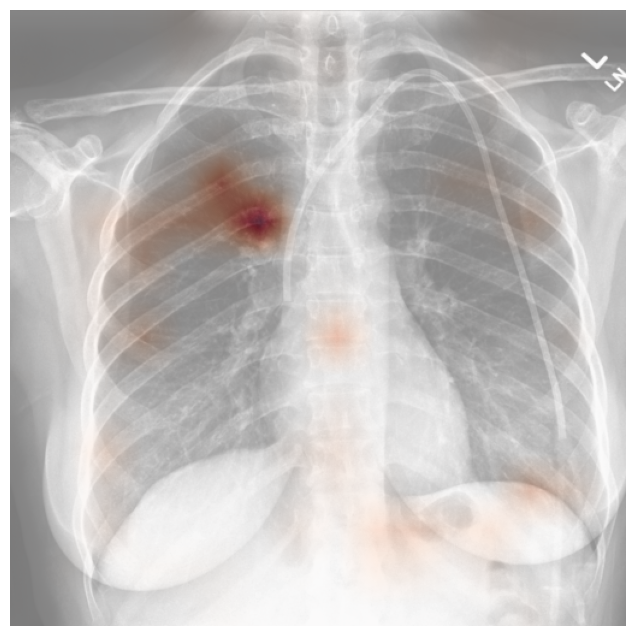} &
        \includegraphics[width=\linewidth]{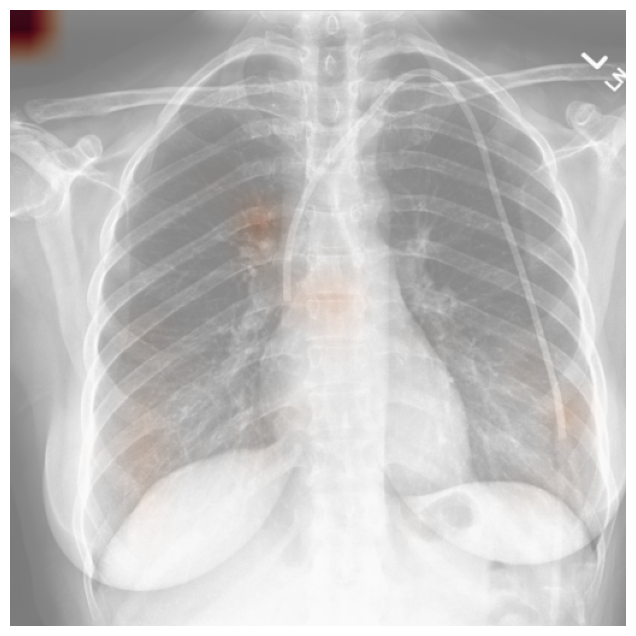} &
        \includegraphics[width=\linewidth]{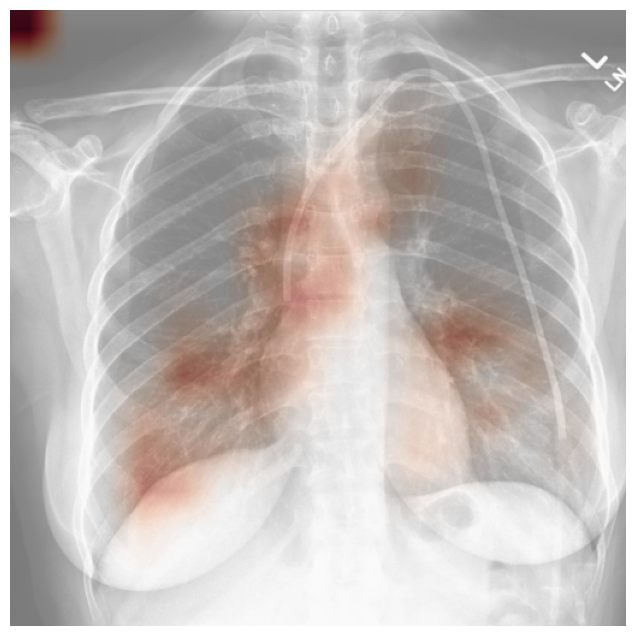} &
        No ground truth mask given for ``Pleural Effusion'' \\

        Support Devices (True) &
        \includegraphics[width=\linewidth]{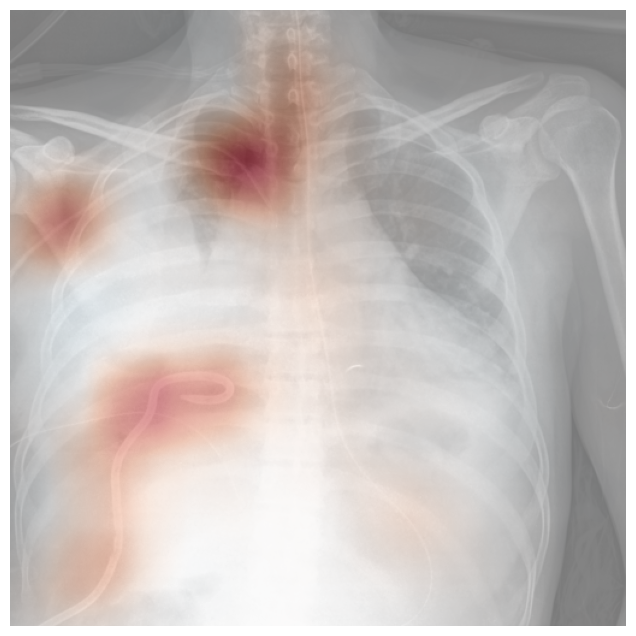} &
        \includegraphics[width=\linewidth]{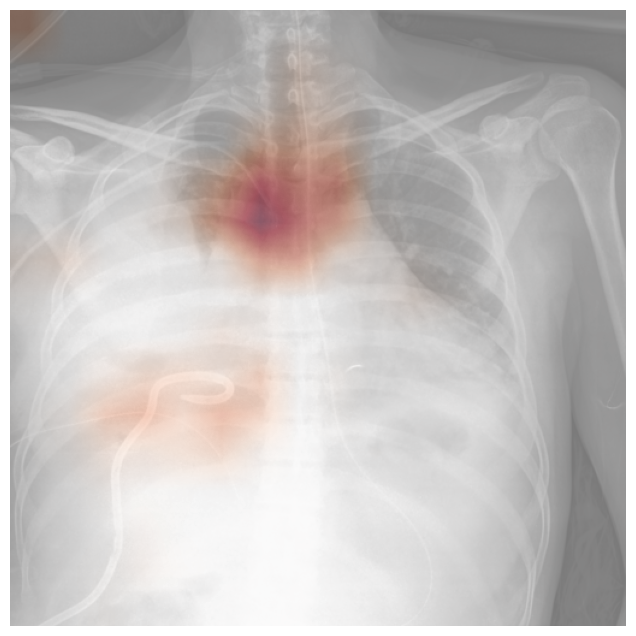} &
        \includegraphics[width=\linewidth]{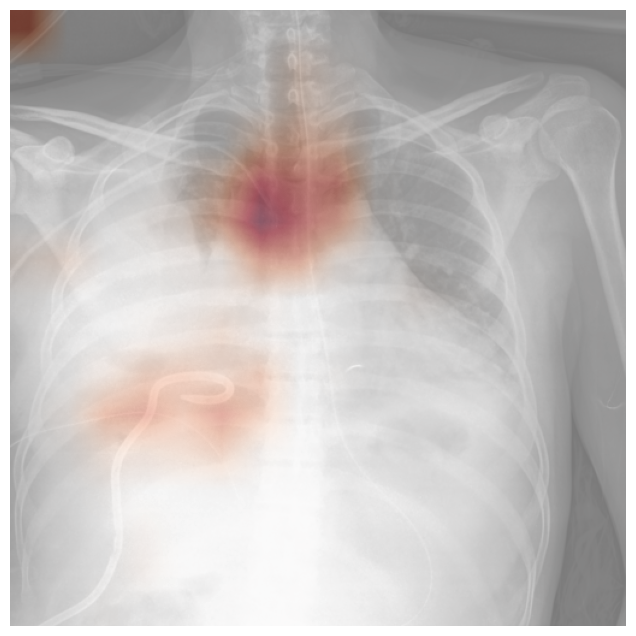} &
        \includegraphics[width=\linewidth]{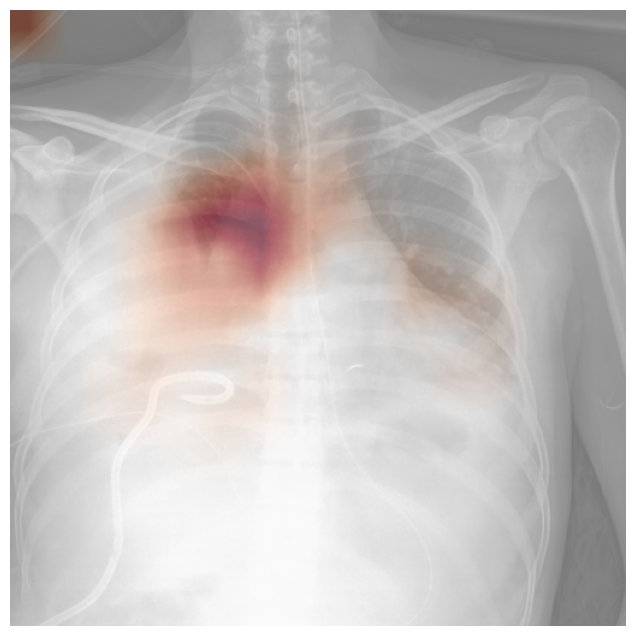} &
        \includegraphics[width=\linewidth]{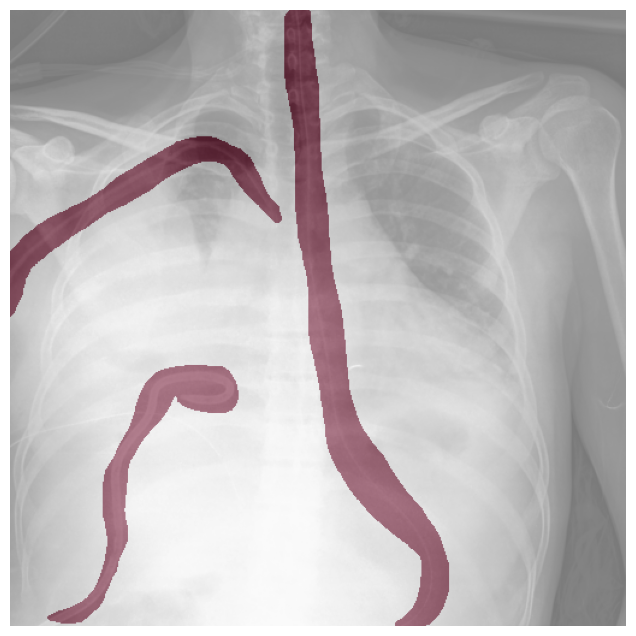} \\

        No Finding (False) &
        \includegraphics[width=\linewidth]{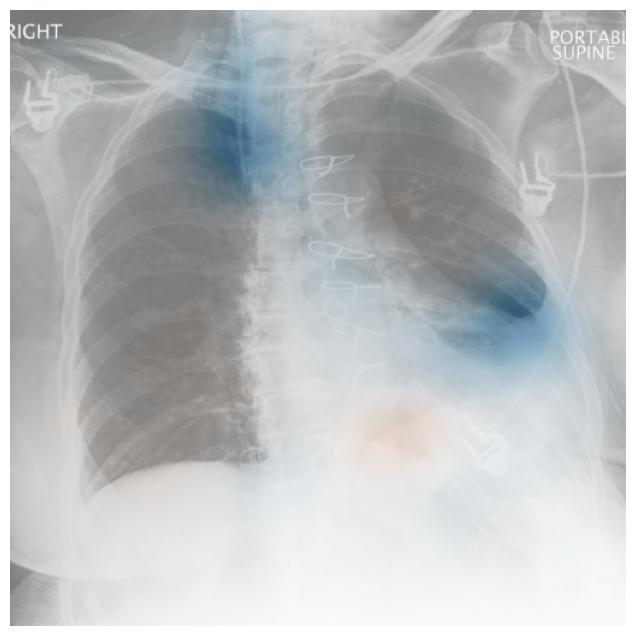} &
        \includegraphics[width=\linewidth]{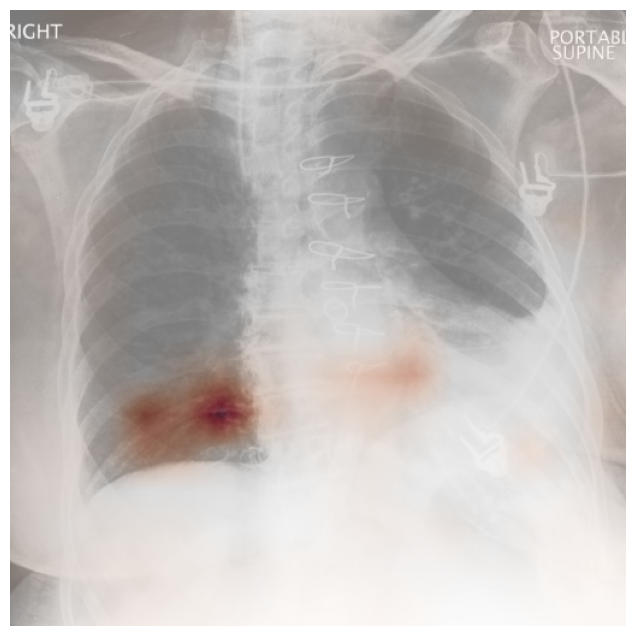} &
        \includegraphics[width=\linewidth]{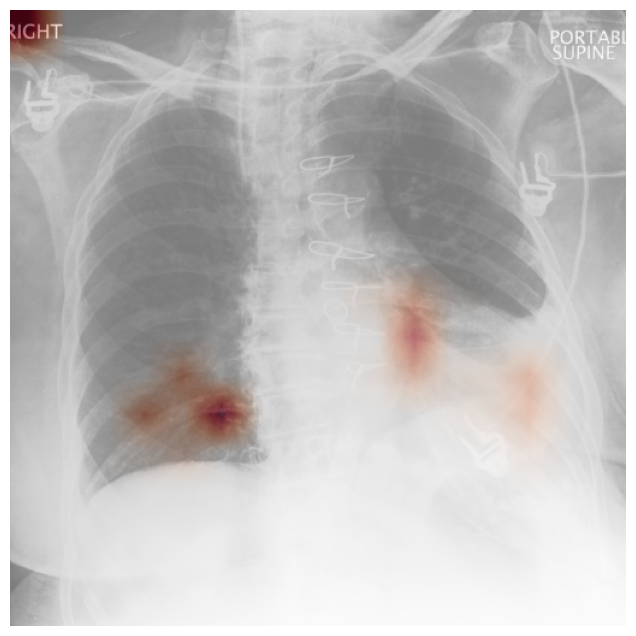} &
        \includegraphics[width=\linewidth]{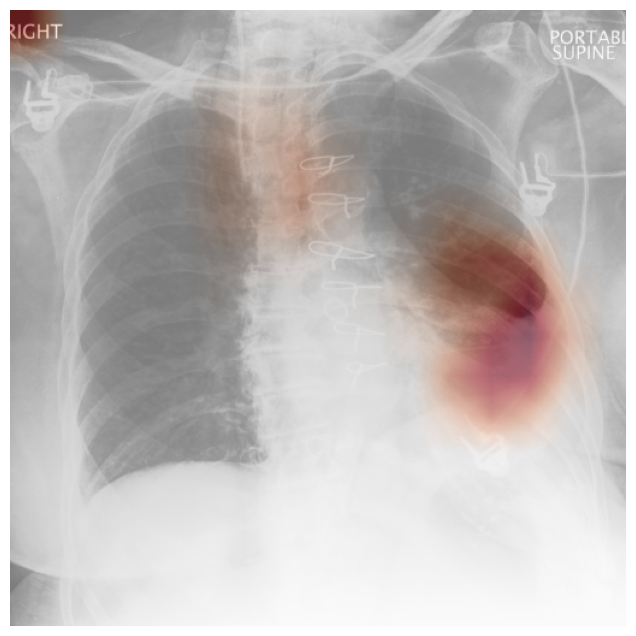} &
        No ground truth mask given for ``No Finding'' \\

        Pneumothorax (False) &
        \includegraphics[width=\linewidth]{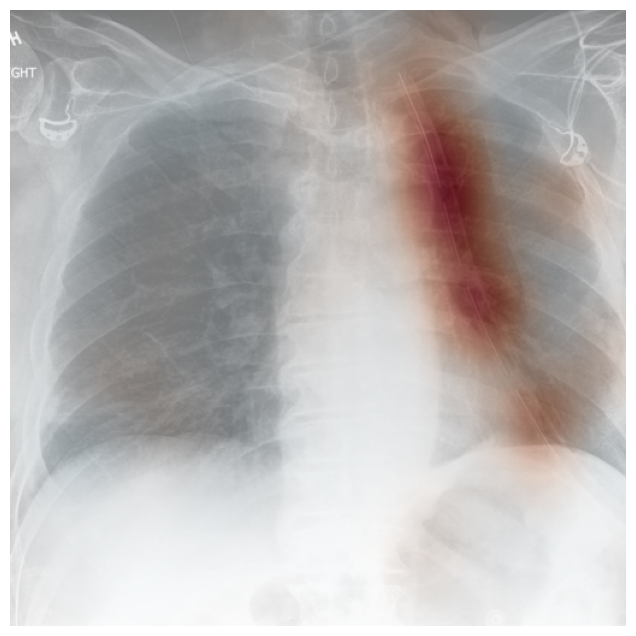} &
        \includegraphics[width=\linewidth]{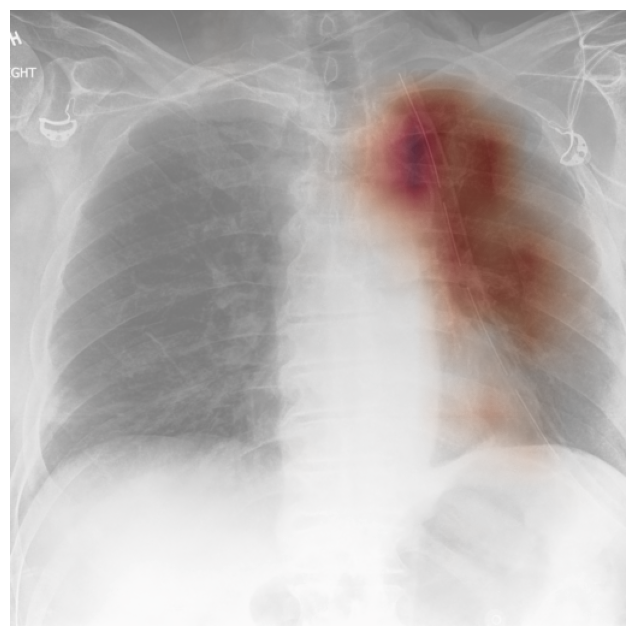} &
        \includegraphics[width=\linewidth]{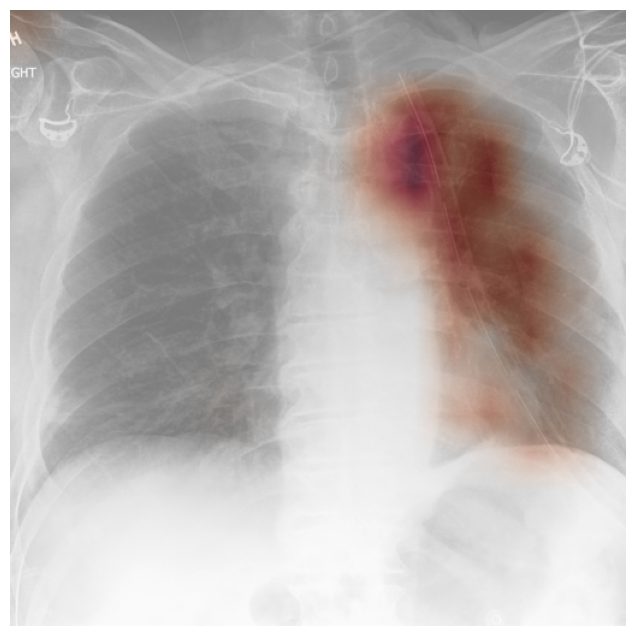} &
        \includegraphics[width=\linewidth]{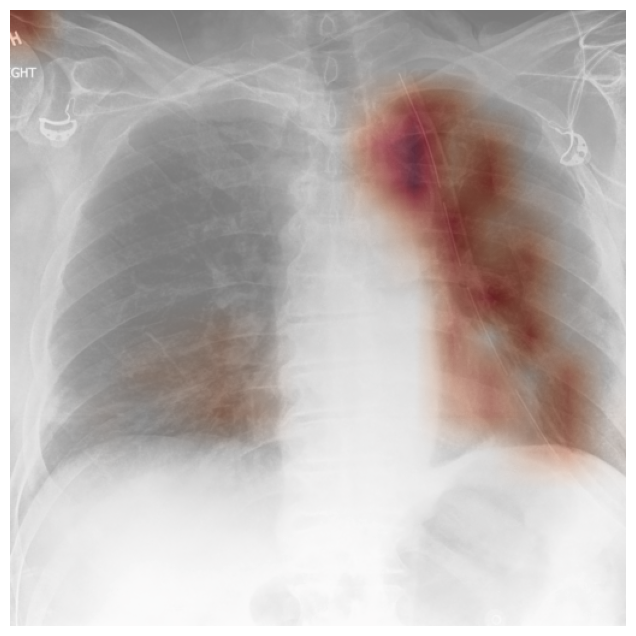} &
        No ground truth mask given for ``Pneumothorax'' \\

        Support Devices (True) &
        \includegraphics[width=\linewidth]{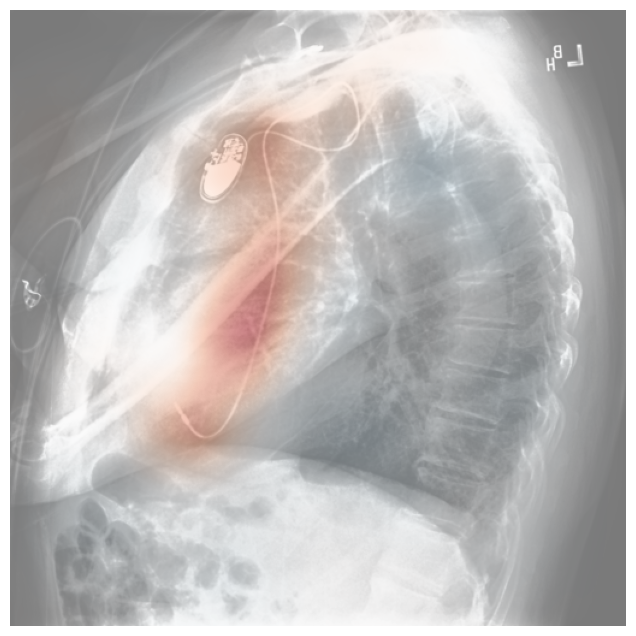} &
        \includegraphics[width=\linewidth]{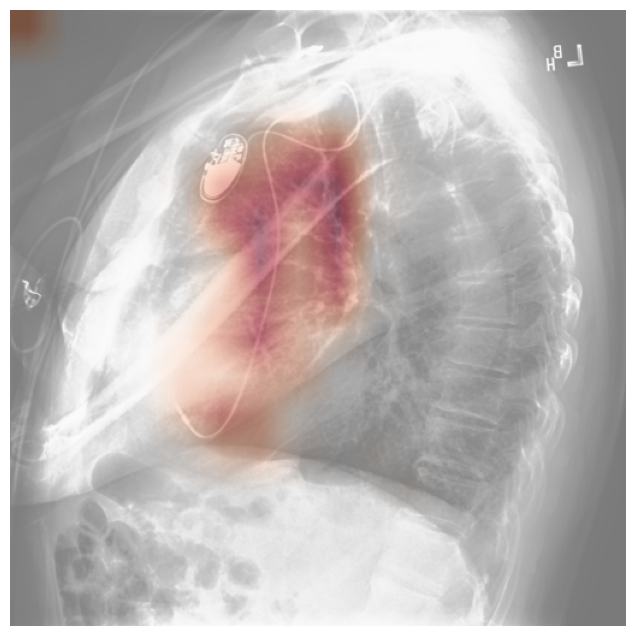} &
        \includegraphics[width=\linewidth]{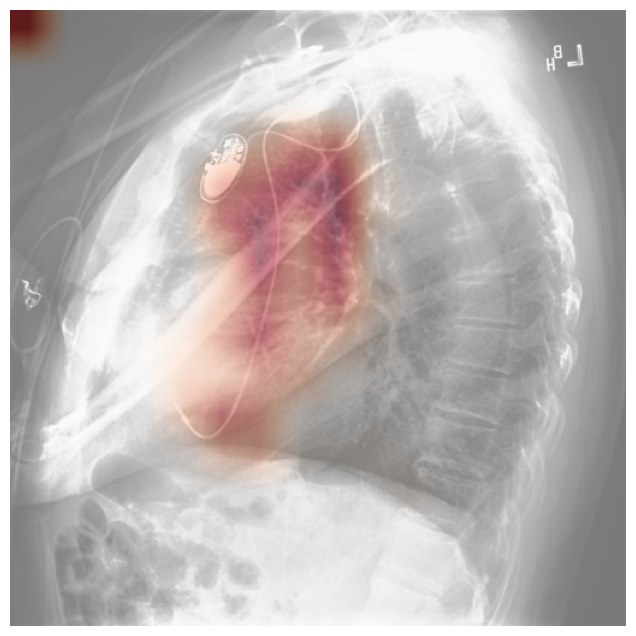} &
        \includegraphics[width=\linewidth]{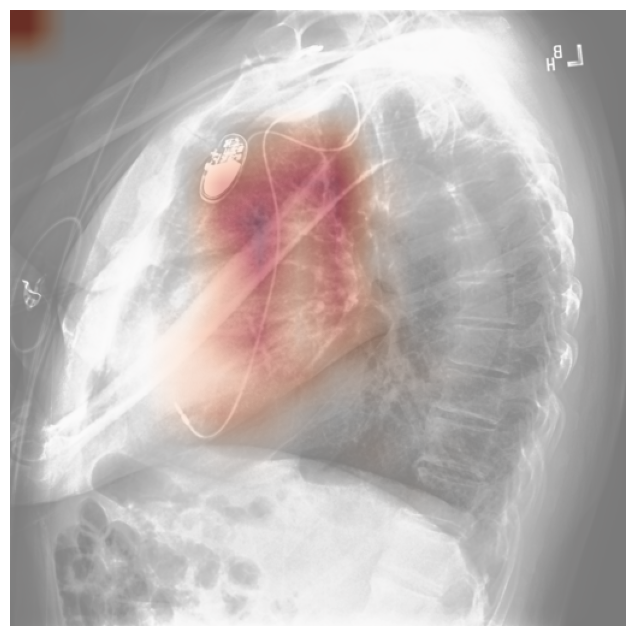} &
        \includegraphics[width=\linewidth]{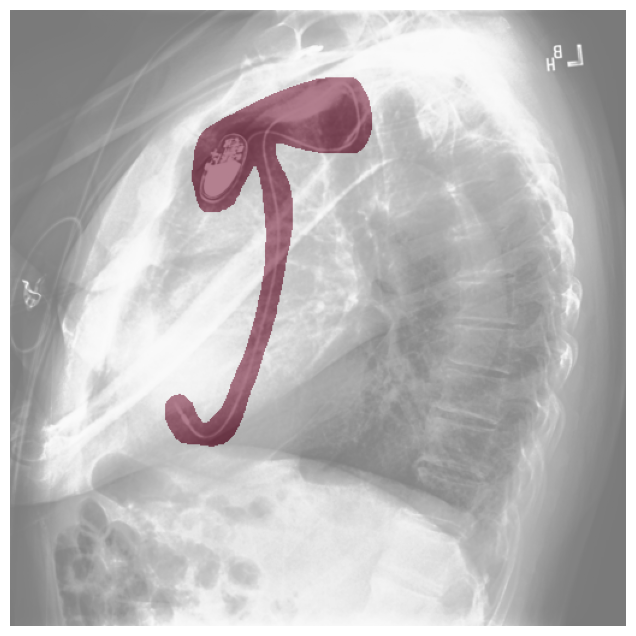} \\
        \bottomrule
    \end{tabular}
    \caption{Representative saliency maps produced by MedicalPatchNet and three post-hoc methods. Each row displays the same chest X-ray for a given pathology together with its ground-truth label (``True'' or ``False''). Columns compare MedicalPatchNet’s raw patch logits with Grad-CAM, Grad-CAM++, and Eigen-CAM applied to an EfficientNet\chreplaced{V2-S}{-B0} baseline. In the MedicalPatchNet maps, \textcolor[HTML]{9E1127}{red} denotes evidence supporting the class and \textcolor[HTML]{0F457E}{blue} denotes evidence against it, whereas Grad-CAM–based maps visualize only positive (red) contributions. Eigen-CAM is class-agnostic and therefore does not generate class-specific saliency maps. Interestingly, for the wrongly diagnosed pneumothorax, all four explainability methods point to the chest tube, revealing that the model used a shortcut, with MedicalPatchnet denoting its course most clearly.}
    \label{tab:activationMaps}
\end{figure}

\section{Discussion}

The results presented in Table \ref{tab:performance_comparison} demonstrate that MedicalPatchNet, using an image size of $512 \times 512$ and a patch size of $64 \times 64$, achieves classification performance comparable to EfficientNet\chreplaced{V2-S}{-B0}, which utilizes the full $512 \times 512$ context.
The results show that the classification performance of MedicalPatchNet matches that of EfficientNet\chreplaced{V2-S}{-B0}, except for pneumonia.
This indicates that classification of these chest X-ray pathologies can be achieved with only local information.
It is not necessary for the different patches and their processing to communicate until the logits are averaged at the end.\chadded{ Importantly, the aim of MedicalPatchNet is not to surpass the original EfficientNet\chreplaced{V2-S}{-B0} in raw classification accuracy, but to retain comparable performance under the additional structural constraint that all evidence is decomposed into local, independently processed patches. In this sense, matching the backbone’s AUROC demonstrates that a substantially more constrained and interpretable aggregation scheme can be imposed without incurring a large loss in diagnostic performance.}
\chadded{A complementary experiment on natural images (PASCAL VOC 2012, Supplementary Material, section ``Evaluation on natural images'') shows that the AUROC gap between MedicalPatchNet and EfficientNet\chreplaced{V2-S}{-B0} is slightly larger on complex multi-object scenes, which is consistent with the intuition that tasks requiring stronger global context benefit more from a fully image-level architecture.}
However, it should be noted that this observation is limited to the pathologies examined here, as one can imagine cases solvable by EfficientNet\chreplaced{V2-S}{-B0} but not by MedicalPatchNet, as discussed in the limitations.
\chadded{We further examined the impact of patch size on classification performance by training MedicalPatchNet with patch sizes of $64\times64$, $128\times128$, and $256\times256$, and comparing it to an image-level EfficientNet\chreplaced{V2-S}{-B0} (effectively a single $512\times512$ patch). On average across all pathologies, the mean AUROC varied only marginally between these configurations (approximately 0.900–0.911), although for some individual findings, such as pneumonia, larger patches yielded noticeably higher AUROC than smaller ones. Detailed per-class results are reported in the Supplementary Material.}

Having reliable explainability methods is essential for the clinical application of neural networks, as these provide insights into their decision processes.
Especially when they make decisions that contradict human reasoning or appear unintuitive at first glance.
For example, if the network learns to classify pathologies based on features that correlate in the dataset but are unrelated in reality, a reliable explainability method allows investigation of such behavior.
DeGrave et al. \cite{DeGrave2021Shortcut} demonstrated that some networks trained for COVID-19 classification rely on shortcuts—such as laterality markers or radiopacity at image borders—rather than on genuine pathological features.
A similar observation was made by Zech et al. \cite{Zech2018VariableGeneralizationPerformance} who discovered that pneumonia classification sometimes relied on laterality markers instead of lung features.
This demonstrates the value of explainability methods.
It also highlights the different goals of explainability methods and weakly supervised segmentation.
Their goals are fundamentally different: weakly supervised segmentation aims to segment the pathology as precisely as possible, whereas explainability methods aim to reflect the neural network’s decision-making process as precisely as possible.
Ideally, these two should align if the network does not rely on shortcuts—but this is not necessarily true, as shown \cite{DeGrave2021Shortcut,Zech2018VariableGeneralizationPerformance}.
Consequently, an evaluation based solely on the ability of a saliency method to localize a pathology may unintentionally favor a method that does not truly explain the network but highlights, for example, random spots in a specific body region over one that highlights the true decision rationale.
\chadded{ Moreover, all localization metrics in this work are reported at the level of individual pathologies and aggregate lesions across the entire thorax; the available datasets do not contain standardized labels for apical, hilar, central, or basal regions, so we cannot stratify localization performance by anatomical subregion.}

Consider the following thought experiment:
Imagine a neural network that achieves high accuracy in fracture detection on a hypothetical dataset.
In this dataset, images with fractures consistently display a laterality marker ``R,'' whereas those without fractures have a marker ``L.''
Suppose the trained network relies entirely on these laterality markers for classification.
Now, consider two explainability methods:
Method $A$ accurately highlights the true basis for classification.
Method $B$, however, randomly highlights bone regions regardless of fracture presence.
If evaluated solely by how often they correctly localize or identify the fracture, method $B$ might appear superior to method $A$, as it occasionally overlaps with fractures by chance, whereas method $A$ consistently highlights only laterality markers and never directly indicates fractures.

This underscores that evaluations relying solely on segmentation accuracy or hit rate are inadequate for comparing explainability methods.\chadded{ A concrete example is cardiomegaly: human annotators often contour or point to the entire enlarged cardiac silhouette, which is clinically natural, whereas saliency methods typically highlight only those edges of the heart or mediastinum that are most discriminative for the model. Both behaviours are reasonable from their respective perspectives, but they yield different spatial patterns and can therefore create apparent performance gaps between human and model explanations when compared against dense radiologist segmentations.}
As described by the authors of the CheXlocalize paper, they evaluate how well post-hoc saliency methods perform on localization tasks \cite{Saporta2022CheXlocalize}.
This, however, does not necessarily align with the goal of explaining the neural network's decision as precisely as possible.
The same holds true for evaluating saliency methods based on the visual assessment of saliency maps by human experts.
Just because a saliency map appears to align with a human expert’s reasoning does not necessarily mean the network relied on the same visual features for classification.
For a radiologist in the thought experiment, examining bones like method $B$ would seem more intuitive when searching for a fracture than focusing on laterality markers like method $A$.

An effective explainability method must inherently reflect only the true underlying reasons behind the network’s classification decisions by design. \chreplaced{In MedicalPatchNet, a key goal was to keep the model a drop-in replacement for a standard image-level classifier: it processes full images, relies solely on image-level labels, and does not use explicit masks or segmentation annotations to select regions. Interpretability is then achieved by restricting raw patch encodings to purely local information.}{ Our approach ensures this by exclusively using local information for raw patch encodings.}
\chreplaced{Including all patches, even those at the image margins, is a deliberate design choice: it keeps the architecture fully annotation-free and allows the resulting saliency maps to expose potential shortcut features in peripheral regions (such as markers or border artifacts), rather than suppressing them a priori through manual masking.}{If a specific patch indicates the presence of a class, it implies the network's decision relies solely on the content of that patch—not on distant features or on the combination of features across different patches.}
There is no communication between the information from different patches until the final averaging step.
Therefore, if the network utilizes information from a patch for classification, the logits corresponding to that patch must directly reflect this.\chadded{ Combined with the drop-in design relative to the underlying backbone, this yields a model that can be used in place of a standard image-level classifier while making all contributing image regions explicitly visible.}
There is no alternative mechanism that allows such information to influence the final prediction unless it is explicitly represented in the generated saliency map.

Although several Multiple Instance Learning approaches \cite{Ilse18MIL,Shao21TransMIL,lu2021data} similarly process images at the patch level and provide interpretability through learned attention mechanisms or transformer-based correlations, these methods have been primarily applied in histopathology. In contrast, MedicalPatchNet is specifically designed for chest X-ray analysis and deliberately avoids learnable pooling or inter-patch interactions. By averaging independently predicted patch logits, our method ensures that each patch’s influence on the final decision is direct and transparent, offering interpretability by design rather than relying on learned attribution weights.

\subsection{Limitations}
Using our architecture comes with limitations.
First, the reliance on only local information does not allow for classifications that depend on broader context, where information from spatially distant image regions must be aggregated—beyond the patch size.
For example, pulmonary congestion is often associated both with an enlarged heart and with bilateral infiltrates. 
Such a condition requires integrating information across distant regions of the image, which cannot be captured by our patch-wise averaging strategy.

\chadded{Second, MedicalPatchNet is primarily sensitive to local shortcut features that are confined to specific regions of the image (e.g.\ laterality markers, text overlays, medical devices, border artifacts). Dataset biases that act globally---such as vendor-specific post-processing or characteristic global contrast and texture patterns---will usually affect all patches similarly, so our saliency maps tend to appear relatively homogeneous and cannot meaningfully isolate the source of such global bias.}

\chadded{Third, the explainability we provide is restricted to spatial attribution at the patch level. Our saliency maps indicate where evidence for a class arises, but not which fine-grained visual concept within a patch is responsible; the internal feature representations of each patch-wise classifier therefore remain a black box and may require complementary concept-level analyses in future work.}

\chadded{Fourth, our quantitative localization analysis does not differentiate between anatomical subregions. Lesions in apical, perihilar, central, or basal lung regions are evaluated jointly for each pathology, because the available datasets do not provide structured region labels. As a result, potential differences in localization behaviour across specific anatomical areas are not captured by our current evaluation and should be investigated in future work once suitable region-level annotations become available.}

\subsection{Conclusion}
This study presents the new inherently explainable MedicalPatchNet architecture, which enables backtracking and decomposition of the exact contribution of each patch to the final classification.
Unlike traditional post-hoc explainability methods that can produce misleading visualizations, MedicalPatchNet incorporates explainability directly into the model's core architecture.
The resulting patch-based saliency maps transparently reflect the exact contributions of individual image regions to each diagnostic decision.
Empirical evaluations on the CheXpert and CheXlocalize datasets show that MedicalPatchNet not only achieves strong classification performance but also \chreplaced{outperforms}{substantially outperforms} existing methods such as Grad-CAM and its variants in pathology localization accuracy\chadded{ when evaluated across true positive, false positive, and false negative cases}.
This demonstrates the feasibility of maintaining high diagnostic precision without sacrificing interpretability.
MedicalPatchNet thus addresses a critical barrier to clinical AI adoption by delivering inherently reliable explanations that are accessible even to practitioners without extensive deep learning expertise.
Future work could adapt this explainability approach to other settings, such as 3D imaging or multimodal classification, enabling quantification of the contribution of different modalities\chadded{. In particular, the same patch-based architecture could be applied to volumetric patches or slabs in CT or MRI, or extended to multi-disease diagnostic tasks beyond chest X-rays, providing analogous self-explainable decompositions in more complex imaging scenarios.}
When applied to neural networks used in clinical settings, this approach may enhance trust, ultimately facilitating safer, more effective patient care through transparent AI-assisted diagnostics.

\bibliography{bibliography}

\noindent\textbf{Acknowledgements}\\
The authors gratefully acknowledge the computing time provided to them at the NHR Center NHR4CES at RWTH Aachen University (project number p0021834). This is funded by the Federal Ministry of Education and Research, and the state governments participating on the basis of the resolutions of the GWK for national high performance computing at universities (www.nhr-verein.de/unsere-partner).
The data used in this publication was managed using the research data management platform Coscine (http://doi.org/10.17616/R31NJNJZ) with storage space of the Research Data Storage (RDS) (DFG: INST222/1261-1) and DataStorage.nrw (DFG: INST222/1530-1) granted by the DFG and Ministry of Culture and Science of the State of North Rhine-Westphalia.
We used generative AI tools for language editing and rephrasing; all scientific content, data, analyses, and conclusions were written and verified by the authors.\\

\noindent\textbf{Funding}\\
This research is supported by the Deutsche Forschungsgemeinschaft - DFG (NE 2136/7-1, NE 2136/3-1, TR 1700/7-1) , the German Federal Ministry of Research, Technology and Space (Transform Liver - 031L0312C, DECIPHER-M, 01KD2420B) and the European Union Research and Innovation Programme (ODELIA - GA 101057091).\\

\noindent\textbf{Author contributions}\\
P.W. conceptualized the study, developed the software, conducted the experiments, and wrote the manuscript. D.T. conceptualized the study, provided supervision, and reviewed the manuscript. C.K., J.N.K., and S.N. provided supervision and contributed to reviewing the manuscript. All authors read and approved the final manuscript.\\

\noindent\textbf{Competing interests}\\
D.T. received honoraria for lectures by Bayer, GE, Roche, Astra Zenica, and Philips and holds shares in StratifAI GmbH, Germany and in Synagen GmbH, Germany.
J.N.K. declares consulting services for Panakeia, AstraZeneca, MultiplexDx, Mindpeak, Owkin, DoMore Diagnostics, and Bioptimus. Furthermore, he holds shares in StratifAI, Synagen, Tremont AI, and Ignition Labs, has received an institutional research grant from GSK, and has received honoraria from AstraZeneca, Bayer, Daiichi Sankyo, Eisai, Janssen, Merck, MSD, BMS, Roche, Pfizer, and Fresenius.\\

\noindent\textbf{Data availability statement}
The CheXpert dataset\cite{Irvin2019Chexpert} used for training is publicly available from the Stanford ML Group at \href{https://stanfordmlgroup.github.io/competitions/chexpert/}{https://stanfordmlgroup.github.io/competitions/chexpert/}. The CheXlocalize dataset\cite{Saporta2022CheXlocalize}, used for evaluation, is publicly available from the Stanford AIMI group at \href{https://stanfordaimi.azurewebsites.net/datasets/23c56a0d-15de-405b-87c8-99c30138950c}{https://stanfordaimi.azurewebsites.net/datasets/23c56a0d-15de-405b-87c8-99c30138950c}. The source code for our model, training, and evaluation, are publicly available on GitHub at \href{https://github.com/TruhnLab/MedicalPatchNet}{github.com/TruhnLab/MedicalPatchNet}.
The model weights are publicly available on Hugging Face: \href{https://huggingface.co/patrick-w/MedicalPatchNet}{https://huggingface.co/patrick-w/MedicalPatchNet}\\

\renewcommand{\thefigure}{S\arabic{figure}}
\renewcommand{\thetable}{S\arabic{table}}
\setcounter{section}{0}
\setcounter{figure}{0}
\setcounter{table}{0}

\flushbottom
%\maketitle
{\raggedright\sffamily\bfseries\LARGE
Supplementary Material for \\MedicalPatchNet: A Patch-Based Self-Explainable AI Architecture for Chest X-ray Classification\par}
%\vspace{50pt}
\vspace{50pt}

\section{Mask Size Evaluation}
\chadded{To assess how lesion size affects localization performance, we stratified the CheXlocalize test set by ground-truth mask area and evaluated MedicalPatchNet separately for smaller and larger lesions. Specifically, for each pathology we split cases into those with a ground-truth segmentation mask area greater than the median and those with an area less than or equal to the median. Across all pathologies with available segmentations, the median fraction of image area covered by a ground-truth mask is 9.92\%.}

\begin{table}[H]
\centering
\begin{tabular}{lcccc}
\toprule
 & \multicolumn{2}{c}{Hit rate} & \multicolumn{2}{c}{mIoU} \\
\cmidrule(lr){2-3}\cmidrule(lr){4-5}
Pathology & $>$Median & $\leq$Median & $>$Median & $\leq$Median \\
\midrule
Lung Opacity               & 0.698 & 0.404 & 0.144 & 0.032 \\
Atelectasis                & 0.746 & 0.364 & 0.133 & 0.017 \\
Cardiomegaly               & 0.475 & 0.308 & 0.143 & 0.003 \\
Consolidation              & 0.769 & 0.409 & 0.261 & 0.030 \\
Edema                      & 0.681 & 0.500 & 0.276 & 0.022 \\
Enlarged Cardiomediastinum & 0.423 & 0.167 & 0.177 & 0.002 \\
Lung Lesion                & 1.000 & 0.231 & 0.206 & 0.005 \\
Pleural Effusion           & 0.788 & 0.345 & 0.324 & 0.075 \\
Pneumothorax               & 0.500 & 0.500 & 0.075 & 0.080 \\
Support Devices            & 0.561 & 0.385 & 0.221 & 0.160 \\
\bottomrule
\end{tabular}
\caption{Hit rate and mean Intersection over Union (mIoU) stratified by mask size. For each pathology, metrics are reported separately for lesions with ground-truth mask area greater than (``$>$Median'') or less than or equal to (``$\leq$Median'') the median lesion size.}
\label{tab:size_stratified_metrics}
\end{table}

\chadded{When comparing the values in Table~\ref{tab:size_stratified_metrics}, a consistent pattern emerges: for almost all pathologies, hit rate and mIoU are markedly lower for lesions with ground-truth masks at or below the median size than for larger lesions. We attribute this effect to the fact that small lesions occupy only a few patches or pixels, making it more difficult for the most salient point to fall inside the ground-truth region and causing the IoU metric to penalize even minor spatial misalignment much more strongly than for larger, more extended pathologies.}

\section{Patch Size Analysis}

\begin{figure}
  \centering
  %\includesvg[width=\textwidth]{figsupplement/auroc_barplot_with_ci.svg}
  \includegraphics[width=\textwidth]{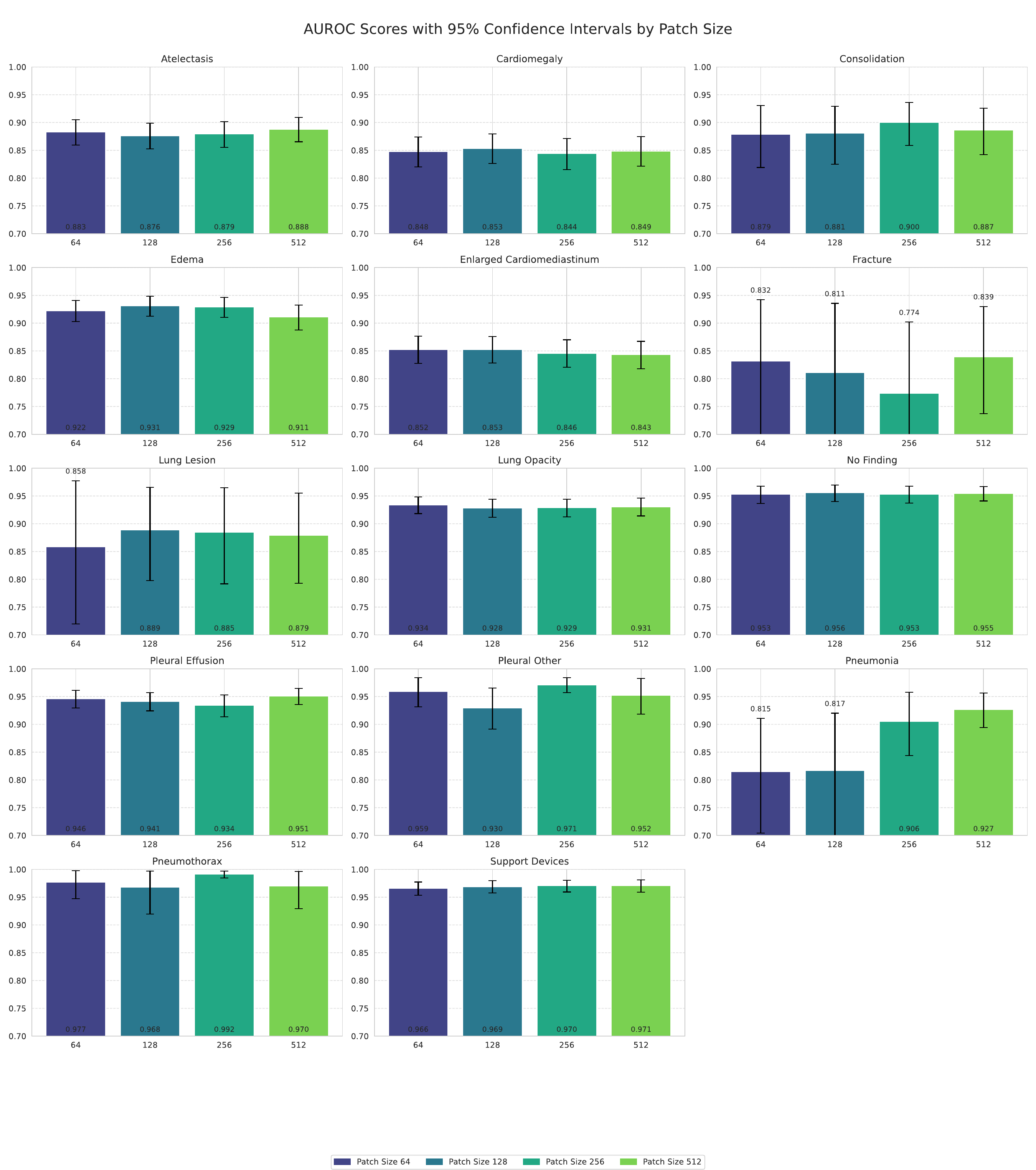}
  \caption{\chadded{Classification performance (AUROC) of MedicalPatchNet for different patch sizes across all CheXpert pathologies. For each class, bars show the AUROC obtained with patch sizes of $64\times64$, $128\times128$, and $256\times256$ pixels, as well as the baseline EfficientNetV2-S model, which corresponds to processing the full $512\times512$ image as a single patch. Error bars indicate 95\% confidence intervals.}}
  \label{fig:patchSizeMultiDisease}
\end{figure}

\begin{figure}
  \centering
  %\includesvg[width=0.5\textwidth]{figsupplement/avg_performance_summary.svg}
  \includegraphics[width=0.5\textwidth]{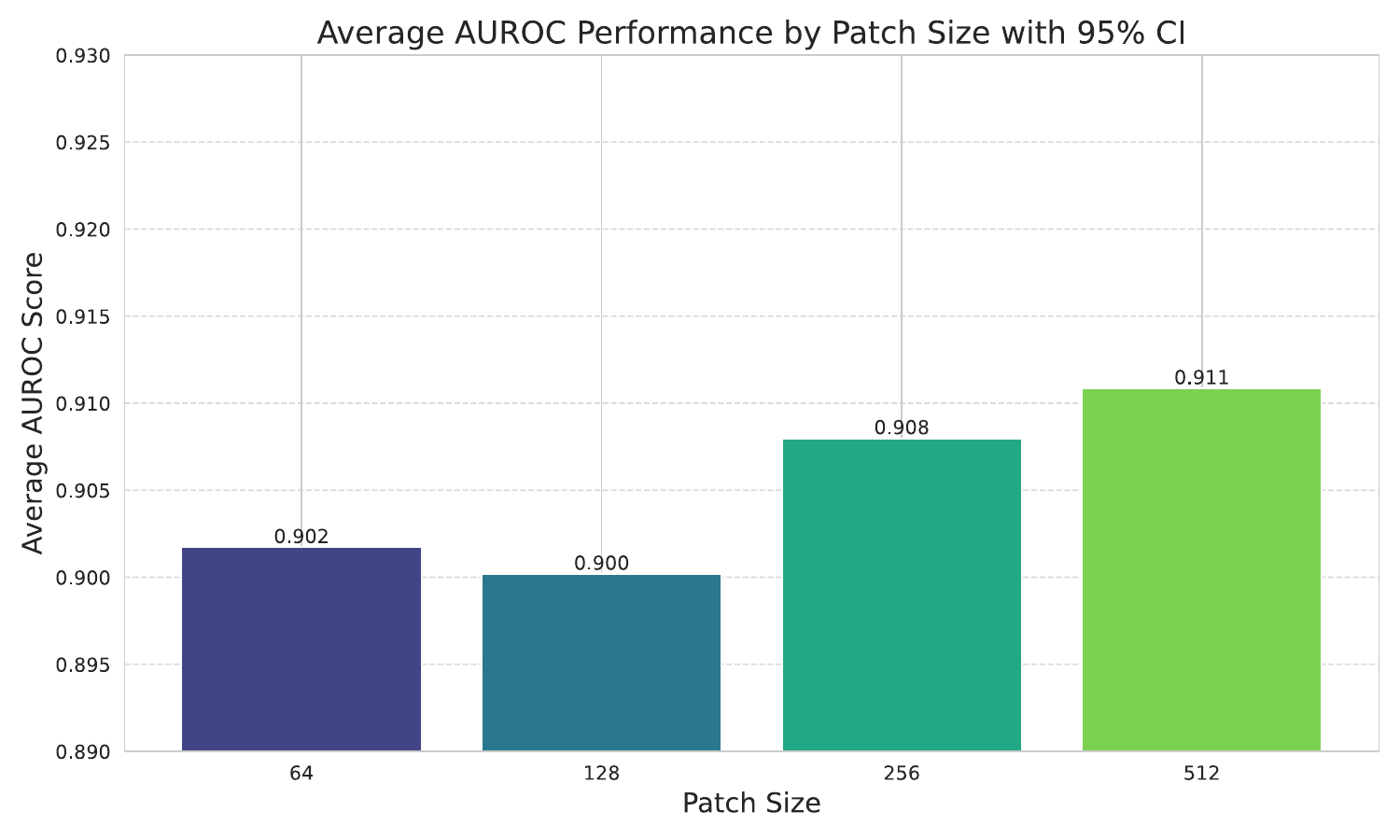}
  \caption{\chadded{Average AUROC over all pathologies as a function of patch size. The four bars correspond to patch sizes of $64\times64$, $128\times128$, $256\times256$, and $512\times512$ pixels (the latter representing the standard EfficientNetV2-S image-level classifier). Error bars denote 95\% confidence intervals.}}
  \label{fig:avgPatchSizeMultiDisease}
\end{figure}

\chadded{To investigate the effect of patch size on classification performance, we evaluated MedicalPatchNet not only with a patch size of $64\times64$ pixels, which yields an $8\times8$ grid of patches per image, but also with patch sizes of $128\times128$ and $256\times256$ pixels, corresponding to $4\times4$ and $2\times2$ patches, respectively. Applying EfficientNetV2-S directly to the full image is equivalent to using a patch size of $512\times512$ pixels with a single patch, i.e.\ a $1\times1$ grid. As illustrated in Figure~\ref{fig:patchSizeMultiDisease}, the impact of patch size on AUROC is pathology-dependent. The most pronounced difference is observed for pneumonia: MedicalPatchNet with a patch size of $64\times64$ achieves an AUROC of 0.815, whereas the variant with a patch size of $256\times256$ reaches an AUROC of 0.906. In contrast, for some findings such as ``Support Devices'', the AUROC values for all tested patch sizes lie in a narrow range between 0.966 and 0.971. When comparing AUROC across patch sizes, as summarized in Figure~\ref{fig:avgPatchSizeMultiDisease}, a general trend emerges in which larger patch sizes yield slightly higher average performance, even though all configurations operate within a relatively small AUROC range of approximately 0.900 to 0.911. It is important to emphasize, however, that increasing the patch size simultaneously reduces the spatial resolution of the saliency maps: once a patch containing a pathology is highlighted, a larger patch size leads to the corresponding heatmap covering a larger area of the image.}
\section{Margin Analysis}
\begin{figure}[H]
    \centering
    % Row 1
    \begin{subfigure}[t]{0.24\textwidth}
        \centering
        \includegraphics[width=\linewidth]{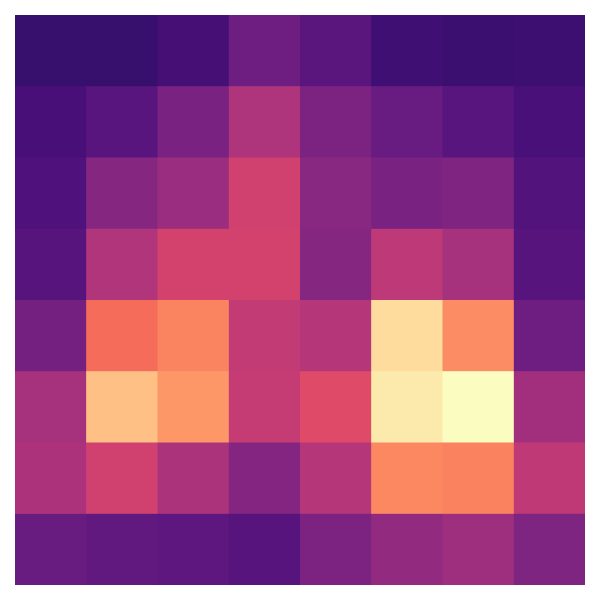}
        \caption{All classes}
        \label{fig:avgmap_onestep_all}
    \end{subfigure}
    \hfill
    \begin{subfigure}[t]{0.24\textwidth}
        \centering
        \includegraphics[width=\linewidth]{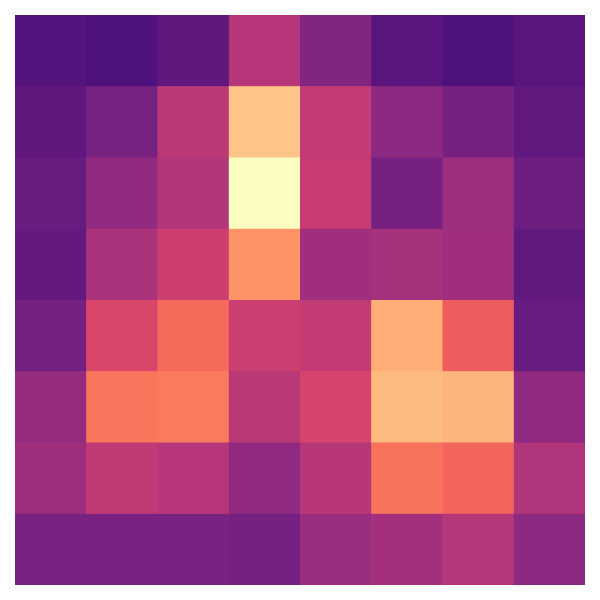}
        \caption{No finding}
        \label{fig:avgmap_onestep_nofinding}
    \end{subfigure}
    \hfill
    \begin{subfigure}[t]{0.24\textwidth}
        \centering
        \includegraphics[width=\linewidth]{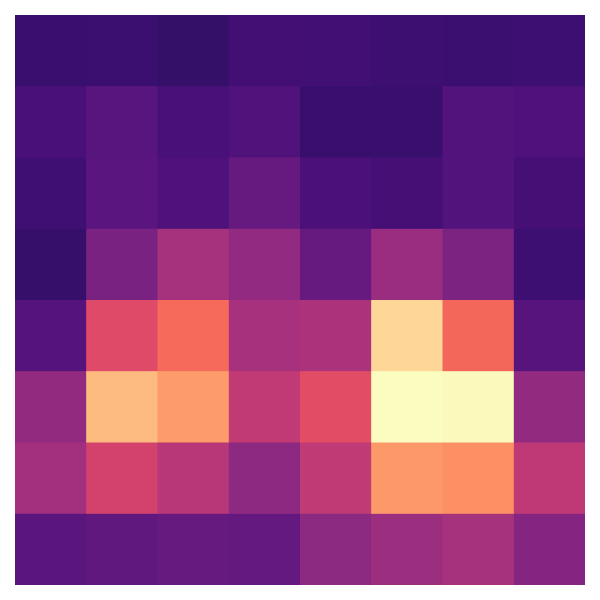}
        \caption{Lung opacity}
        \label{fig:avgmap_onestep_lungopacity}
    \end{subfigure}
    \hfill
    \begin{subfigure}[t]{0.24\textwidth}
        \centering
        \includegraphics[width=\linewidth]{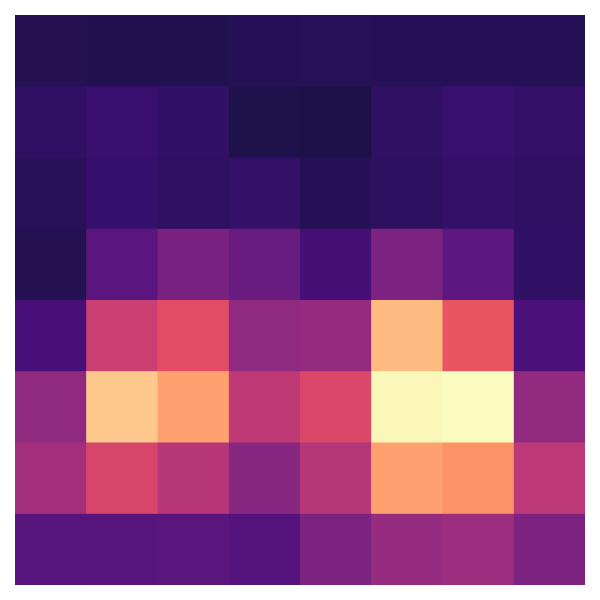}
        \caption{Atelectasis}
        \label{fig:avgmap_onestep_atelectasis}
    \end{subfigure}

    \vspace{0.6em}

    % Row 2
    \begin{subfigure}[t]{0.24\textwidth}
        \centering
        \includegraphics[width=\linewidth]{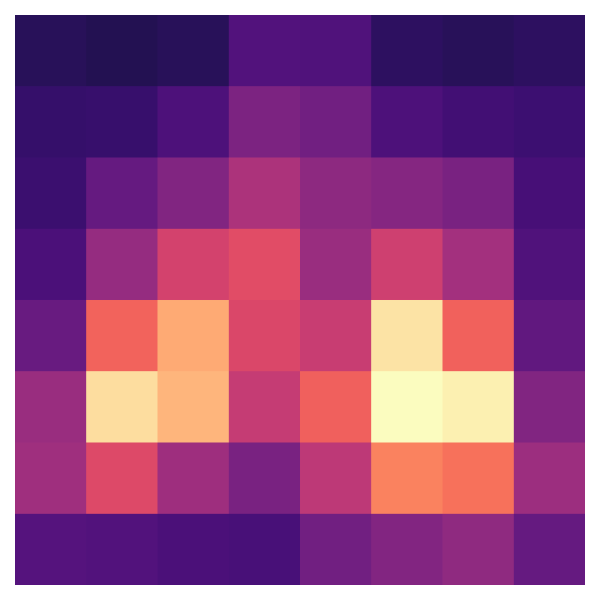}
        \caption{Cardiomegaly}
        \label{fig:avgmap_onestep_cardiomegaly}
    \end{subfigure}
    \hfill
    \begin{subfigure}[t]{0.24\textwidth}
        \centering
        \includegraphics[width=\linewidth]{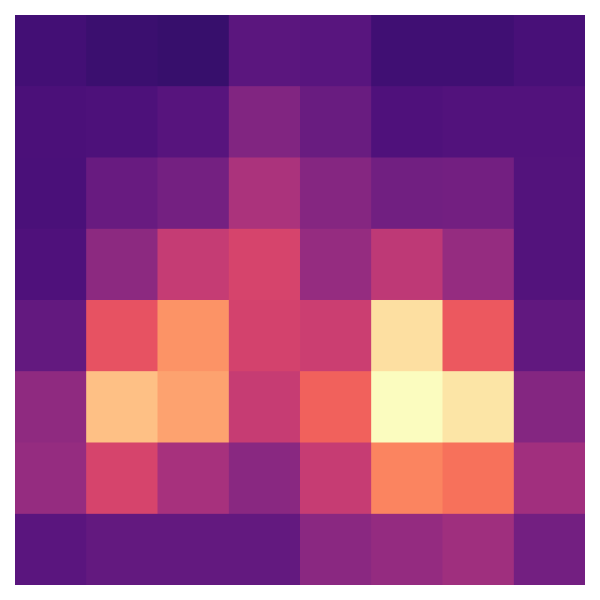}
        \caption{Enlarged cardiomediastinum}
        \label{fig:avgmap_onestep_enlargedcardio}
    \end{subfigure}
    \hfill
    \begin{subfigure}[t]{0.24\textwidth}
        \centering
        \includegraphics[width=\linewidth]{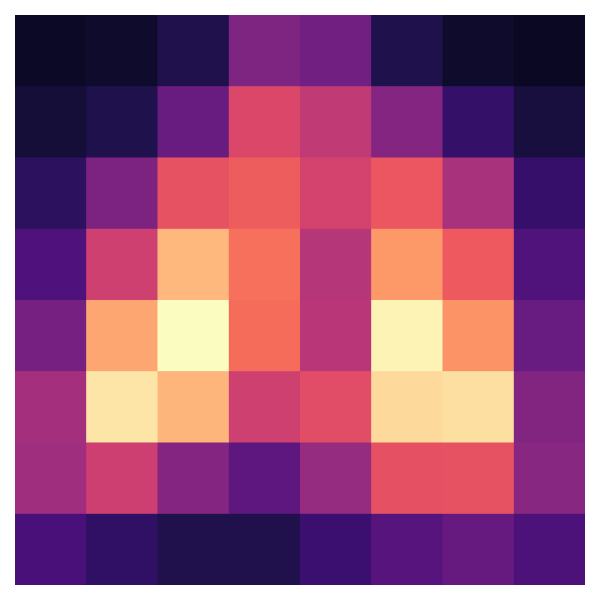}
        \caption{Edema}
        \label{fig:avgmap_onestep_edema}
    \end{subfigure}
    \hfill
    \begin{subfigure}[t]{0.24\textwidth}
        \centering
        \includegraphics[width=\linewidth]{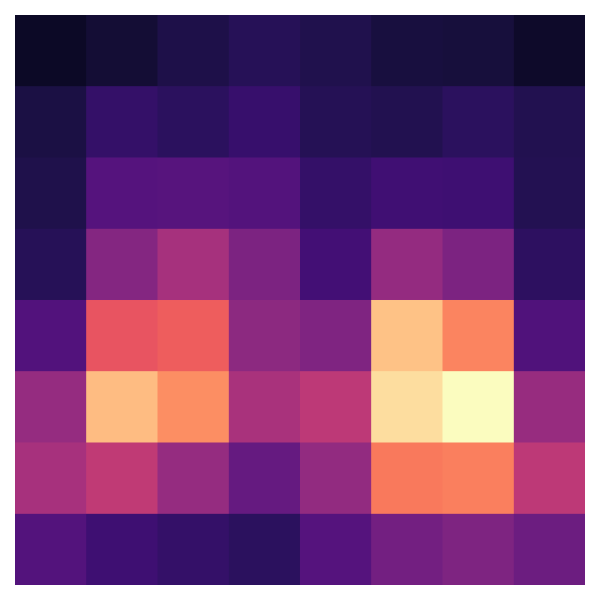}
        \caption{Consolidation}
        \label{fig:avgmap_onestep_consolidation}
    \end{subfigure}

    \vspace{0.6em}

    % Row 3
    \begin{subfigure}[t]{0.24\textwidth}
        \centering
        \includegraphics[width=\linewidth]{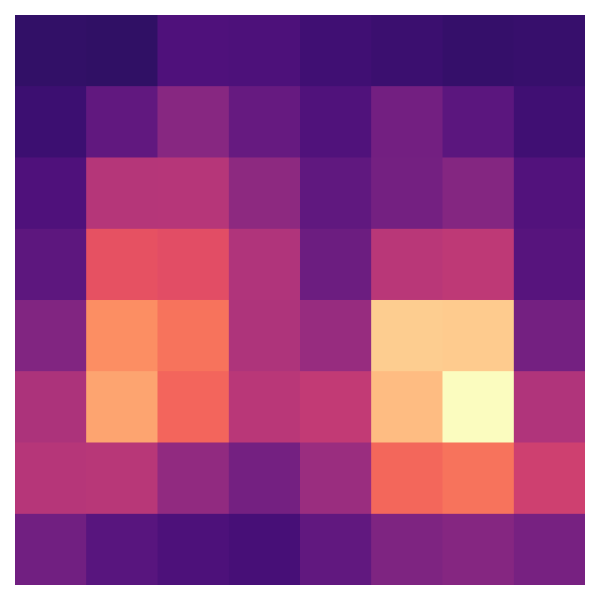}
        \caption{Pneumonia}
        \label{fig:avgmap_onestep_pneumonia}
    \end{subfigure}
    \hfill
    \begin{subfigure}[t]{0.24\textwidth}
        \centering
        \includegraphics[width=\linewidth]{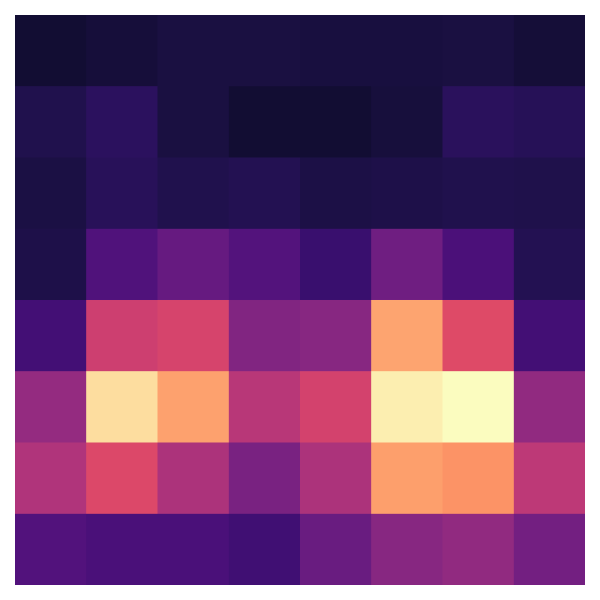}
        \caption{Pleural effusion}
        \label{fig:avgmap_onestep_pleuraleffusion}
    \end{subfigure}
    \hfill
    \begin{subfigure}[t]{0.24\textwidth}
        \centering
        \includegraphics[width=\linewidth]{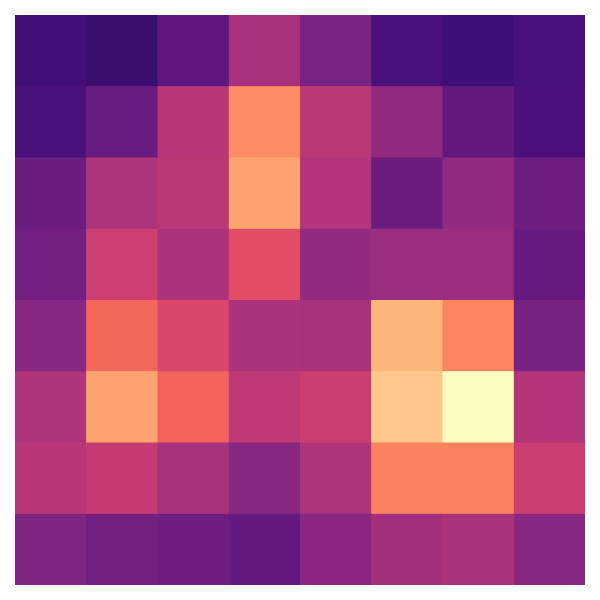}
        \caption{Pleural other}
        \label{fig:avgmap_onestep_pleuralother}
    \end{subfigure}
    \hfill
    \begin{subfigure}[t]{0.24\textwidth}
        \centering
        \includegraphics[width=\linewidth]{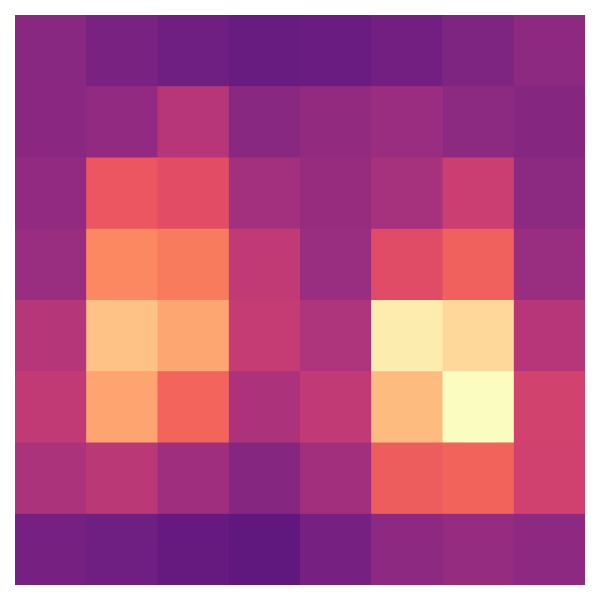}
        \caption{Pneumothorax}
        \label{fig:avgmap_onestep_pneumothorax}
    \end{subfigure}

    \vspace{0.6em}

    % Row 4 (last slot intentionally left empty)
    \begin{subfigure}[t]{0.24\textwidth}
        \centering
        \includegraphics[width=\linewidth]{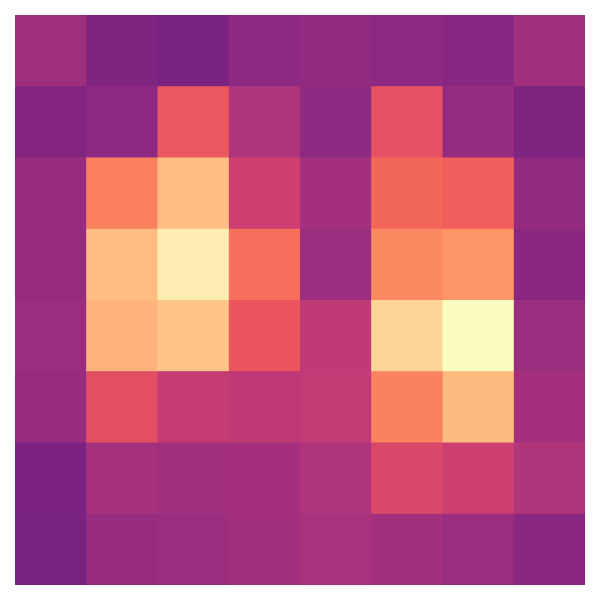}
        \caption{Lung lesion}
        \label{fig:avgmap_onestep_lunglesion}
    \end{subfigure}
    \hfill
    \begin{subfigure}[t]{0.24\textwidth}
        \centering
        \includegraphics[width=\linewidth]{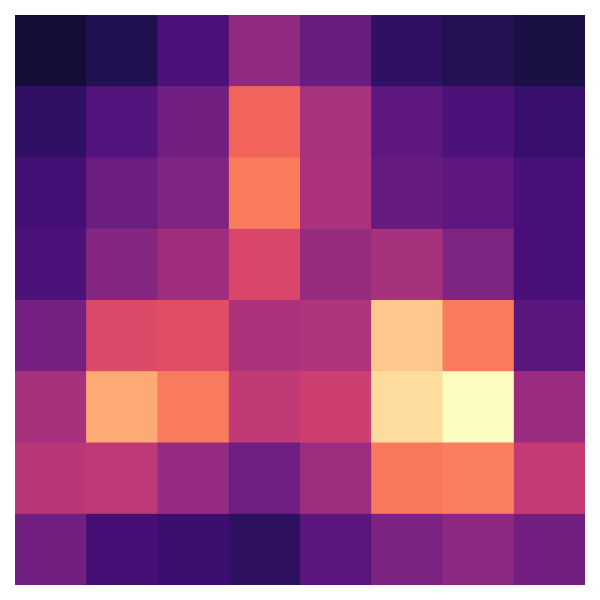}
        \caption{Fracture}
        \label{fig:avgmap_onestep_fracture}
    \end{subfigure}
    \hfill
    \begin{subfigure}[t]{0.24\textwidth}
        \centering
        \includegraphics[width=\linewidth]{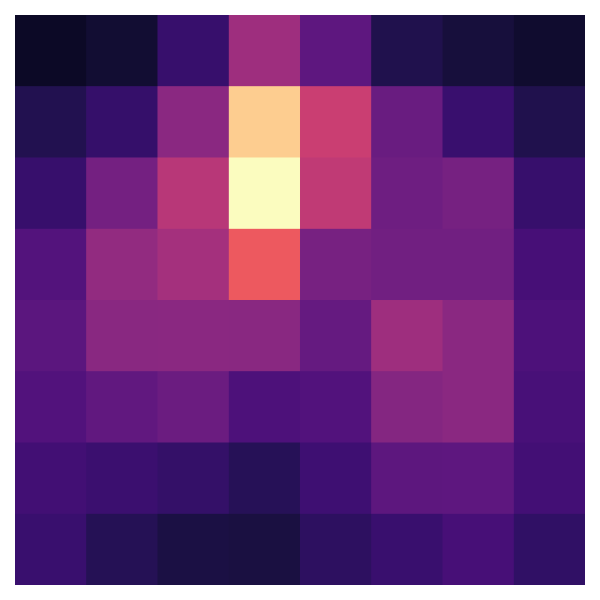}
        \caption{Support devices}
        \label{fig:avgmap_onestep_supportdevices}
        
    \end{subfigure}
    \hfill
    \begin{subfigure}[t]{0.24\textwidth}
        \includegraphics[width=\linewidth]{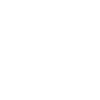}
    \end{subfigure}
    % fourth in this row intentionally omitted to keep a 4x4 layout with one empty slot

    \caption{Patch-wise average absolute logit magnitude for a single decoding step, averaged over the CheXpert test set. The first map shows the mean over all classes; the remaining maps correspond to individual CheXpert labels (the Airspace opacity label is reported as Lung opacity).}
    \label{fig:margin_onestep_cluster}
\end{figure}

MedicalPatchNet was deliberately designed as a drop-in replacement for conventional image-level classifiers (such as ResNet, EfficientNet, or Vision Transformers). Consequently, it operates directly on the full radiograph without requiring any additional inputs, for example lung masks that would restrict the analysis to predefined regions of interest. This raises the question to what extent patches at the image margins, which are often clinically less relevant, contribute to the final predictions. To investigate this, we analysed how strongly individual patches influence the output by computing the average absolute patch logits over the CheXpert test set. Figure~\ref{fig:margin_onestep_cluster} shows these mean absolute logit magnitudes, first aggregated over all classes and then separately for each CheXpert label. As visible from these maps, the patches that most strongly influence the classification form a lung-shaped region in the central part of the image, whereas patches at the margins generally contribute much less.

\section{Evaluation on natural images}
\chadded{To assess how well our architecture transfers to non-medical imagery, we additionally evaluated it on multi-label object classification using the PASCAL VOC 2012 dataset\cite{everingham2011pascalVOC}. We used the official validation split for testing and randomly divided the official training split into 80\% training and 20\% validation images. Each image was assigned all object classes that were present in its annotations.}

\chadded{For this experiment, we trained MedicalPatchNet with the same optimization and augmentation settings as in the main paper. Input images were zero-padded to a square canvas, resized to $512\times512$ pixels, and subdivided into a grid of $8\times8$ non-overlapping patches of size $64\times64$ pixels. As backbone, we again used EfficientNetV2-S.}

\chadded{We compared MedicalPatchNet to a standard EfficientNetV2-S baseline that processes the full image in a single pass and therefore has access to the complete global context. Both models were trained under identical conditions, differing only in their patch-based versus image-level architectures. The resulting performance metrics on the VOC 2012 validation set for different training durations are summarized in Table~\ref{tab:naturalIMG}.}

\thispagestyle{empty}
\begin{table}[H]
\centering
\begin{center}
\begin{tabular}{ c  c c  c ccc}
\toprule
  & \multicolumn{2}{c}{10 epochs} & \multicolumn{2}{c}{30 epochs} & \multicolumn{2}{c}{100 epochs}\\
  \cmidrule(lr){2-3} \cmidrule(lr){4-5} \cmidrule(lr){6-7}
  & MedPatchNet  & EffNetV2-S & MedPatchNet & EffNetV2-S & MedPatchNet & EffNetV2-S\\
\midrule
 AUROC & 0.915 & 0.980 & 0.959 & 0.974 & 0.951 & 0.972 \\ 
 Accuracy & 0.945 & 0.979 & 0.966 & 0.977 & 0.965 & 0.977 \\  
 Sensitivity & 0.294 & 0.822 & 0.640 & 0.814 & 0.644 & 0.820 \\  
 Specificity & 0.988 & 0.990 & 0.989 & 0.990 & 0.988 & 0.989 \\   
 Precision & 0.640 & 0.873 & 0.822 & 0.853 & 0.823 & 0.849 \\   
 F1-Score & 0.368 & 0.845 & 0.717 & 0.831 & 0.721 & 0.834 \\   
 \bottomrule
\end{tabular}
\end{center}
\caption{\chadded{Classification performance on the PASCAL VOC 2012 validation set for MedicalPatchNet (``MedPatchNet'') and a standard EfficientNetV2-S classifier (``EffNetV2-S'') after 10, 30, and 100 training epochs. All metrics are averaged over classes in the multi-label setting.}}
\label{tab:naturalIMG}
\end{table}

\chadded{Compared to the chest X-ray experiments, the AUROC gap between MedicalPatchNet and EfficientNetV2-S on natural images is slightly larger: after 30 epochs, EfficientNetV2-S outperforms MedicalPatchNet by an absolute AUROC margin of 0.015. Nonetheless, MedicalPatchNet reaches competitive performance while retaining its inherent patch-wise explainability.}

\begin{figure}[htbp]
    \centering
    % 1. Bild
    \begin{subfigure}[t]{0.24\textwidth}
        \centering
        \includegraphics[width=\linewidth]{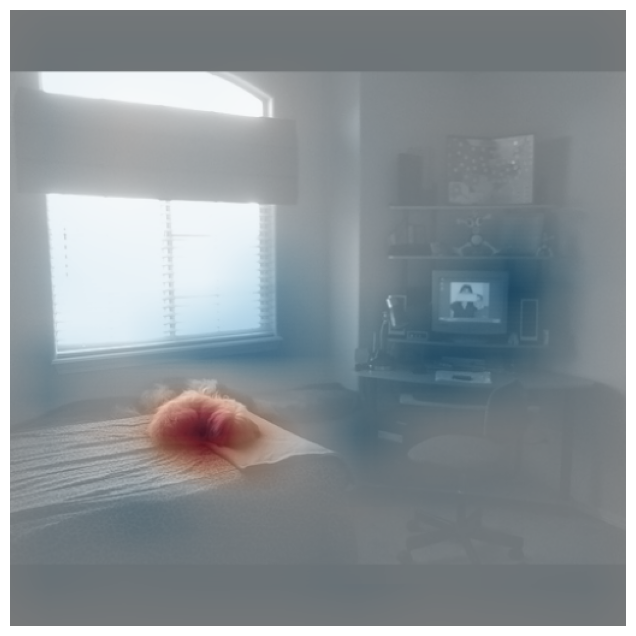}
        \caption{Dog}
        \label{fig:voc_example_1}
    \end{subfigure}
    \hfill
    % 2. Bild
    \begin{subfigure}[t]{0.24\textwidth}
        \centering
        \includegraphics[width=\linewidth]{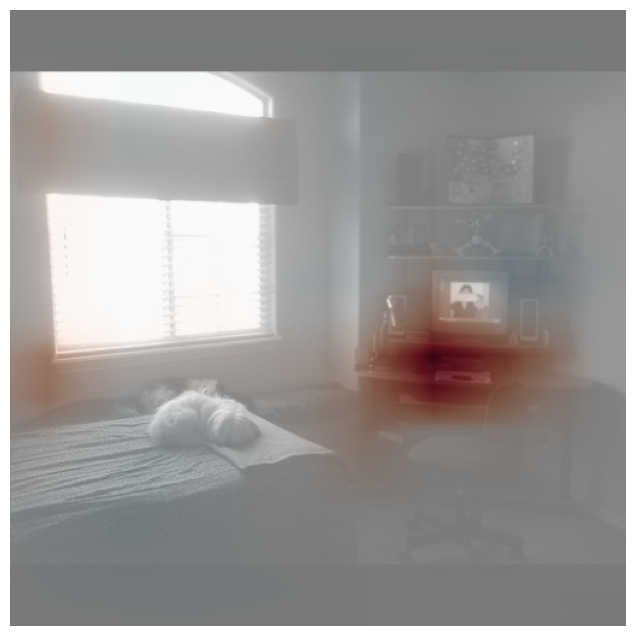}
        \caption{TV-Monitor}
        \label{fig:voc_example_2}
    \end{subfigure}
    \hfill
    % 3. Bild
    \begin{subfigure}[t]{0.24\textwidth}
        \centering
        \includegraphics[width=\linewidth]{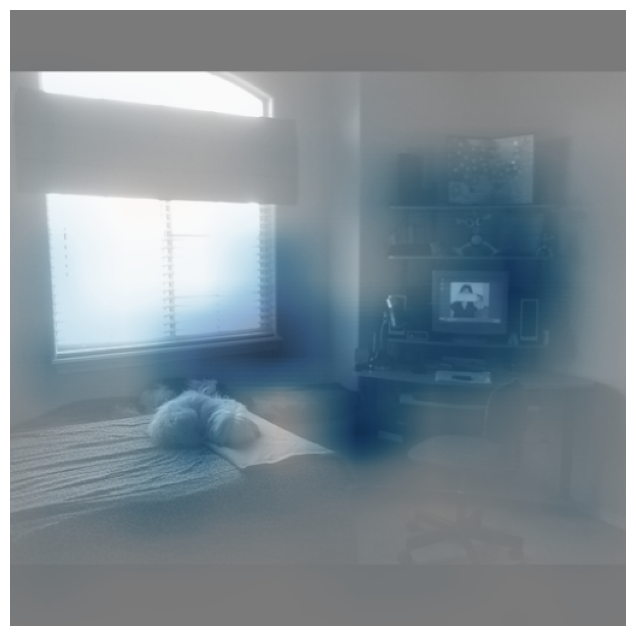}
        \caption{Aeroplane}
        \label{fig:voc_example_3}
    \end{subfigure}
    \hfill
    % 4. Bild
    \begin{subfigure}[t]{0.24\textwidth}
        \centering
        \includegraphics[width=\linewidth]{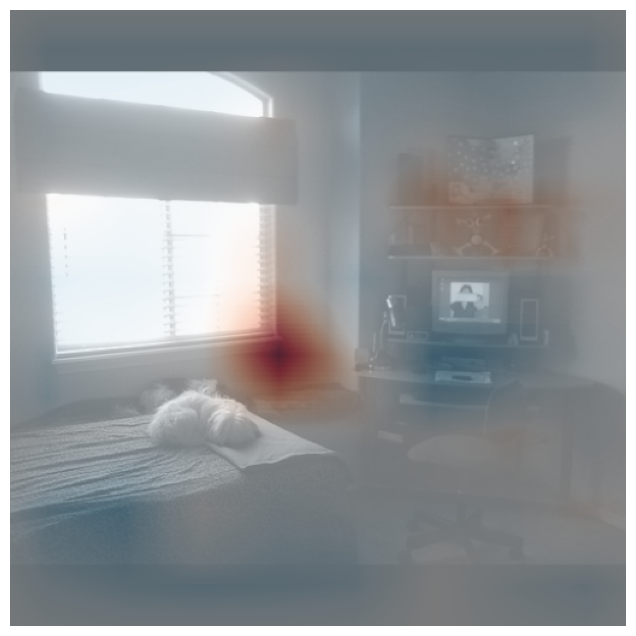}
        \caption{Potted Plant}
        \label{fig:voc_example_4}
    \end{subfigure}
    \vspace{1.2em}

    % 1. Bild
    \begin{subfigure}[t]{0.24\textwidth}
        \centering
        \includegraphics[width=\linewidth]{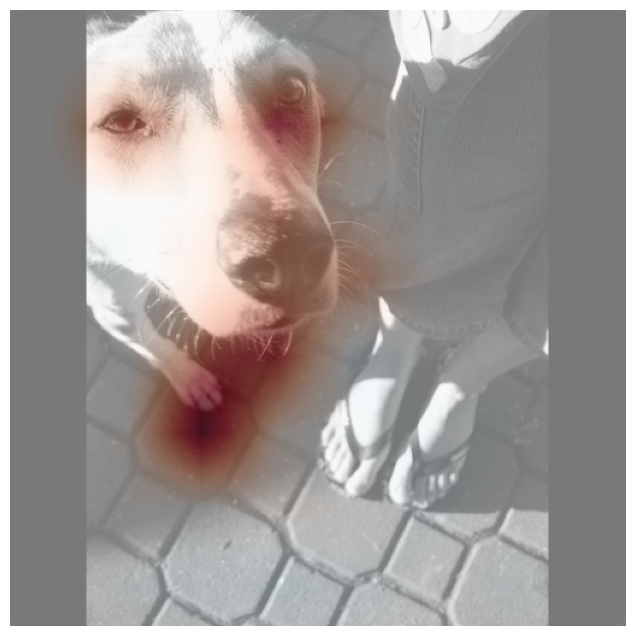}
        \caption{Dog}
        \label{fig:voc_example_1}
    \end{subfigure}
    \hfill
    % 2. Bild
    \begin{subfigure}[t]{0.24\textwidth}
        \centering
        \includegraphics[width=\linewidth]{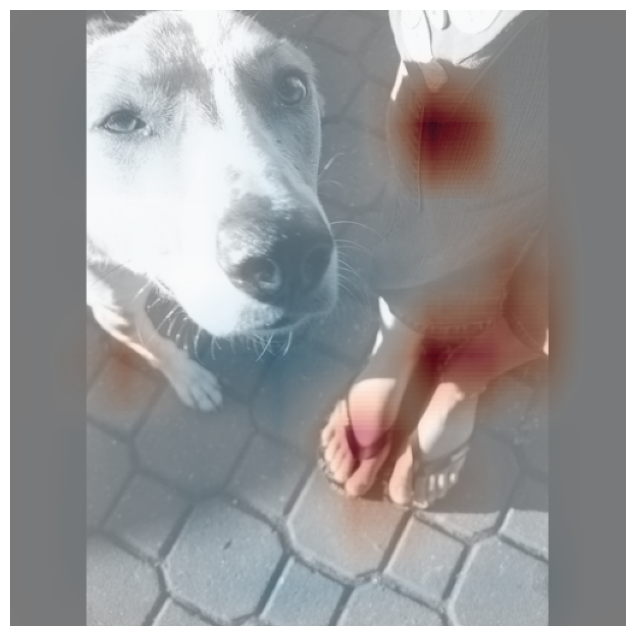}
        \caption{Person}
        \label{fig:voc_example_2}
    \end{subfigure}
    \hfill
    % 3. Bild
    \begin{subfigure}[t]{0.24\textwidth}
        \centering
        \includegraphics[width=\linewidth]{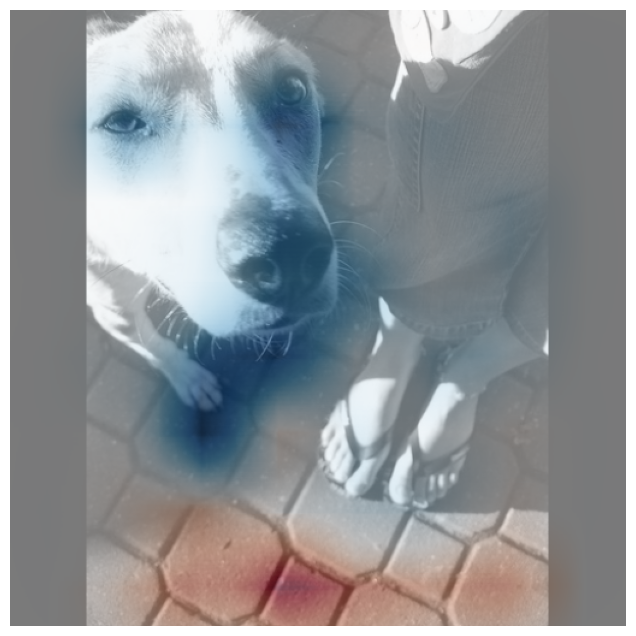}
        \caption{Bird}
        \label{fig:voc_example_3}
    \end{subfigure}
    \hfill
    % 4. Bild
    \begin{subfigure}[t]{0.24\textwidth}
        \centering
        \includegraphics[width=\linewidth]{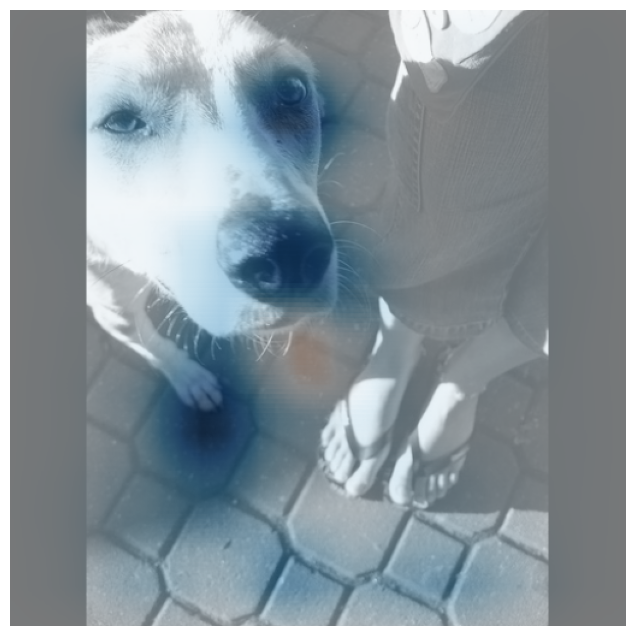}
        \caption{Sheep}
        \label{fig:voc_example_4}
    \end{subfigure}
    \vspace{1.2em}
    % 1. Bild
    \begin{subfigure}[t]{0.24\textwidth}
        \centering
        \includegraphics[width=\linewidth]{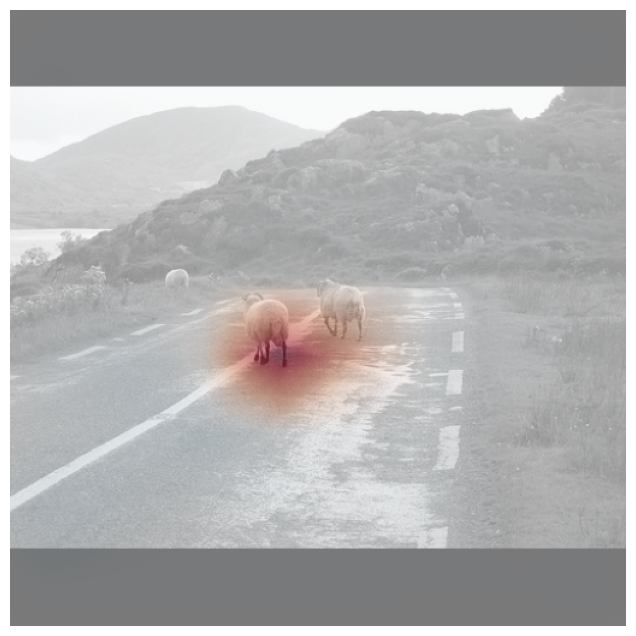}
        \caption{Sheep}
        \label{fig:voc_example_1}
    \end{subfigure}
    \hfill
    % 2. Bild
    \begin{subfigure}[t]{0.24\textwidth}
        \centering
        \includegraphics[width=\linewidth]{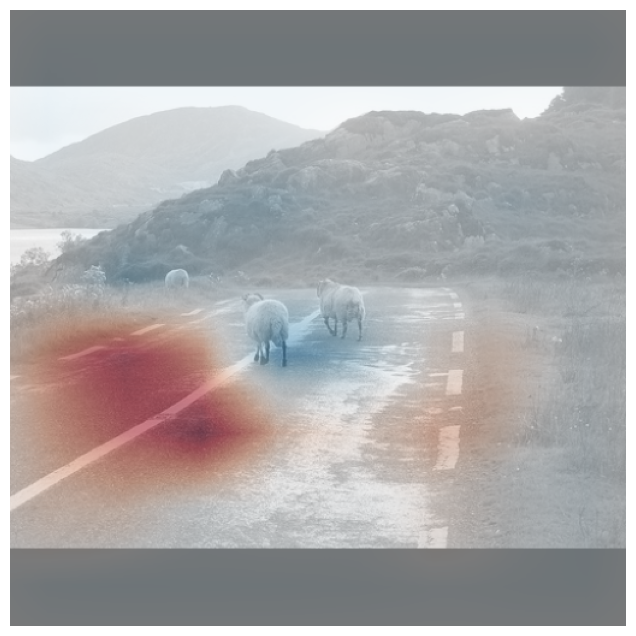}
        \caption{Car}
        \label{fig:voc_example_2}
    \end{subfigure}
    \hfill
    % 3. Bild
    \begin{subfigure}[t]{0.24\textwidth}
        \centering
        \includegraphics[width=\linewidth]{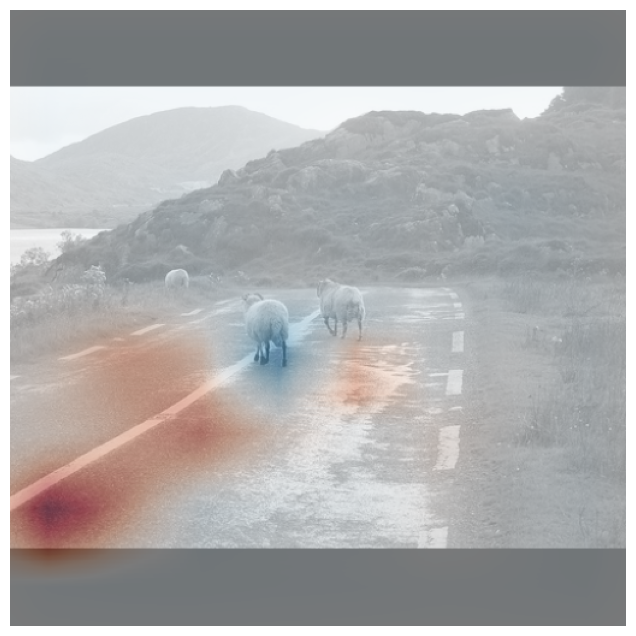}
        \caption{Bicycle}
        \label{fig:voc_example_3}
    \end{subfigure}
    \hfill
    % 4. Bild
    \begin{subfigure}[t]{0.24\textwidth}
        \centering
        \includegraphics[width=\linewidth]{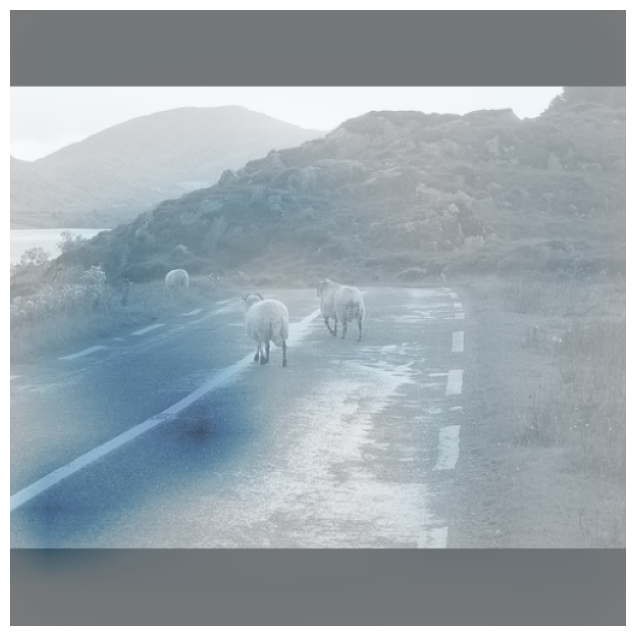}
        \caption{Sofa}
        \label{fig:voc_example_4}
    \end{subfigure}
    \caption{Representative VOC 2012 examples with corresponding saliency maps produced by MedicalPatchNet.}
    \label{fig:voc_examples}
\end{figure}

\thispagestyle{empty}
\begin{table}[H]
\centering
%\resizebox{\textwidth}{!}{%
\begin{tabular}{llllll}
\toprule
 & \multicolumn{2}{c}{MedicalPatchNet} & \multicolumn{3}{c}{EfficientNet\chreplaced{V2-S}{-B0}} \\
\cmidrule(lr){2-3} \cmidrule(lr){4-6}
Pathology & Scaled Encodings & Raw Encodings & Grad-CAM & Grad-CAM++ & Eigen-CAM \\
\midrule
Lung Opacity & 0.040 \scriptsize{[0.036--0.044]} & 0.040 \scriptsize{[0.036--0.044]} & 0.097 \scriptsize{[0.087--0.108]} & \textbf{0.101} \scriptsize{[0.091--0.112]} & \underline{0.101} \scriptsize{[0.090--0.111]} \\
Atelectasis & 0.023 \scriptsize{[0.019--0.026]} & 0.023 \scriptsize{[0.019--0.026]} & \underline{0.057} \scriptsize{[0.049--0.066]} & \textbf{0.061} \scriptsize{[0.052--0.070]} & 0.057 \scriptsize{[0.049--0.066]} \\
Cardiomegaly & 0.036 \scriptsize{[0.032--0.041]} & 0.036 \scriptsize{[0.032--0.041]} & \underline{0.061} \scriptsize{[0.053--0.069]} & \textbf{0.063} \scriptsize{[0.055--0.071]} & 0.060 \scriptsize{[0.052--0.068]} \\
Consolidation & \textbf{0.044} \scriptsize{[0.030--0.059]} & \underline{0.015} \scriptsize{[0.010--0.021]} & 0.012 \scriptsize{[0.008--0.017]} & 0.012 \scriptsize{[0.008--0.017]} & 0.012 \scriptsize{[0.008--0.017]} \\
Edema & \textbf{0.106} \scriptsize{[0.086--0.129]} & 0.048 \scriptsize{[0.039--0.059]} & 0.047 \scriptsize{[0.038--0.057]} & \underline{0.050} \scriptsize{[0.040--0.060]} & 0.048 \scriptsize{[0.039--0.058]} \\
Enlarged Card. & 0.078 \scriptsize{[0.071--0.085]} & 0.078 \scriptsize{[0.071--0.085]} & 0.129 \scriptsize{[0.117--0.140]} & \textbf{0.136} \scriptsize{[0.123--0.148]} & \underline{0.133} \scriptsize{[0.121--0.145]} \\
Lung Lesion & \underline{0.006} \scriptsize{[0.002--0.010]} & \textbf{0.029} \scriptsize{[0.011--0.050]} & 0.003 \scriptsize{[0.001--0.005]} & 0.003 \scriptsize{[0.001--0.006]} & 0.003 \scriptsize{[0.001--0.006]} \\
Pleural Effusion & \textbf{0.109} \scriptsize{[0.088--0.132]} & \underline{0.061} \scriptsize{[0.049--0.074]} & 0.031 \scriptsize{[0.024--0.039]} & 0.033 \scriptsize{[0.025--0.041]} & 0.026 \scriptsize{[0.020--0.032]} \\
Pneumothorax & \textbf{0.079} \scriptsize{[0.030--0.135]} & \underline{0.060} \scriptsize{[0.019--0.114]} & 0.004 \scriptsize{[0.001--0.006]} & 0.004 \scriptsize{[0.001--0.006]} & 0.002 \scriptsize{[0.001--0.004]} \\
Support Devices & \underline{0.166} \scriptsize{[0.156--0.176]} & \textbf{0.171} \scriptsize{[0.158--0.184]} & 0.075 \scriptsize{[0.068--0.083]} & 0.075 \scriptsize{[0.068--0.082]} & 0.047 \scriptsize{[0.041--0.052]} \\
\midrule
Mean & \textbf{0.069} & \underline{0.056} & 0.052 & 0.054 & 0.049 \\
\bottomrule
\end{tabular}
%}
\caption{Mean Intersection over Union (mIoU) for all cases (true positive, false positive, and false negative). The highest and second-highest means per row are bold and underlined, respectively. Brackets denote 95\% confidence intervals.}
\label{tab:miou_comparison}
\end{table}

\begin{table}[H]
\centering
%\resizebox{\textwidth}{!}{%
\begin{tabular}{llllll}
\toprule
 & \multicolumn{2}{c}{MedicalPatchNet} & \multicolumn{3}{c}{EfficientNet\chreplaced{V2-S}{-B0}} \\
\cmidrule(lr){2-3} \cmidrule(lr){4-6}
Pathology & Scaled Encodings & Raw Encodings & Grad-CAM & Grad-CAM++ & Eigen-CAM \\
\midrule
Lung Opacity & 0.087 \scriptsize{[0.081--0.092]} & 0.087 \scriptsize{[0.082--0.092]} & 0.210 \scriptsize{[0.197--0.223]} & \textbf{0.219} \scriptsize{[0.205--0.232]} & \underline{0.218} \scriptsize{[0.204--0.232]} \\
Atelectasis & 0.086 \scriptsize{[0.079--0.092]} & 0.086 \scriptsize{[0.079--0.092]} & 0.213 \scriptsize{[0.195--0.232]} & \textbf{0.228} \scriptsize{[0.210--0.247]} & \underline{0.216} \scriptsize{[0.197--0.234]} \\
Cardiomegaly & 0.139 \scriptsize{[0.135--0.144]} & 0.138 \scriptsize{[0.134--0.143]} & \underline{0.232} \scriptsize{[0.220--0.244]} & \textbf{0.241} \scriptsize{[0.229--0.252]} & 0.228 \scriptsize{[0.217--0.239]} \\
Consolidation & \textbf{0.255} \scriptsize{[0.217--0.295]} & 0.232 \scriptsize{[0.196--0.268]} & 0.237 \scriptsize{[0.195--0.282]} & \underline{0.238} \scriptsize{[0.195--0.284]} & 0.233 \scriptsize{[0.186--0.280]} \\
Edema & 0.296 \scriptsize{[0.265--0.323]} & 0.288 \scriptsize{[0.262--0.312]} & 0.377 \scriptsize{[0.350--0.402]} & \textbf{0.403} \scriptsize{[0.377--0.426]} & \underline{0.389} \scriptsize{[0.364--0.412]} \\
Enlarged Card. & 0.175 \scriptsize{[0.171--0.180]} & 0.175 \scriptsize{[0.171--0.180]} & 0.289 \scriptsize{[0.279--0.300]} & \textbf{0.305} \scriptsize{[0.295--0.315]} & \underline{0.299} \scriptsize{[0.289--0.310]} \\
Lung Lesion & 0.071 \scriptsize{[0.038--0.116]} & 0.109 \scriptsize{[0.065--0.162]} & 0.146 \scriptsize{[0.082--0.206]} & \textbf{0.171} \scriptsize{[0.087--0.259]} & \underline{0.164} \scriptsize{[0.072--0.251]} \\
Pleural Effusion & \textbf{0.221} \scriptsize{[0.193--0.248]} & 0.069 \scriptsize{[0.048--0.092]} & 0.174 \scriptsize{[0.148--0.201]} & \underline{0.183} \scriptsize{[0.156--0.210]} & 0.145 \scriptsize{[0.121--0.169]} \\
Pneumothorax & 0.167 \scriptsize{[0.085--0.263]} & \textbf{0.235} \scriptsize{[0.153--0.322]} & \underline{0.232} \scriptsize{[0.156--0.311]} & 0.231 \scriptsize{[0.145--0.322]} & 0.158 \scriptsize{[0.083--0.234]} \\
Support Devices & \underline{0.186} \scriptsize{[0.177--0.196]} & \textbf{0.218} \scriptsize{[0.207--0.228]} & 0.160 \scriptsize{[0.151--0.169]} & 0.159 \scriptsize{[0.149--0.168]} & 0.099 \scriptsize{[0.091--0.106]} \\
\midrule
Mean & 0.168 & 0.164 & \underline{0.227} & \textbf{0.238} & 0.215 \\
\bottomrule
\end{tabular}
%}
\caption{Mean Intersection over Union (mIoU) evaluated over the true positive cases only. The highest and second-highest means per row are bold and underlined, respectively. Brackets denote 95\% confidence intervals.}
\end{table}
\begin{figure}
\centering
\includegraphics[width=\linewidth]{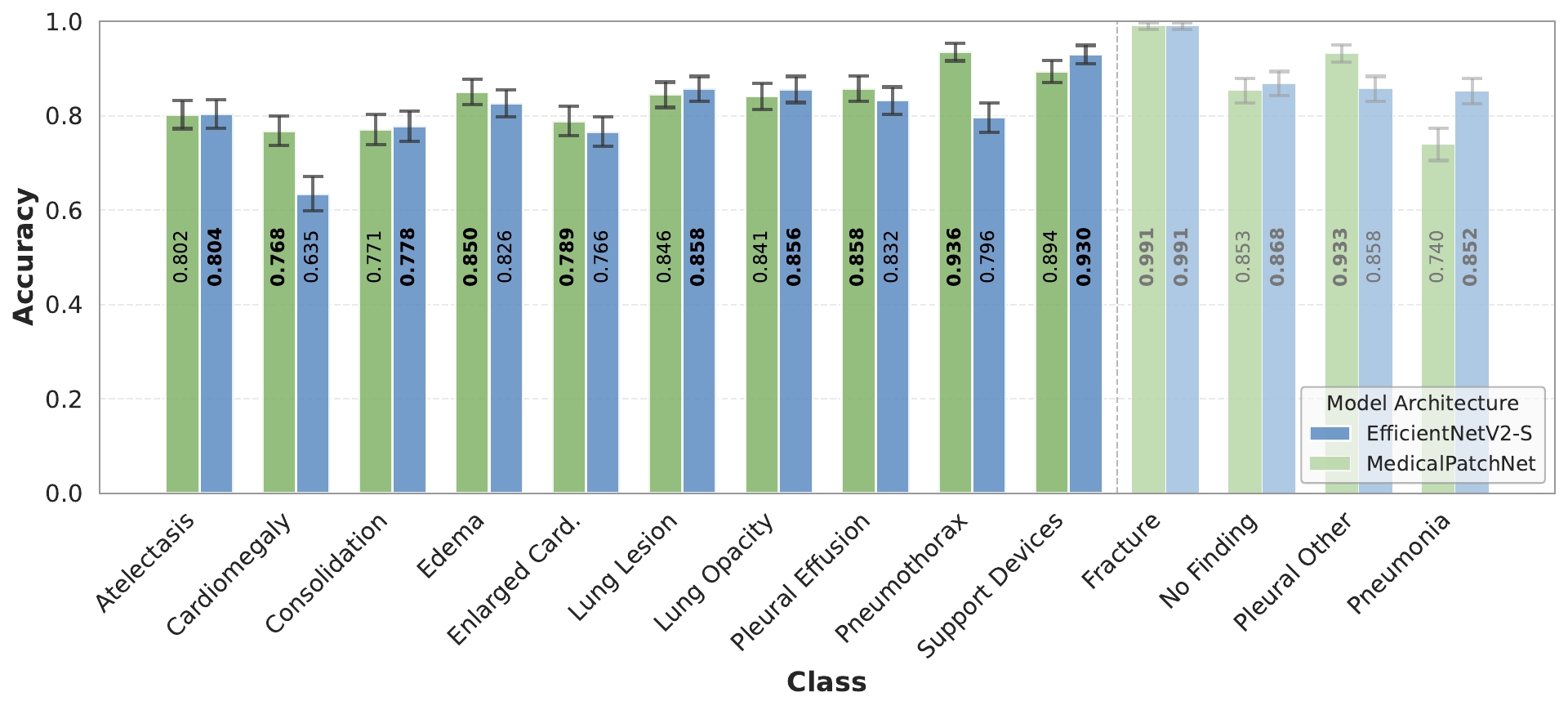}\caption{Comparison of accuracy between MedicalPatchNet and EfficientNet\chreplaced{V2-S}{-B0}. The classification threshold for each class was determined by maximizing the sum of sensitivity and specificity on the validation set. Error bars indicate 95\% confidence intervals. \chdeleted{(*) Note: The Airspace Opacity class in CheXpert is termed Lung Opacity.}}
\label{fig:auroc_comparison}
\end{figure}

\begin{figure}
\centering
\includegraphics[width=\linewidth]{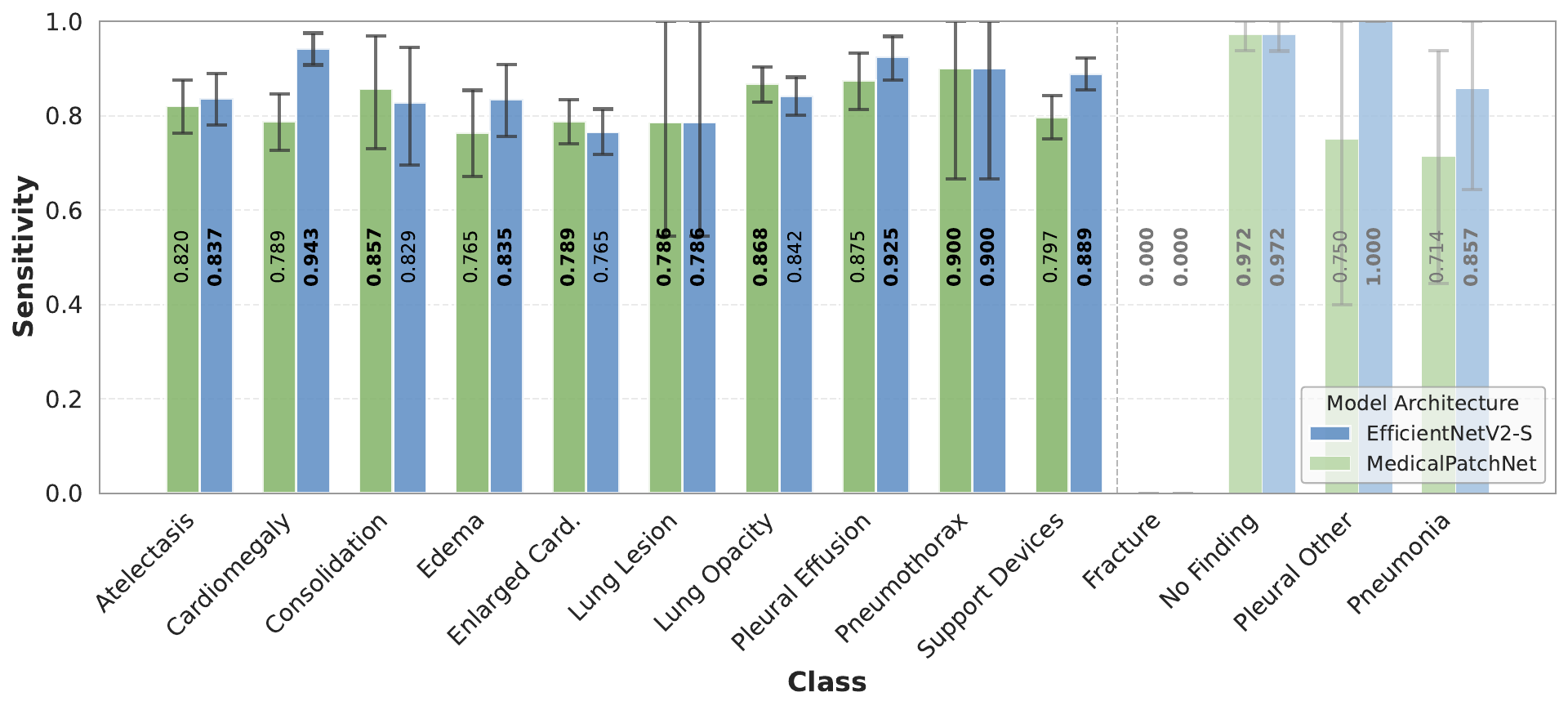}\caption{Comparison of sensitivity between MedicalPatchNet and EfficientNet\chreplaced{V2-S}{-B0}. The classification threshold for each class was determined by maximizing the sum of sensitivity and specificity on the validation set. Error bars indicate 95\% confidence intervals. \chdeleted{(*) Note: The Airspace Opacity class in CheXpert is termed Lung Opacity.}}
\label{fig:auroc_comparison}
\end{figure}

\begin{figure}
\centering
\includegraphics[width=\linewidth]{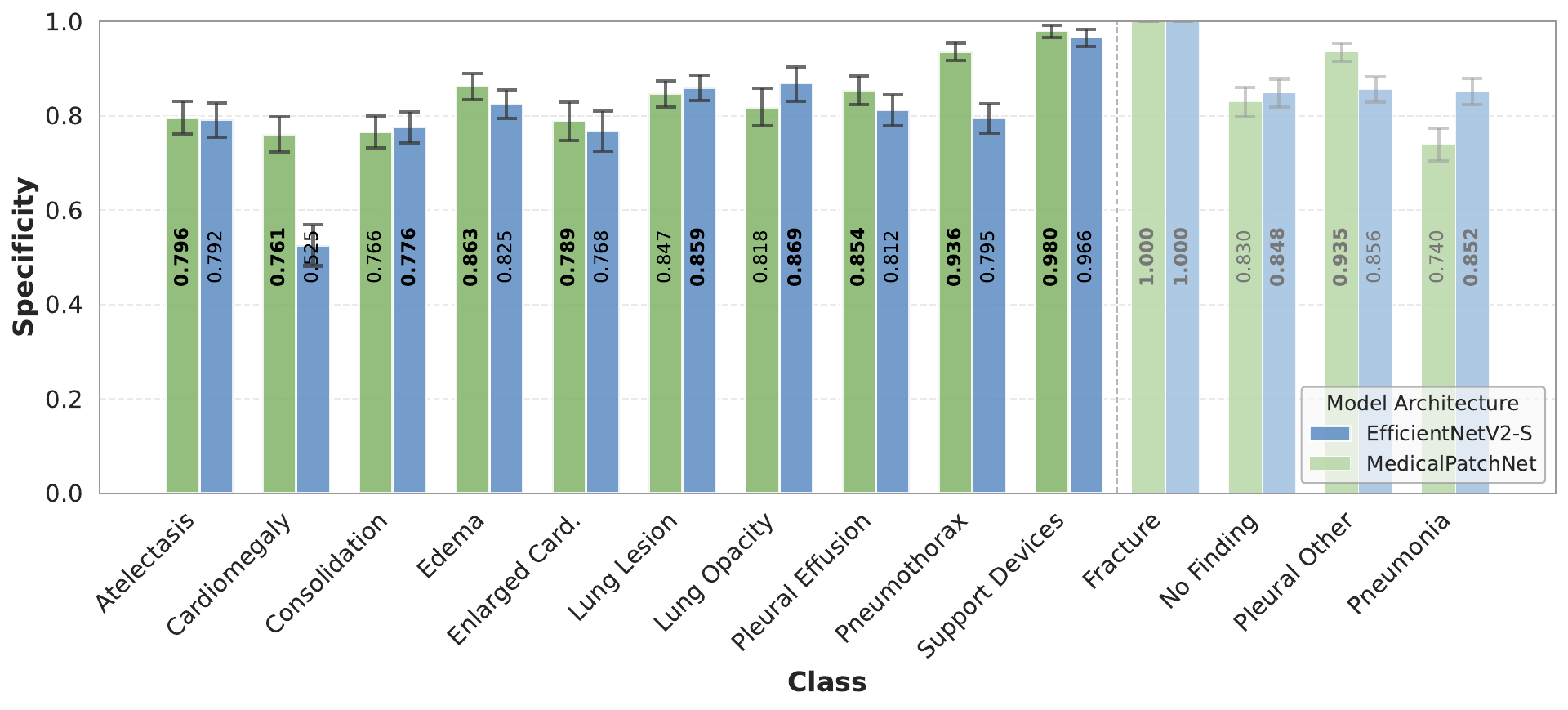}\caption{Comparison of specificity between MedicalPatchNet and EfficientNet\chreplaced{V2-S}{-B0}. The classification threshold for each class was determined by maximizing the sum of sensitivity and specificity on the validation set. Error bars indicate 95\% confidence intervals. \chdeleted{(*) Note: The Airspace Opacity class in CheXpert is termed Lung Opacity.}}
\label{fig:auroc_comparison}
\end{figure}

\section{Further Examples}
\begin{figure}[!p]
    \centering
    \scriptsize
    \setlength{\tabcolsep}{2pt}
    \renewcommand{\arraystretch}{1}
    \begin{tabular}{@{}
        L{3.0cm}
        C{0.14\textwidth}
        C{0.14\textwidth}
        C{0.14\textwidth}
        C{0.14\textwidth}
        C{0.14\textwidth}@{}}
        \toprule
        & MedicalPatchNet (raw) (our) & Grad\,-CAM & Grad\,-CAM++ & Eigen\,-CAM & Ground Truth \\
        \midrule

        Atelectasis (True) &
        \includegraphics[width=\linewidth]{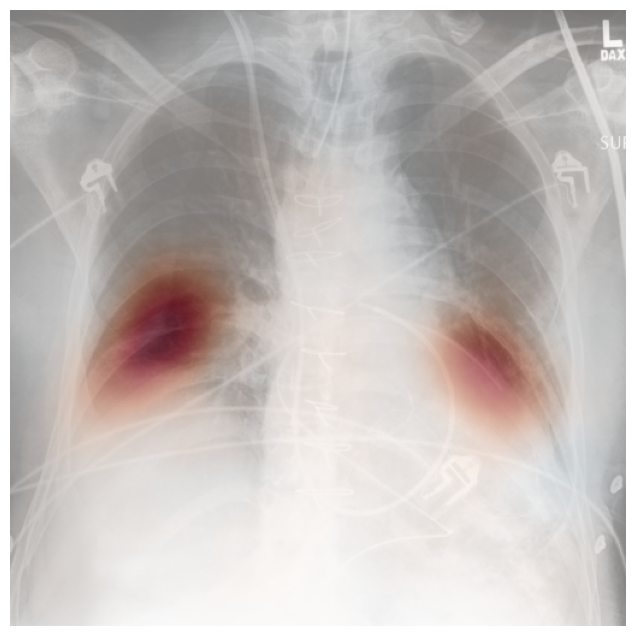} &
        \includegraphics[width=\linewidth]{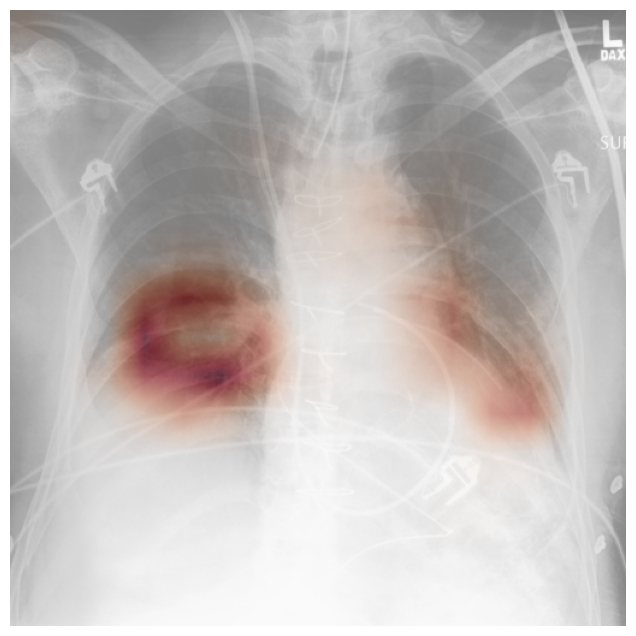} &
        \includegraphics[width=\linewidth]{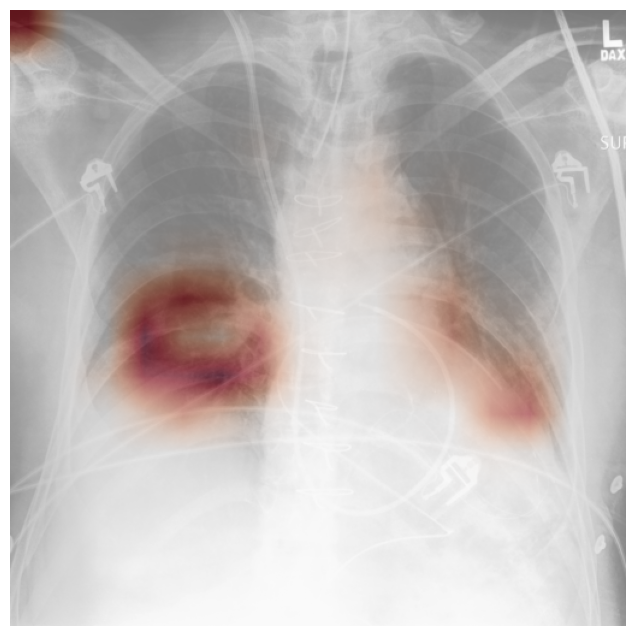} &
        \includegraphics[width=\linewidth]{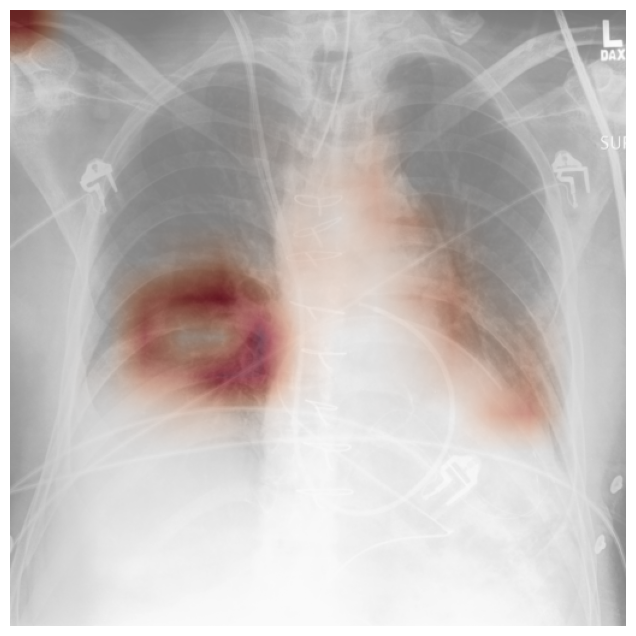} &
        \includegraphics[width=\linewidth]{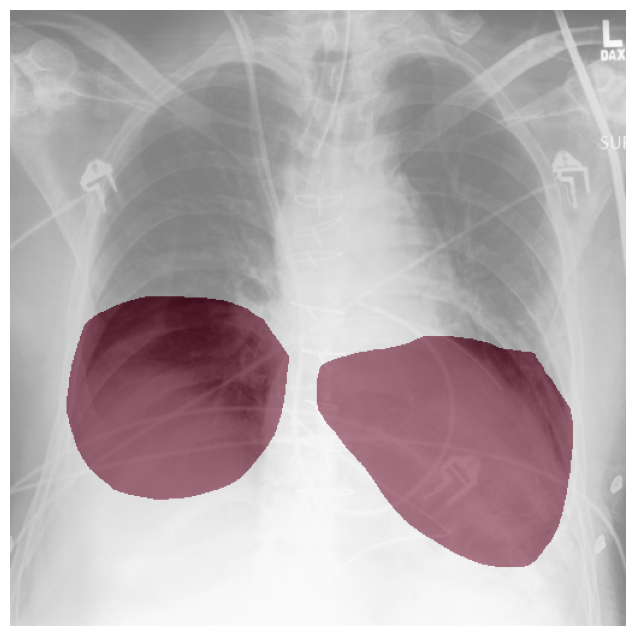} \\

        Airspace Opacity (True) &
        \includegraphics[width=\linewidth]{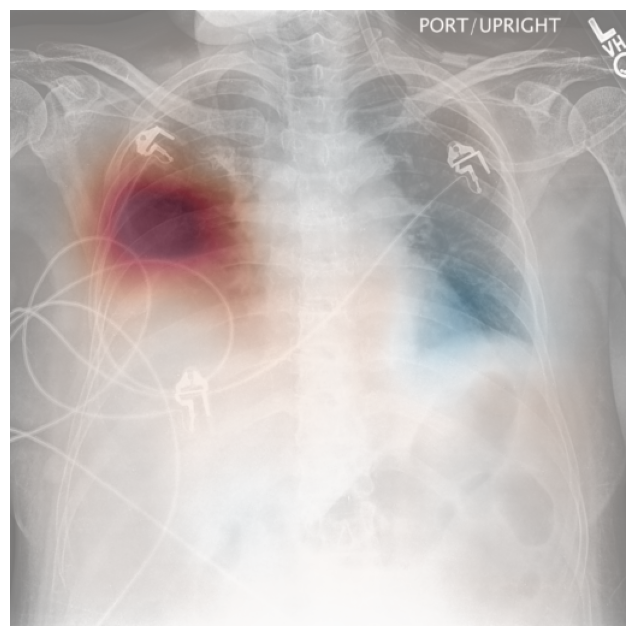} &
        \includegraphics[width=\linewidth]{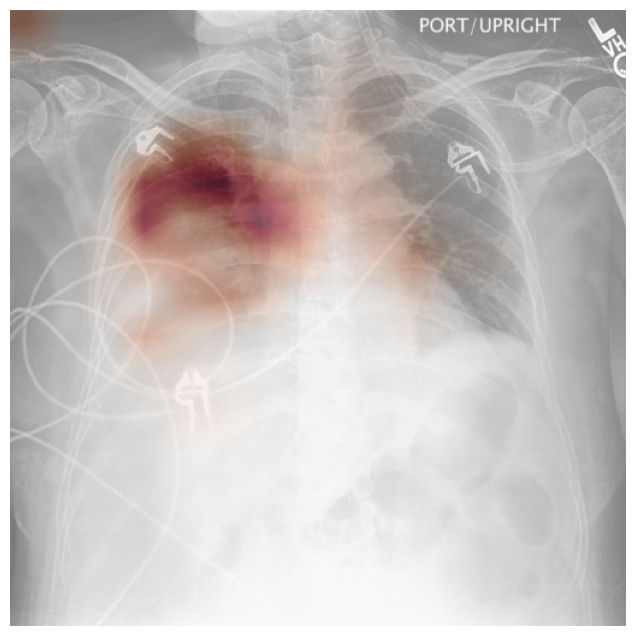} &
        \includegraphics[width=\linewidth]{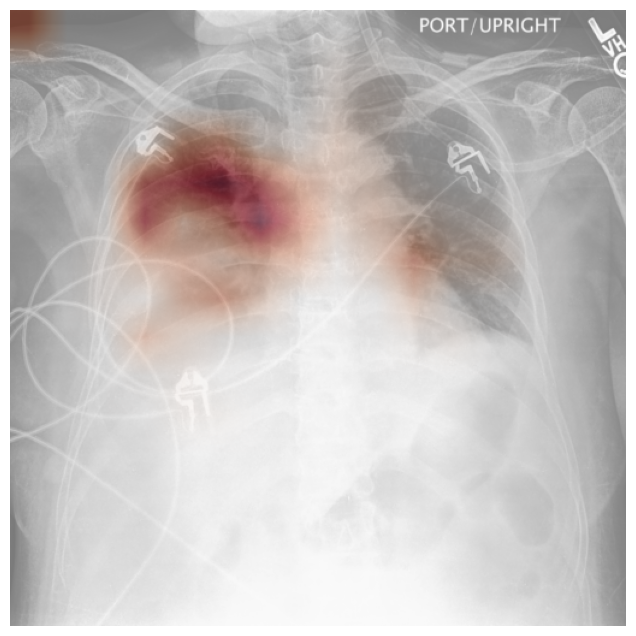} &
        \includegraphics[width=\linewidth]{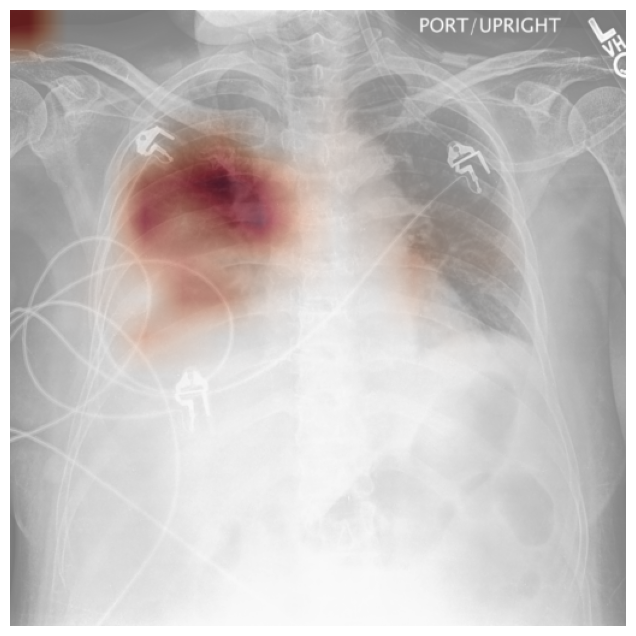} &
        \includegraphics[width=\linewidth]{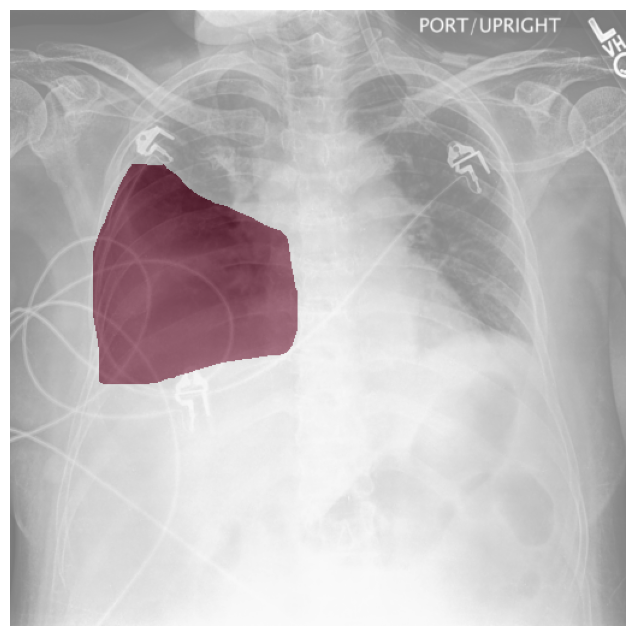} \\

        Edema (True) &
        \includegraphics[width=\linewidth]{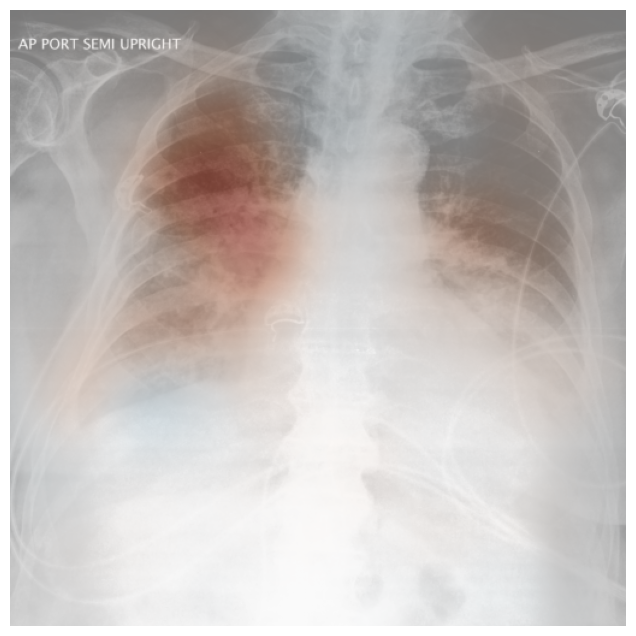} &
        \includegraphics[width=\linewidth]{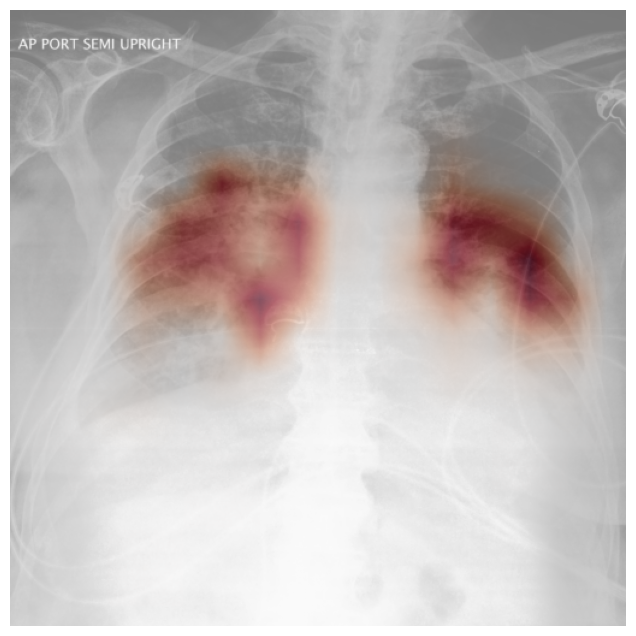} &
        \includegraphics[width=\linewidth]{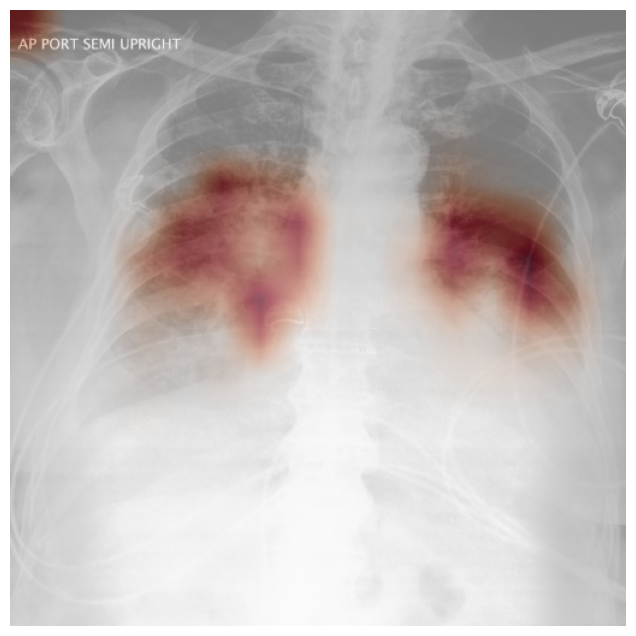} &
        \includegraphics[width=\linewidth]{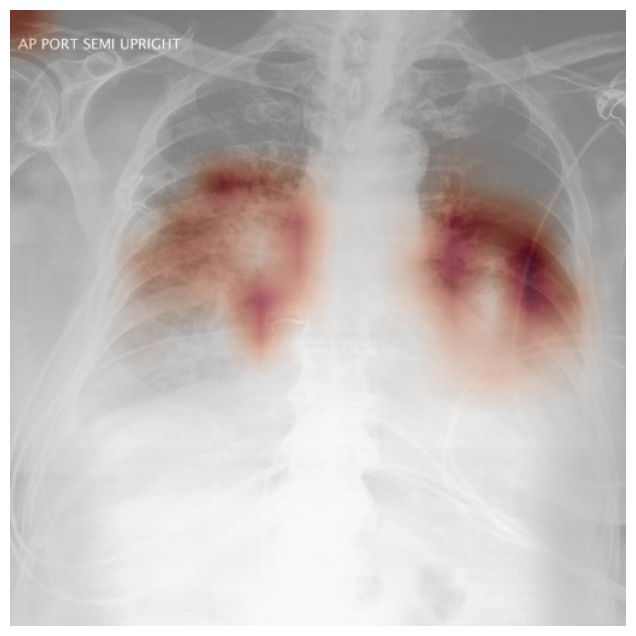} &
        \includegraphics[width=\linewidth]{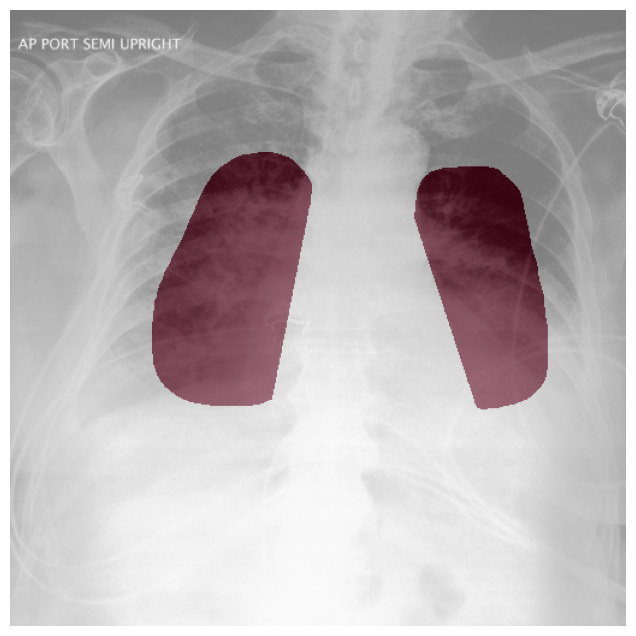} \\

        Lung Lesion (True) &
        \includegraphics[width=\linewidth]{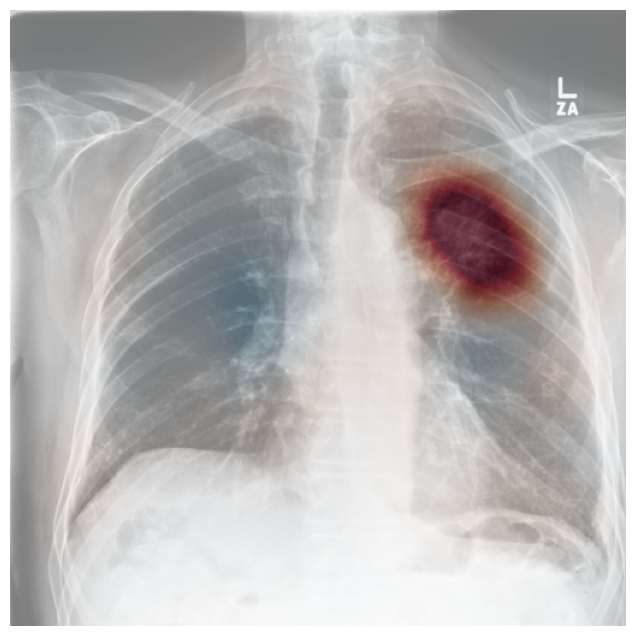} &
        \includegraphics[width=\linewidth]{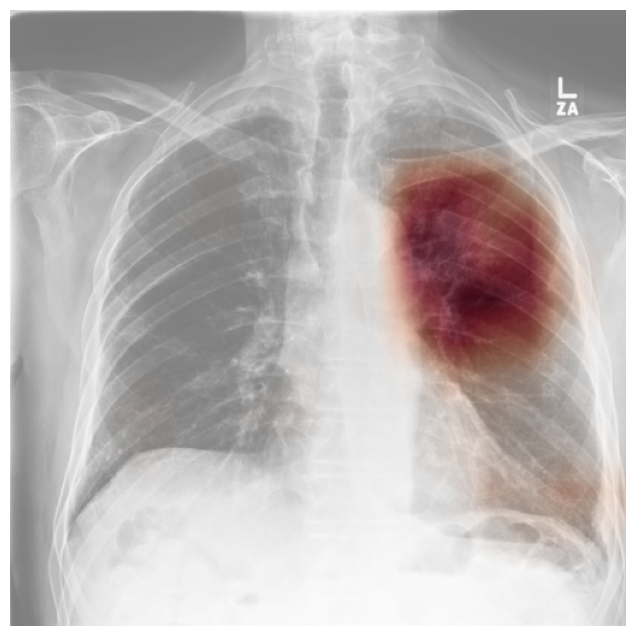} &
        \includegraphics[width=\linewidth]{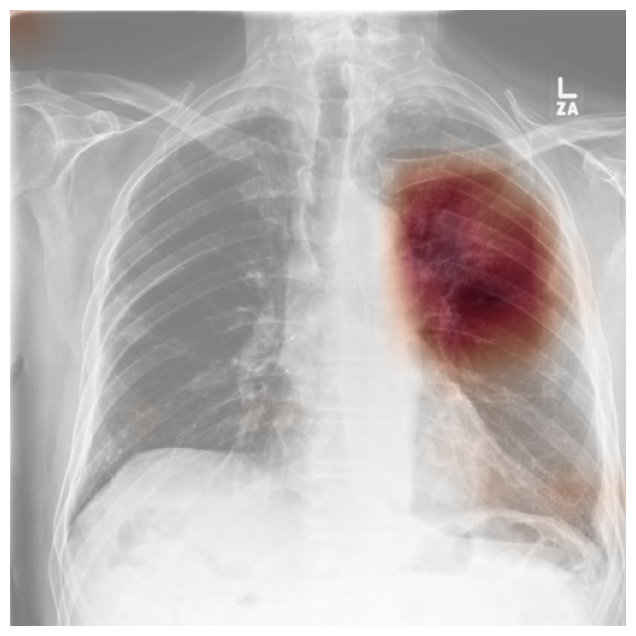} &
        \includegraphics[width=\linewidth]{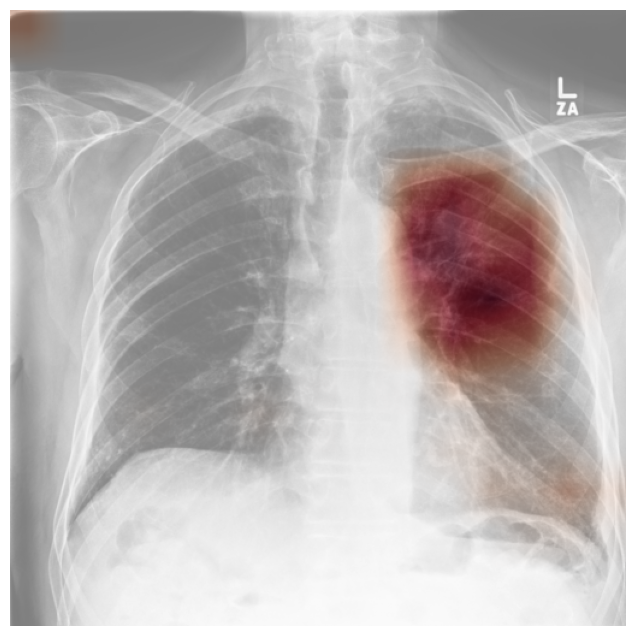} &
        \includegraphics[width=\linewidth]{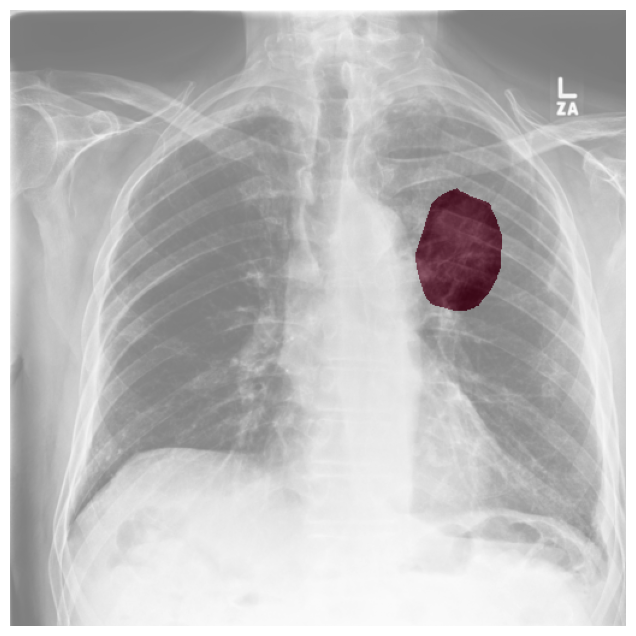} \\

        Cardiomegaly (True) &
        \includegraphics[width=\linewidth]{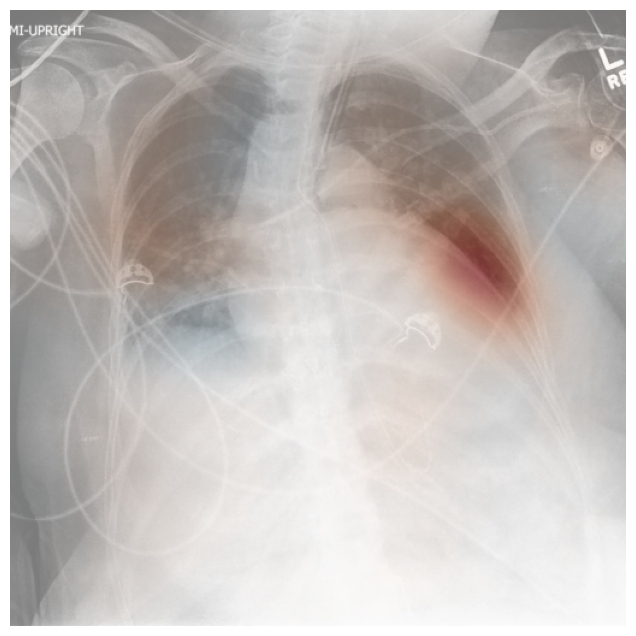} &
        \includegraphics[width=\linewidth]{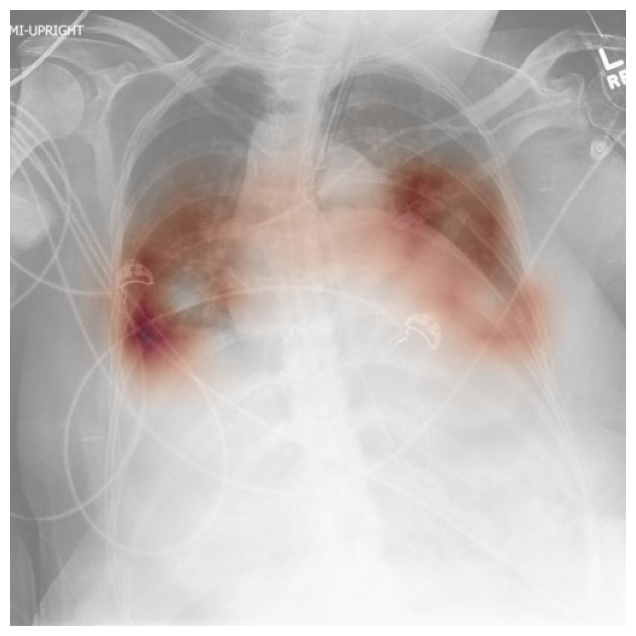} &
        \includegraphics[width=\linewidth]{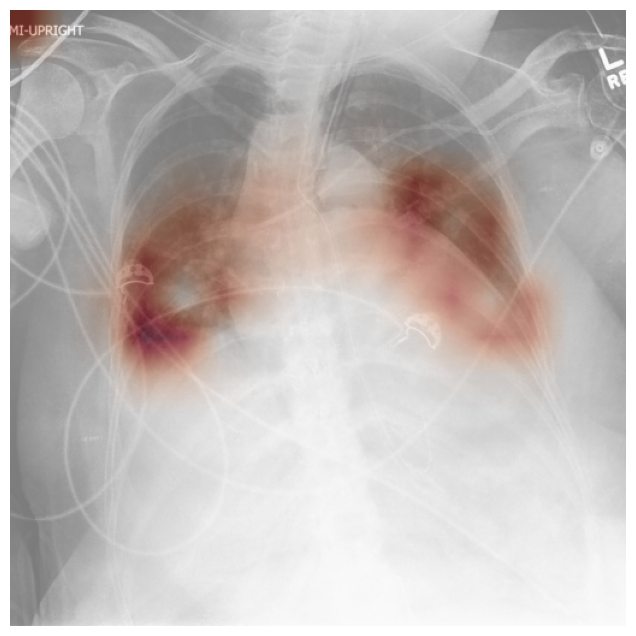} &
        \includegraphics[width=\linewidth]{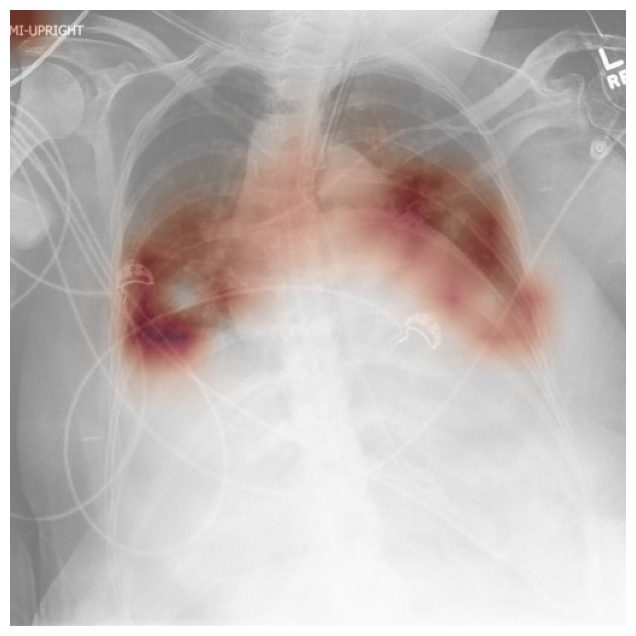} &
        \includegraphics[width=\linewidth]{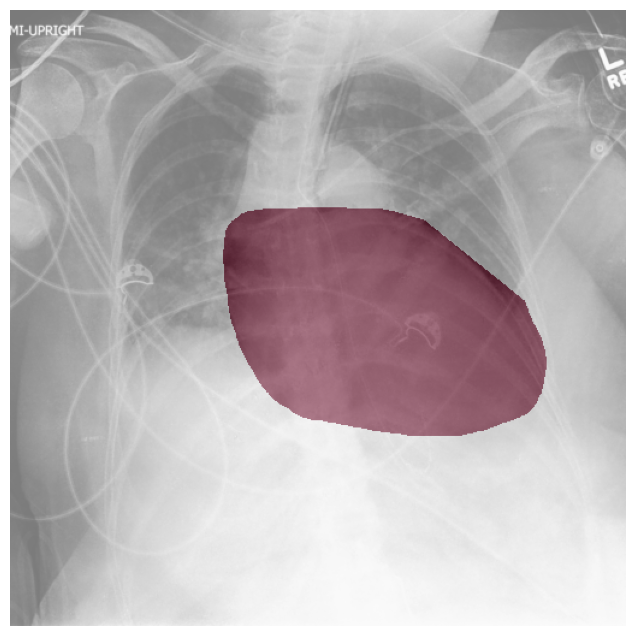} \\

        Consolidation (True) &
        \includegraphics[width=\linewidth]{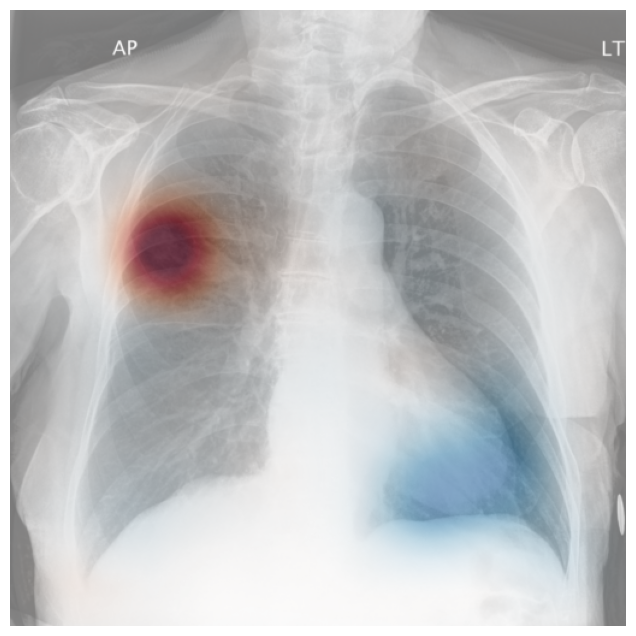} &
        \includegraphics[width=\linewidth]{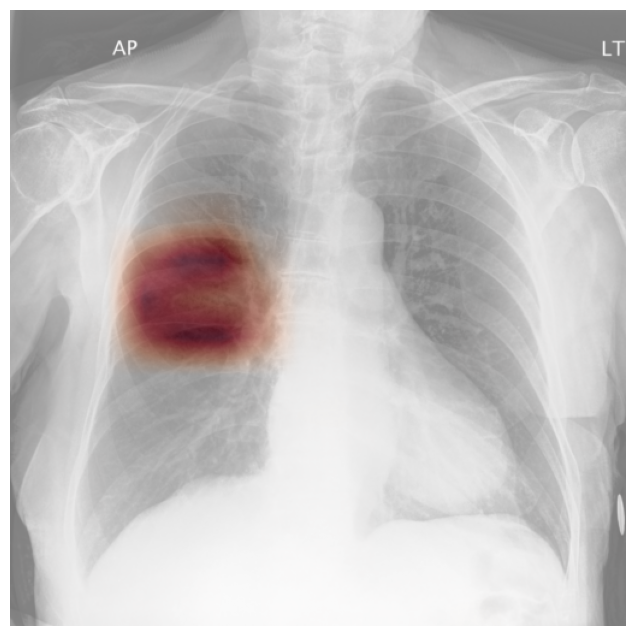} &
        \includegraphics[width=\linewidth]{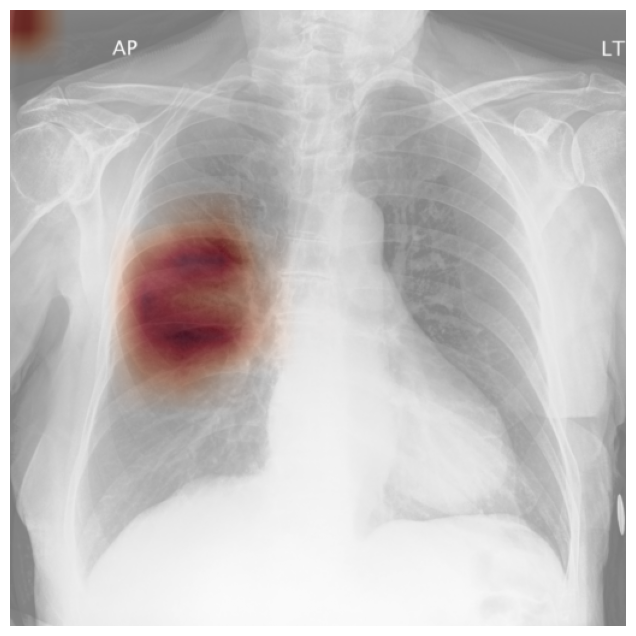} &
        \includegraphics[width=\linewidth]{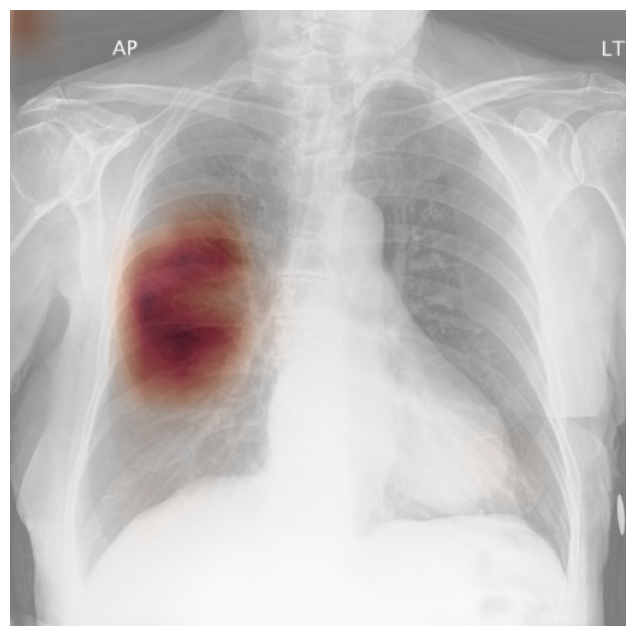} &
        \includegraphics[width=\linewidth]{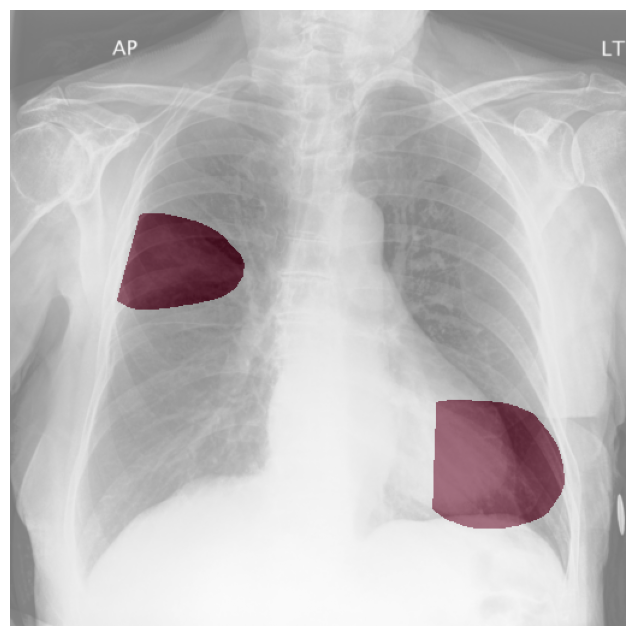} \\

        Enlarged Cardiomediastinum (True) &
        \includegraphics[width=\linewidth]{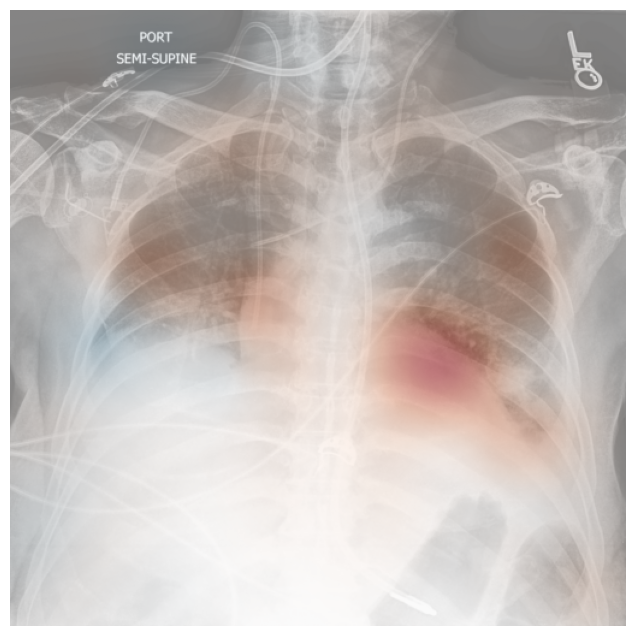} &
        \includegraphics[width=\linewidth]{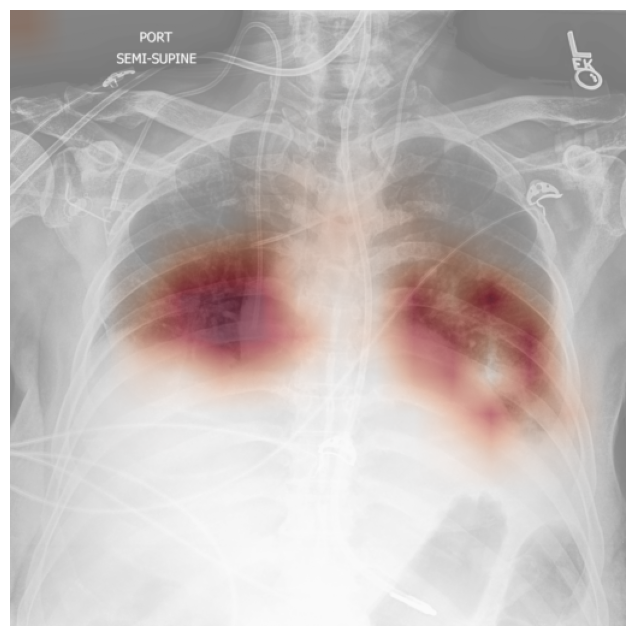} &
        \includegraphics[width=\linewidth]{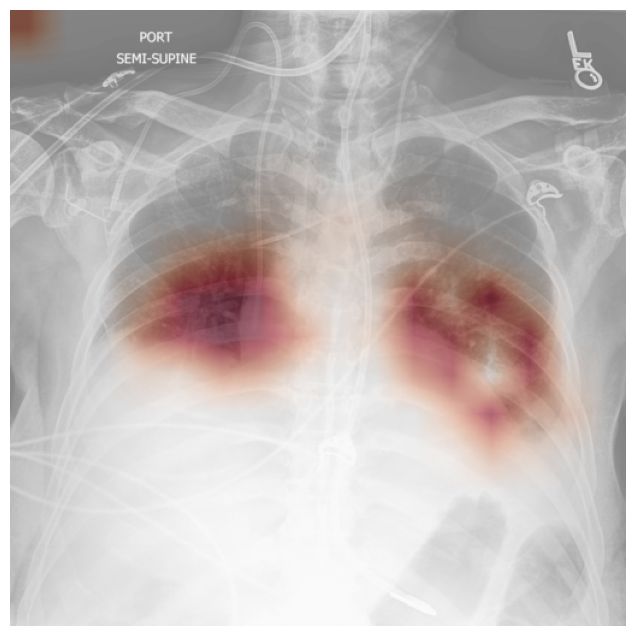} &
        \includegraphics[width=\linewidth]{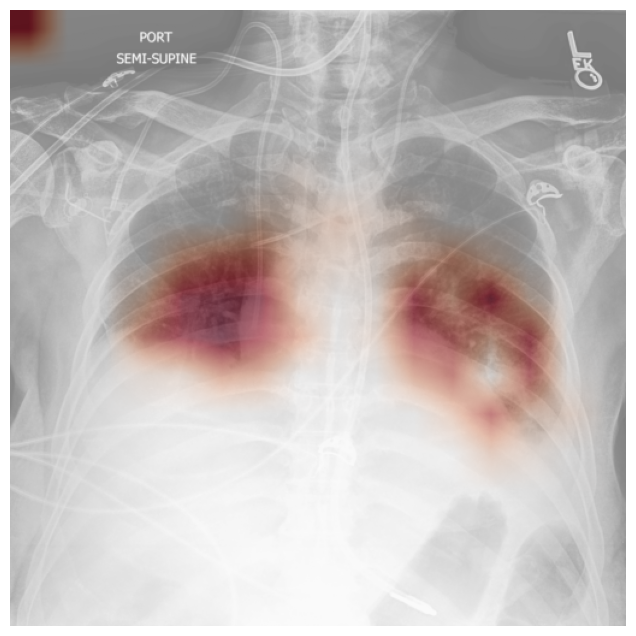} &
        \includegraphics[width=\linewidth]{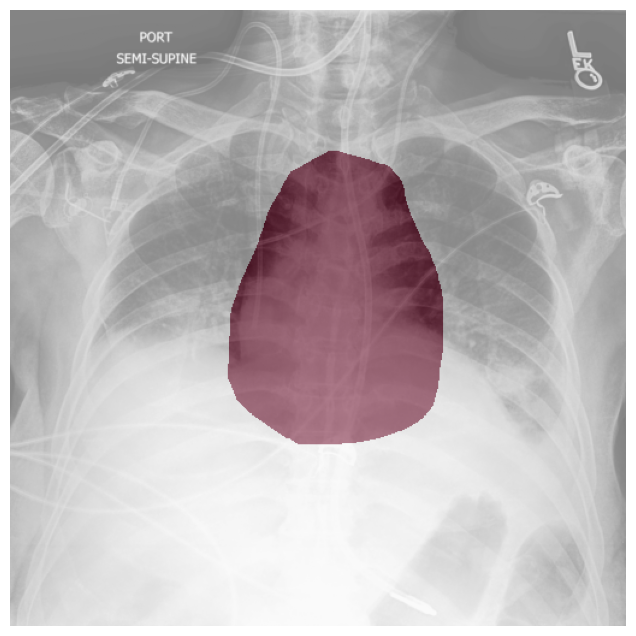} \\

        \bottomrule
    \end{tabular}
    \caption{Representative saliency maps produced by MedicalPatchNet and three post-hoc methods. Each row displays the same chest X-ray for a given pathology together with its ground-truth label (``True'' or ``False''). Columns compare MedicalPatchNet’s raw patch logits with Grad-CAM, Grad-CAM++, and Eigen-CAM applied to an EfficientNetV2-S baseline. In the MedicalPatchNet maps, \textcolor[HTML]{9E1127}{red} denotes evidence supporting the class and \textcolor[HTML]{0F457E}{blue} denotes evidence against it, whereas Grad-CAM–based maps visualize only positive (red) contributions. Eigen-CAM is class-agnostic and therefore does not generate class-specific saliency maps. Interestingly, for the wrongly diagnosed pneumothorax, all four explainability methods point to the chest tube, revealing that the model used a shortcut, with MedicalPatchnet denoting its course most clearly.}
    \label{fig:activationMaps}
\end{figure}

\flushbottom
\newpage
\newpage

\subsection*{Performance Difference Explanation}
Our results differ from those in the original CheXlocalize paper\cite{Saporta2022CheXlocalize} because of methodological choices, not inherent limitations of our approach: we trained and evaluated single models—one MedicalPatchNet and one EfficientNet\chreplaced{V2-S}{-B0}—whereas Saporta et al. generated 120 checkpoints for each of three models, chose the ten best checkpoints per pathology, and ensembled those selections, a strategy that inflates metrics relative to single-model evaluation. The preprint version of their work\footnote{\url{https://www.medrxiv.org/content/10.1101/2021.02.28.21252634v1}} reports single-network results comparable to ours, reinforcing that the observed gap stems from ensemble versus single-model evaluation.

\section{Self-Explainable Methods} 
As our method falls in the category of self explainable methods, its natural to compare it to other self explainable approaches.
For this we compare it to the ProtoPNet\cite{Chen2019ThisLooksLikeThat} and the PIPNet\cite{PIPNet}. We haven't made a comparison to the XProtoNet\cite{kim2021xprotonet}, as the code was not publicly available to reproduce it.

\subsection*{ProtoPNet}

ProtoPNet\cite{Chen2019ThisLooksLikeThat} is a self-explainable architecture that classifies images by comparing internal image representations to a small set of learned prototypes. Each prototype is intended to represent a recurring visual pattern, and predictions are obtained by aggregating evidence of the form ``this region looks like that prototype''. In contrast to post-hoc saliency methods, the explanation is part of the forward pass: the model decision is explicitly constructed from prototype matches.

\subsubsection*{Prototype matching in latent space}
At a high level, ProtoPNet consists of (i) a convolutional backbone that maps an input image $x$ to a spatial feature map $z=f(x)$, (ii) a set of learnable prototypes $\{p_j\}$ that live in the same latent space as local feature vectors of $z$, and (iii) a linear classification layer on top of prototype similarity scores.

Concretely, each prototype $p_j$ is compared to all spatial locations of the feature map $z$ by computing a distance (typically an $\ell_2$ distance in feature space) between $p_j$ and each local feature vector of $z$. This yields a spatial distance map per prototype, which is converted into a similarity map. ProtoPNet then performs a spatial max operation per prototype, so that each prototype contributes a single scalar similarity score given by its best-matching location in the image. The final class logits are computed as a weighted sum of these prototype scores. In our multi-label setting, the resulting logits are passed through a sigmoid to obtain per-label probabilities.

\subsubsection*{Push / projection step (prototype grounding)}
A key ingredient for interpretability is the prototype grounding procedure (often called push or projection). After (or during) training, each prototype is replaced by the latent feature vector of the closest training patch (in the backbone feature space), typically constrained to come from an image that contains the prototype's target class. This step ties each abstract prototype vector to an actual, human-inspectable training image region: the model can now display, for each prototype, the training patch it was projected onto and the corresponding most similar patch in a test image. Importantly, this grounding does not change the inference rule (matching in latent space + max aggregation + linear classifier); it only enforces that the stored prototypes correspond to real training patches, making the ``this looks like that'' explanations visually meaningful.

\subsubsection*{Results on CheXlocalize}
Following the reviewer suggestion, we evaluated ProtoPNet on the CheXlocalize test set (668 samples, 14 labels). Overall performance was close to random, with a macro-average AUROC of 0.502. Notably, threshold selection degenerated for most labels: the optimal threshold was 0.0 for 13/14 labels, which results in predicting nearly all cases as positive (sensitivity $\approx 1.0$ and specificity $\approx 0.0$), consistent with poor separability of the predicted scores.

\begin{table}[H]
\centering
\begin{tabular}{lccccc}
\toprule
Label & AUROC & Accuracy & Sensitivity & Specificity & F1 \\
\midrule
No Finding & 0.523 & 0.163 & 1.000 & 0.000 & 0.281 \\
Enlarged Cardiomediastinum & 0.507 & 0.446 & 1.000 & 0.000 & 0.617 \\
Cardiomegaly & 0.521 & 0.262 & 1.000 & 0.000 & 0.415 \\
Lung Opacity & 0.479 & 0.464 & 1.000 & 0.000 & 0.634 \\
Lung Lesion & 0.490 & 0.021 & 1.000 & 0.000 & 0.041 \\
Edema & 0.512 & 0.127 & 1.000 & 0.000 & 0.226 \\
Consolidation & 0.525 & 0.052 & 1.000 & 0.000 & 0.100 \\
Pneumonia & 0.492 & 0.021 & 1.000 & 0.000 & 0.041 \\
Atelectasis & 0.493 & 0.266 & 1.000 & 0.000 & 0.421 \\
Pneumothorax & 0.493 & 0.015 & 1.000 & 0.000 & 0.030 \\
Pleural Effusion & 0.520 & 0.180 & 1.000 & 0.000 & 0.305 \\
Pleural Other & 0.493 & 0.012 & 1.000 & 0.000 & 0.024 \\
Fracture & 0.497 & 0.991 & 0.000 & 1.000 & 0.000 \\
Support Devices & 0.489 & 0.472 & 1.000 & 0.000 & 0.641 \\
\midrule
Average & 0.502 & 0.249 & 0.929 & 0.071 & 0.270 \\
\bottomrule
\end{tabular}%
\caption{ProtoPNet performance on the CheXlocalize test set (14-label multi-label classification). Metrics are reported per label and averaged across labels.}
\label{tab:protopnet_chexlocalize}
\end{table}

\subsubsection*{Why ProtoPNet underperformed}
We hypothesize that the poor performance is primarily caused by a mismatch between ProtoPNet's training objective and the properties of chest X-ray multi-label classification.

First, ProtoPNet was originally designed for single-label classification, where each image belongs to exactly one class. Its prototype assignment and separation mechanisms implicitly assume that evidence should be class-exclusive: patches that support class $A$ are simultaneously treated as evidence against class $B$. In chest X-rays, however, labels frequently co-occur (e.g., cardiomegaly with edema), so the same image region can legitimately support multiple labels. In such a setting, enforcing strong separation between prototypes of different classes can become contradictory and may push the representation toward trivial solutions.

Second, ProtoPNet aggregates prototype evidence via a spatial max over the feature map. This makes the prediction highly sensitive to a single best-matching location per prototype. For radiographs, many findings are subtle, extended, or context-dependent (e.g., cardiomegaly depends on global thoracic context rather than a single local texture). A max-based evidence model can therefore be brittle: it may either lock onto ubiquitous, non-specific structures (ribs, mediastinum edges, text overlays, devices) or produce near-constant similarity scores that offer little discriminative power, both of which are consistent with AUROCs close to 0.5.

Third, strong class imbalance (especially for rare labels such as fracture) interacts poorly with a fixed prototype budget per class. If prototypes cannot reliably ground onto truly class-specific patches during projection, they may collapse onto background patterns; the subsequent linear layer can then drift into degenerate threshold behaviour (here: near-always-positive predictions for most labels and near-always-negative for fracture), yielding high sensitivity but vanishing specificity.

Overall, these observations suggest that ProtoPNet, in its standard form, is not well aligned with the multi-label, co-occurring, and often spatially diffuse nature of chest X-ray findings. Addressing this likely requires multi-label-aware prototype objectives (e.g., relaxing inter-class separation for co-occurring labels), alternative evidence pooling (beyond a pure max), and/or prototype allocation strategies that reflect the prevalence and heterogeneity of each pathology.

\subsection*{PIPNet}

\subsubsection*{Method overview}
PIP-Net\cite{PIPNet} (Patch-based Intuitive Prototypes Network) is a prototype-based, inherently interpretable classifier that aims to explain predictions via a set of prototypical visual patterns. Conceptually, the model acts like a ``scoring sheet'': it measures how strongly an input image matches each learned prototype and then aggregates these prototype scores with non-negative class weights. The non-negativity constraint is intended to support case-based reasoning, where prototypes can only add evidence for a class (rather than subtracting it).

\subsubsection*{Prototype scoring and aggregation}
PIPNet turns an image into a set of prototype activations and then combines them into class scores in a strictly additive way.
Conceptually, the backbone produces a spatial feature map with $K$ channels; PIPNet treats each channel as one prototype, so in our configuration ($K=768$ with a ConvNeXt-Tiny backbone) the model has 768 prototypes.

At each spatial location, the model computes how strongly each prototype ``matches'' the local image content, and then normalizes these responses across the $K$ prototypes (so prototypes compete for attention at each location). For each prototype $k$, PIPNet keeps only its strongest match anywhere in the image, yielding one scalar score $s_k(x)$ per prototype (intuitively: ``how much does the image contain something that looks like prototype $k$?'').

A final non-negative linear layer then aggregates these prototype scores into one score per class by a weighted sum,
\[
\ell_c(x)=\sum_{k=1}^{K} w_{c,k}\, s_k(x), \qquad w_{c,k}\ge 0,
\]
so prototypes can only contribute positive evidence for a class. Explanations follow directly: a class is supported by those prototypes with large contributions $w_{c,k}s_k(x)$, and each such prototype can be localized by the image region where its maximum activation was attained.

\subsubsection*{Training recipe used in our experiments}
We used the authors' public reference implementation from their GitHub repository (\href{https://github.com/M-Nauta/PIPNet}{github.com/M-Nauta/PIPNet}) and applied it to the CheXpert dataset.
In this implementation, training follows a two-stage procedure.

In the first stage, the network is trained without using class labels to shape the prototype space.
The objective encourages prototype activations to be stable under image augmentations (i.e., two augmented views of the same image should activate similar prototypes) and includes an additional regularization term that discourages overly sparse prototype usage.
During this stage, the final classification layer is kept fixed (not updated), so the model focuses on learning a prototype representation rather than class mappings.

In the second stage, the model is fine-tuned with supervision to learn how prototypes map to target classes via the non-negative classification layer.
The implementation enforces sparsity in this layer by repeatedly shrinking small weights toward zero during training, which promotes a small set of ``active'' prototypes per class.
At inference time, weak prototype activations are additionally thresholded (set to zero), so only sufficiently strong prototype matches contribute to the final class scores.
As a result, the overall decision is intentionally driven by a limited number of high-confidence prototype matches, which is meant to make explanations concise and prototype-based.

\subsubsection*{Results on chest X-ray classification}
We evaluated PIPNet on the CheXpert-derived test set used throughout this work (668 images, 14 labels; the same image subset as the CheXlocalize test split, but evaluated for classification only). A critical failure mode emerged: the trained model produced meaningful predictions for only two labels, while all other labels collapsed to degenerate outputs (Table~\ref{tab:pipnet_chexpert}). Concretely, No Finding achieved an AUROC of 0.8374 and Support Devices reached an AUROC of 0.6127, whereas the remaining 12 labels yielded zero scores across all reported metrics in our evaluation.

\begin{table}[H]
\centering
%\resizebox{\textwidth}{!}{%
\begin{tabular}{lccccc}
\toprule
Label & AUROC & Accuracy & Sensitivity & Specificity & F1 \\
\midrule
No Finding & 0.8374 & 0.8323 & 0.7982 & 0.8390 & 0.6084 \\
Enlarged Cardiomediastinum & 0.0000 & 0.0000 & 0.0000 & 0.0000 & 0.0000 \\
Cardiomegaly & 0.0000 & 0.0000 & 0.0000 & 0.0000 & 0.0000 \\
Lung Opacity & 0.0000 & 0.0000 & 0.0000 & 0.0000 & 0.0000 \\
Lung Lesion & 0.0000 & 0.0000 & 0.0000 & 0.0000 & 0.0000 \\
Edema & 0.0000 & 0.0000 & 0.0000 & 0.0000 & 0.0000 \\
Consolidation & 0.0000 & 0.0000 & 0.0000 & 0.0000 & 0.0000 \\
Pneumonia & 0.0000 & 0.0000 & 0.0000 & 0.0000 & 0.0000 \\
Atelectasis & 0.0000 & 0.0000 & 0.0000 & 0.0000 & 0.0000 \\
Pneumothorax & 0.0000 & 0.0000 & 0.0000 & 0.0000 & 0.0000 \\
Pleural Effusion & 0.0000 & 0.0000 & 0.0000 & 0.0000 & 0.0000 \\
Pleural Other & 0.0000 & 0.0000 & 0.0000 & 0.0000 & 0.0000 \\
Fracture & 0.0000 & 0.0000 & 0.0000 & 0.0000 & 0.0000 \\
Support Devices & 0.6127 & 0.6332 & 0.2349 & 0.9887 & 0.3766 \\
\midrule
Average & 0.1036 & 0.1047 & 0.0738 & 0.1305 & 0.0704 \\
\bottomrule
\end{tabular}%
%}
\caption{\label{tab:pipnet_chexpert}PIPNet performance on the CheXpert-derived test set (668 images, 14 labels). Values are point estimates (no bootstrap confidence intervals available for this experiment).}
\end{table}

\subsubsection*{Why might PIPNet fail here?}
The observed collapse is most plausibly explained by a combination of interacting factors. Importantly, the following are hypotheses inferred from the observed behavior and from the training configuration; confirming them would require targeted ablations.

First, we hypothesize a mismatch between the default PIPNet learning objective and the CheXpert setting. CheXpert is genuinely multi-label, with frequent co-occurrence of findings. In contrast, the training recipe we used enforces competition between classes, which effectively encourages a single ``winning'' label per image. In such a regime, gradients may be systematically biased toward frequent or visually dominant labels, while suppressing learning signals for concurrent positives. This could explain why the model appears to concentrate on a very small subset of labels and fails to develop usable decision rules for most findings.

Second, we hypothesize that the combination of (i) a non-negative classifier, (ii) explicit sparsity enforcement during training, and (iii) inference-time thresholding of weak prototype activations can lead to premature collapse. If small class-to-prototype weights are repeatedly driven to exactly zero early in training, many labels may lose most of their effective supervised pathways. Once a label has effectively ``lost access'' to prototypes, recovering becomes difficult because its gradients cannot easily re-establish informative prototype-to-class connections under the same sparsifying dynamics.

Third, we hypothesize that the self-supervised pretraining objective can dominate representation learning in this medical domain. Prototype channels may become highly invariant to augmentations (as intended) but not sufficiently aligned with subtle radiographic cues required for pathology discrimination. The supervised stage would then need to re-purpose these prototypes for clinical findings while simultaneously operating under strong sparsity pressure, which is an unfavorable optimization problem.

Fourth, extreme class imbalance likely exacerbates all of the above. Several findings are rare in the evaluated split (e.g., pneumothorax, fracture, pleural other), and in the absence of explicit imbalance-aware weighting, these labels contribute comparatively little to the loss. We hypothesize that this makes them particularly prone to being ignored during optimization, further increasing the probability of degenerate outputs.

Finally, some thoracic findings are spatially diffuse or context-dependent (e.g., global opacity patterns or shape changes spanning large anatomical regions). We hypothesize that capturing such signals via a small number of local max-pooled prototype matches may be intrinsically difficult, which could further bias the model toward trivial solutions focused on a few visually distinct categories.

Overall, in the present configuration, PIPNet did not provide a viable performance--interpretability trade-off for chest X-ray multi-label classification: for most labels, the model failed to learn usable predictors, and consequently prototype-based explanations would not be meaningful in those cases.

\subsection*{Why we did not perform an extensive CheXlocalize evaluation for ProtoPNet and PIPNet}
In principle, CheXlocalize would be the natural benchmark to compare localization behavior across self-explainable methods. However, localization metrics on CheXlocalize are only meaningful if the underlying classifier has learned a non-trivial decision function for the target labels; otherwise, saliency maps (and any thresholded masks derived from them) largely reflect noise or degenerate prediction behavior rather than clinically interpretable evidence.

In our experiments, both ProtoPNet and PIPNet were trained on CheXpert using the authors' public implementations and their default training hyperparameters. Under this setting, neither method produced a reliable multi-label chest X-ray classifier: ProtoPNet remained close to chance level (macro AUROC $\approx 0.5$) with largely degenerate threshold behavior, and PIPNet collapsed to meaningful predictions for only two labels while yielding trivial outputs for the remaining classes. Given these failure modes at the classification level, a full CheXlocalize localization analysis would not provide informative or fair conclusions about explainability, because the resulting maps would primarily quantify artifacts of collapse rather than genuine model reasoning.

\section{Generalizability across backbones (ConvNeXt)}
To assess whether MedicalPatchNet’s patch-wise, self-explainable aggregation generalizes beyond the EfficientNet family, we repeated the CheXpert/CheXlocalize classification evaluation using a ConvNeXt-Base backbone. We trained (i) \emph{MedicalPatchNet -- ConvNextBase} and (ii) a standard \emph{ConvNext -- 512$\times$512} image-level classifier under identical preprocessing and optimization settings. Table~\ref{tab:convnext_backbone} reports mean AUROC, accuracy, sensitivity, and specificity averaged across the 10 CheXlocalize labels (*) and across all 14 CheXpert labels (All). For the 10-label average (*), \emph{Fracture, No Finding, Pleural Other, and Pneumonia} are excluded.

\begin{table}[t]
\centering
\begin{tabular}{lcccccccc}
\toprule
 & \multicolumn{2}{c}{AUROC} & \multicolumn{2}{c}{Accuracy} & \multicolumn{2}{c}{Sensitivity} & \multicolumn{2}{c}{Specificity}\\
\cmidrule(lr){2-3} \cmidrule(lr){4-5} \cmidrule(lr){6-7} \cmidrule(lr){8-9}
Model & * & All & * & All & * & All & * & All\\
\midrule
MedicalPatchNet -- ConvNextBase & 0.903 & 0.898 & 0.853 & 0.863 & 0.737 & 0.719 & 0.862 & 0.869 \\
ConvNext -- 512$\times$512      & 0.903 & 0.899 & 0.857 & 0.865 & 0.755 & 0.737 & 0.867 & 0.871 \\
\midrule
Difference
& 0.000 & -0.001 & -0.004 & -0.002 & -0.018 & -0.018 & -0.005 & -0.002 \\
\bottomrule
\end{tabular}
\caption{Classification performance comparison using a ConvNeXt backbone. The metrics represent average values computed across the 10 classes from the CheXlocalize dataset (*) and across all 14 classes from the CheXpert dataset (All). For the 10-class average (*), Fracture, No Finding, Pleural Other, and Pneumonia were excluded.}
\label{tab:convnext_backbone}
\end{table}

\end{document}